\documentclass{article}
\usepackage{xcolor}
\usepackage[final]{corl_2026} 

\usepackage{adjustbox}
\usepackage{booktabs}
\usepackage{csquotes}
\MakeOuterQuote{"}
\usepackage{enumitem}
\usepackage{etoolbox}
\usepackage{graphicx}
\usepackage{amsmath} 
\usepackage{amssymb}
\usepackage[noabbrev,capitalise]{cleveref}
\usepackage{makecell}
\usepackage{mleftright}
\usepackage{multirow}
\usepackage{paracol}
\usepackage{siunitx}
\usepackage{subcaption}
\usepackage{tabularx}
\usepackage{wrapfig}
\usepackage{xfrac}
\usepackage{tikz}
\usetikzlibrary{arrows.meta,calc,colorbrewer,positioning}
\usepackage{pgfplots}
\pgfplotsset{compat=1.18}
\pgfplotsset{
  log ticks with fixed point,
}
\pgfplotsset{
  ticklabel style={
    /pgf/number format/1000 sep={}
  }
}
\pgfplotsset{
  ticklabel style={font=\small}
}
\usepgfplotslibrary{groupplots}
\usepackage{pgfplotstable}

\hypersetup{
  pdfauthor={Jan Schneider, Mridul Mahajan, Le Chen, Simon Guist, Bernhard Schölkopf, Ingmar Posner, Dieter Büchler},
  pdftitle={Sim-to-Real Transfer for Muscle-Actuated Robots via Generalized Actuator Networks},
  pdfkeywords={sim-to-real, reinforcement learning, muscle actuation, PAM, tendons}
}

\title{Sim-to-Real Transfer for Muscle-Actuated Robots via Generalized Actuator Networks}

\author{Jan Schneider$^{1,2}$, Mridul Mahajan$^{3}$, Le Chen$^{1}$, Simon Guist$^{1}$, \\ 
\vspace{0.2cm}
\bfseries Bernhard Schölkopf$^{1,4}$, Ingmar Posner$^{2}$, and Dieter Büchler$^{5,6,7}$ \\
$^{1}$Max Planck Institute for Intelligent Systems, Tübingen, $^{2}$University of Oxford,\\$^{3}$Boston University, $^{4}$ELLIS Institute Tübingen, $^{5}$Johannes Kepler University Linz, \\
$^{6}$CIFAR AI Chair, $^{7}$Alberta Machine Intelligence Institute (Amii)}%

\DeclareMathOperator{\invdyn}{invdyn}
\DeclareMathOperator{\step}{step}

\DeclareMathOperator{\Sd}{SD}

\newcommand{\includeTrimmedBallInCupImg}[2]{%
  \includegraphics[width=#1,trim={1200 150 1100 750},clip]{#2}%
}

\newcommand{\includeMoreTrimmedBallInCupImg}[2]{%
  \includegraphics[width=#1,trim={1350 350 1330 725},clip]{#2}%
}
\newcommand{\includeMoreTrimmedShiftedLeftBallInCupImg}[2]{%
  \includegraphics[width=#1,trim={1000 350 1680 725},clip]{#2}%
}

\newcommand{\qreal}[1]{\boldsymbol{q}_{#1}}
\newcommand{\qpred}[1]{\boldsymbol{\hat{q}}_{#1}}
\newcommand{\qsim}[1]{\boldsymbol{\tilde{q}}_{#1}}
\newcommand{\qdreal}[1]{\boldsymbol{\dot{q}}_{#1}}
\newcommand{\qdpred}[1]{\boldsymbol{\dot{\hat{q}}}_{#1}}
\newcommand{\qdsim}[1]{\boldsymbol{\dot{\tilde{q}}}_{#1}}
\newcommand{\qddreal}[1]{\boldsymbol{\ddot{q}}_{#1}}
\newcommand{\qddpred}[1]{\boldsymbol{\ddot{\hat{q}}}_{#1}}
\newcommand{\qddsim}[1]{\boldsymbol{\ddot{\tilde{q}}}_{#1}}
\newcommand{\dt}{\Delta t}
\newcommand{\M}[1]{\boldsymbol{M}\mleft(#1\mright)}
\newcommand{\Minv}[1]{\boldsymbol{M}\mleft(#1\mright)^{-1}}
\newcommand{\cor}[2]{\boldsymbol{c}\mleft(#1, #2\mright)}
\newcommand{\grav}[1]{\boldsymbol{g}\mleft(#1\mright)}
\newcommand{\tor}[1]{\boldsymbol{\tau}_{#1}}
\newcommand{\torpred}[1]{\boldsymbol{\hat{\tau}}_{#1}}
\newcommand{\ctrl}[1]{\boldsymbol{u}_{#1}}
\newcommand{\qsimzero}[1]{\boldsymbol{\bar{q}}_{#1}}
\newcommand{\qdsimzero}[1]{\boldsymbol{\dot{\bar{q}}}_{#1}}

\pgfmathsetmacro{\golden}{(1+sqrt(5))/2}
\pgfmathsetmacro{\invgolden}{1/\golden}
\newlength{\goldenheight}

\makeatletter
\pgfplotsset{
    golden width landscape/.code={
        \pgfkeys{/pgfplots/width=#1}%
        \pgfmathsetlength{\goldenheight}{#1 * \invgolden}%
        \pgfkeys{/pgfplots/height=\goldenheight}%
        \pgfkeys{/pgfplots/scale only axis}%
    }
}
\pgfplotsset{
    golden width portrait/.code={
        \pgfkeys{/pgfplots/width=#1}%
        \pgfmathsetlength{\goldenheight}{#1 * \golden}%
        \pgfkeys{/pgfplots/height=\goldenheight}%
        \pgfkeys{/pgfplots/scale only axis}%
    }
}
\makeatother

\newcommand{\insetxpos}{0.115\linewidth}
\newcommand{\insetypos}{0.162\linewidth}
\newcommand{\insetwidth}{0.2\linewidth}
\newcommand{\insetheight}{0.27\linewidth}

\newcommand{\datasetsizefontsize}{\footnotesize}
\newcommand{\datasetsizeheight}{0.43\linewidth}

\definecolor{lightRedPink}{HTML}{FBB4AE}
\definecolor{softBlue}{HTML}{B3CDE3}
\definecolor{lessSoftBlue}{HTML}{3C8BCC}
\definecolor{mintGreen}{HTML}{CCEBC5}
\definecolor{lavender}{HTML}{DECBE4}  
\definecolor{peach}{HTML}{FED9A6}
\definecolor{paleYellow}{HTML}{FFFFCC}
\definecolor{beige}{HTML}{E5D8BD}
\definecolor{lightPink}{HTML}{FDDAEC}
\definecolor{offWhite}{HTML}{F2F2F2}  

\newlength{\explPlotWidth}
\newcommand{\pmin}[1]{p_\mathrm{min}^{(#1)}}
\newcommand{\pmax}[1]{p_\mathrm{max}^{(#1)}}
\newcommand{\pdes}[1]{p_\mathrm{com}^{(#1)}}
\newcommand{\pobs}[1]{p\ifblank{#1}{}{^{(#1)}}}

\DeclareSIUnit\bar{bar}

\begin{document}

\maketitle

\begin{abstract}
Tendon drives paired with soft muscle actuation enable faster and safer robots while potentially accelerating skill acquisition.
Still, these systems are rarely used in practice due to inherent nonlinearities, friction, and hysteresis, which complicate modeling and control.
So far, these challenges have hindered policy transfer from simulation to real systems. 
To bridge this gap, we propose a sim-to-real pipeline that learns a neural network model of this complex actuation and leverages established rigid body simulation for the arm dynamics and interactions with the environment.
Our method, called Generalized Actuator Network~(GenAN), enables actuation model identification across a wide range of robots by learning directly from joint position trajectories rather than requiring torque sensors.
Using GenAN on PAMY2, a tendon-driven robot powered by pneumatic artificial muscles, we successfully deploy dynamic but precise goal-reaching, ball-in-a-cup, and table tennis policies, trained entirely in simulation.
To the best of our knowledge, this result constitutes the first successful sim-to-real transfer for a four-degrees-of-freedom muscle-actuated robot arm.
\end{abstract}

\begin{figure}[h]
    \vspace{0.2cm}
    \begin{subfigure}[b]{0.42\linewidth}
        \includeTrimmedBallInCupImg{0.495\linewidth}{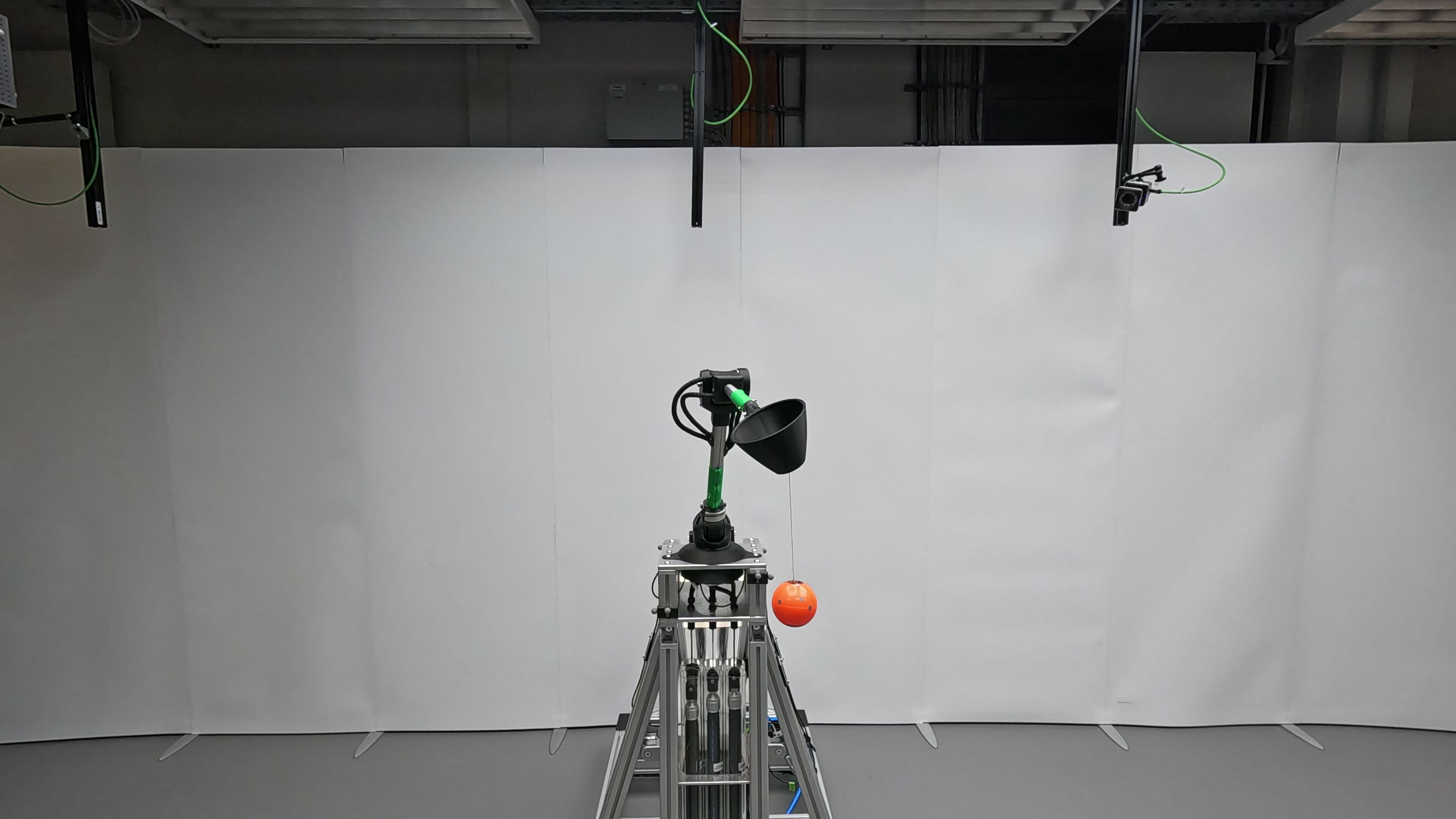}\hfill
        \includeTrimmedBallInCupImg{0.495\linewidth}{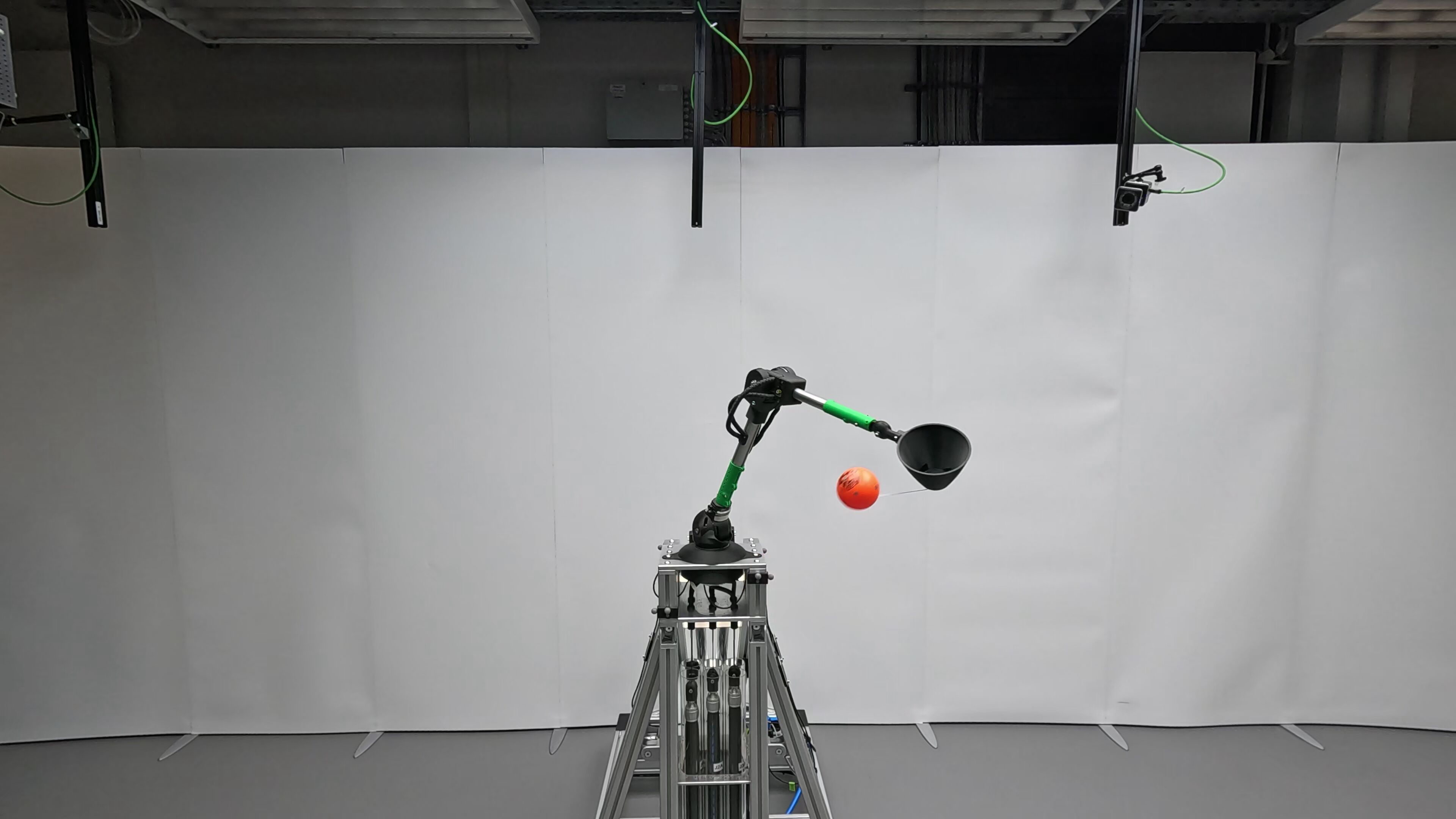}\\[0.005\linewidth]
        \includeTrimmedBallInCupImg{0.495\linewidth}{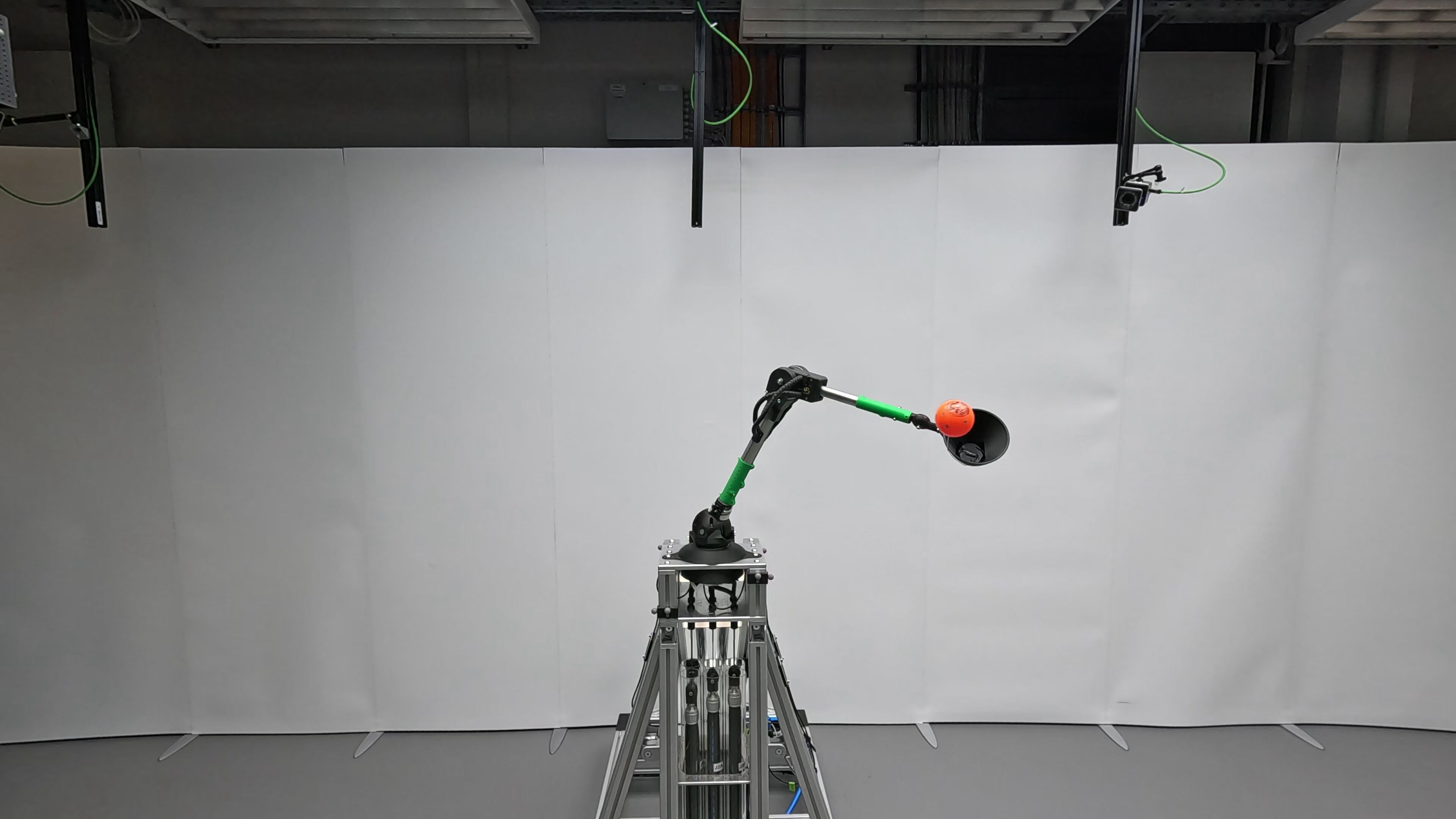}\hfill
        \includeTrimmedBallInCupImg{0.495\linewidth}{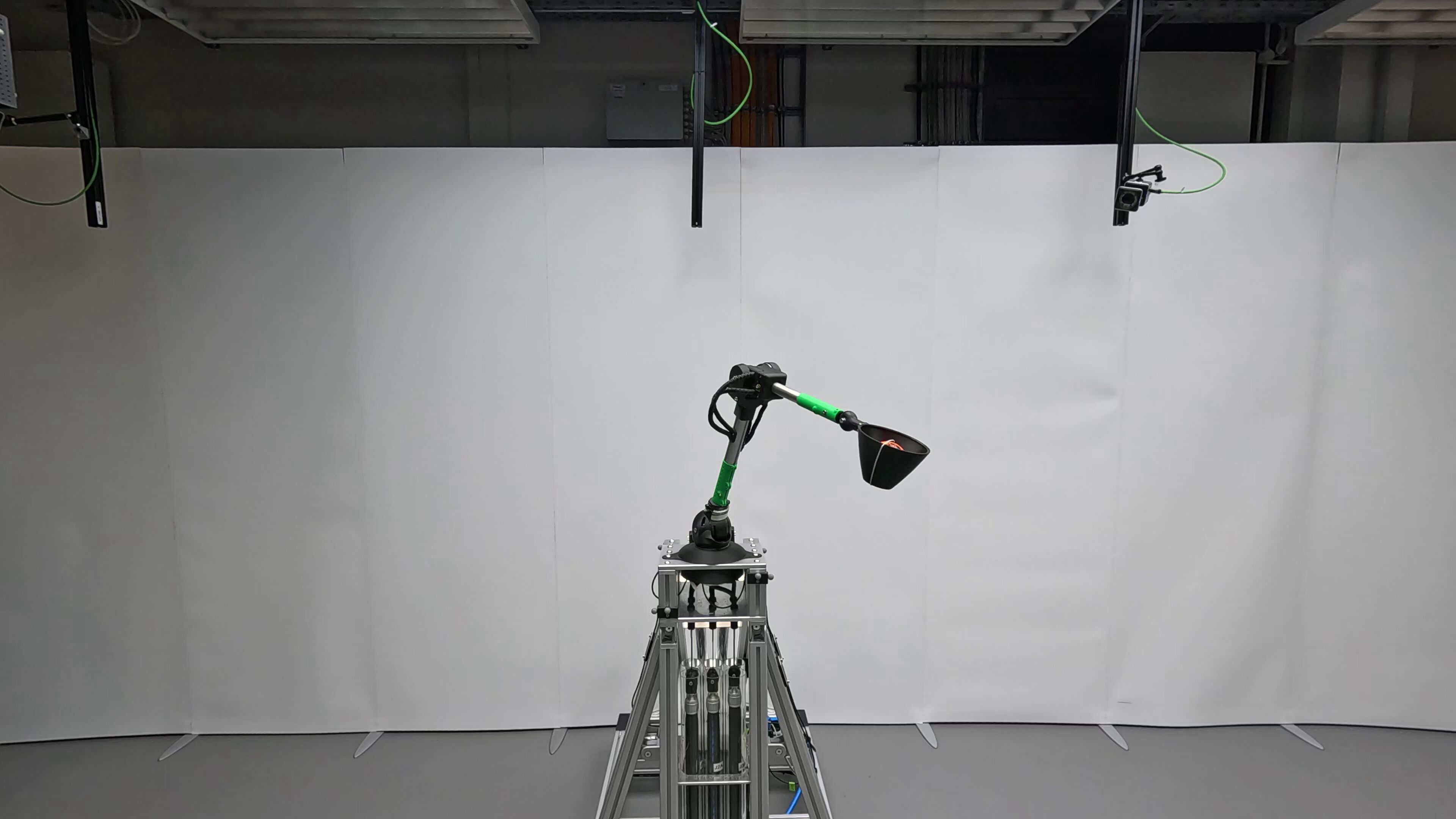}
    \end{subfigure}
    \hfill
    \begin{subfigure}[b]{0.5687\linewidth}
        \includegraphics[width=\linewidth,trim={230 0 640 360},clip]{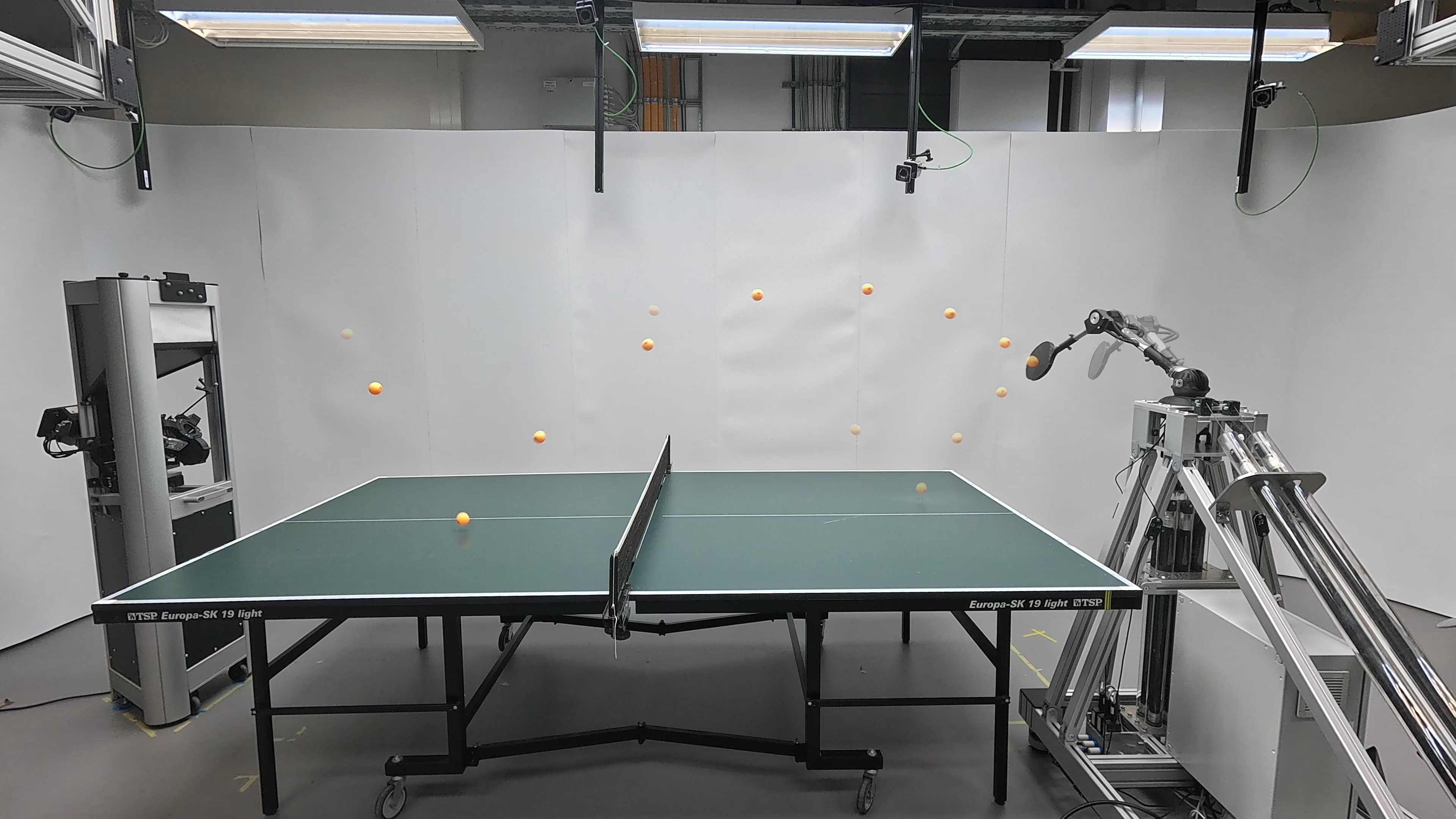}
    \end{subfigure}
    \caption{
        Rollouts of the ball-in-a-cup (left) and table tennis (right) policies on a tendon-driven robot actuated by pneumatic artificial muscles.
        The policies were trained entirely in simulation with a learned actuator model and transferred zero-shot to the physical robot.
    }
    \label{fig:ball_in_a_cup_rollout}
\end{figure}

\section{Introduction}
\label{sec:introduction}
Tendon-driven robot arms paired with soft actuation present a promising alternative to classical rigid and motor-driven systems~\cite{buchler2016lightweight,mori2018high,guist2024safe,kawaharazuka2019component}.
These designs substantially reduce the moving masses by placing the actuators at the base.
Combined with powerful pneumatic actuators, e.g., pneumatic artificial muscles~(PAMs), these systems excel at athletic tasks~\cite{mori2018high,guist2024safe}.
Furthermore, the lightweight design and mechanical compliance greatly reduce contact forces upon collision~\cite{guist2024safe}, making these robots safer to operate around humans even at higher speeds.
\Citet{wochner2023learning} also found that muscle actuation can facilitate more sample-efficient skill learning.

Despite these advantages and the prevalence of muscle and tendon actuation in biological systems, robots are rarely equipped with such actuation. 
The primary obstacle lies in the inherent modeling challenges that impede precise control~\cite{hao2017modeling,minh2010cascade,qin2022active}.
Muscle dynamics are highly nonlinear, subject to hysteresis, and can exhibit time-varying properties, e.g., due to temperature fluctuations. 
Tendons introduce friction that depends on the joint positions since they are routed through the robot. 
Consequently, existing works often resort to learning-based approaches, such as iterative learning control~\cite{ma2022learning} or reinforcement learning~(RL)~\cite{buchler2022learning}, when generating behaviors with these robots.

Many recent successes in robot learning, including locomotion~\cite{radosavovic2024real,seo2025learning}, table tennis~\cite{su2025hitter,durr2026outplaying}, and soccer~\cite{xu2025learning}, have been enabled by sim-to-real transfer. 
Such methods allow for learning complex behaviors without requiring vast amounts of interactions with the physical robot.
Through the use of massive parallelization, enabled by GPU-based simulators, these techniques also vastly reduce the training time, sometimes from hours or days to mere minutes~\cite{rudin2022learning,seo2025learning}.
Prolonged motion execution on a robot is also energy-intensive, accelerates mechanical wear, and typically necessitates extensive safety considerations.
Moreover, automatically resetting the physical environment to a fixed initial condition can be difficult, especially if the robot is handling external objects.

A common technique to bridge the \emph{sim-to-real gap}, i.e., the difference between simulated and real dynamics, is \emph{domain randomization~(DR)}~\cite{peng2018sim,muratore2022robot}.
DR adds noise to the physics parameters during training to robustify the policy to dynamics variations.
However, for muscle-actuated robots, the sim-to-real gap is vastly greater due to the aforementioned modeling challenges.
Even DR relies on approximately correct dynamics models, which have remained elusive for PAM-actuated robots~\cite{tondu2012modelling,buchler2018control}.
Compensating for this wide sim-to-real gap with DR would require increasing the parameter noise significantly, which typically degrades policy performance~\cite{tiboni2024domain}.
An alternative to sim-to-real learning for muscle-actuated robots is hybrid-sim-and-real (HySR) training~\cite{buchler2022learning}, where the objects are simulated but the robot remains real.
However, even the sample-efficient approach of Guist et al.~\cite{guist2023hindsight} requires many hours of real-world interactions, limiting the scalability to complex tasks.

The core idea of this work is to utilize a known dynamics model for the analytically tractable model components while learning the complex actuation dynamics from data, making use of expressive neural network models.
We take inspiration from the actuator network framework~\cite{hwangbo2019learning}, which learns an actuation model from torque measurements and focuses on more well-behaved series elastic actuators.
This method is not applicable to many muscle- and tendon-actuated robots, as these systems are typically not equipped with torque sensors.
Therefore, we introduce the \emph{Generalized Actuator Network~(GenAN)}, which learns the actuator dynamics directly from joint position measurements.
Since our method requires only joint encoders, which are available on most robot platforms, it \emph{generalizes} actuator model learning to a wide range of robots with different actuation types.
To the best of our knowledge, we demonstrate the \textit{first} sim-to-real transfer of dynamic policies for a multi-joint, muscle- and tendon-actuated robot (see~\cref{fig:ball_in_a_cup_rollout}).
Our contributions are threefold:
\begin{enumerate}[topsep=0pt, itemsep=0pt, parsep=0pt]
    \item We introduce the Generalized Actuator Network (GenAN), a learned actuator model for robots without torque sensors that achieves state-of-the-art accuracy through a differentiable position-matching loss.
    \item Using GenAN, we demonstrate the first successful sim-to-real transfer for a four-degrees-of-freedom robot arm with complex muscle and tendon actuation.
    \item We explore the utility of GenAN ensembles for preventing policy overfitting to model uncertainty, especially in low-data regimes.
\end{enumerate}
Links to code, datasets, and videos are available on the \href{https://webdav.tuebingen.mpg.de/genan/}{project webpage}.

\section{Related work}
\label{sec:related_work}
This paper tackles sim-to-real reinforcement learning for muscle-actuated robots and builds on ideas from actuator model learning.
We discuss the relation to the existing literature in these fields.

\subsection{Sim-to-real transfer with learned actuator models}

The idea of utilizing \emph{actuator networks}, i.e., learned actuator models, for sim-to-real learning was first introduced by \Citet{hwangbo2019learning} in the context of quadruped locomotion.
They train a neural network to predict joint torques produced by the series elastic actuation of a quadruped robot.
For the target labels, they use torque measurements, which limits their method to robots that are equipped with torque sensors.
They demonstrate zero-shot sim-to-real transfer for locomotion and fall recovery behaviors.
The approach was subsequently used in several works on quadruped locomotion~\cite{rudin2022learning,lee2020learning,ji2023dribblebot,eichmann2025lauron}, demonstrating its utility for learning agile locomotion in diverse terrains.

Despite these successes, the application of actuator networks beyond quadruped locomotion is only slowly gaining traction.
\Citet{spinelli2024reinforcement} learn end-effector control of an excavator using a neural network model of its hydraulic actuation.
\Citet{yuryev2026tendon} learn a model of forces transferred via a tendon and use it for trajectory tracking with a one-degree-of-freedom motor-actuated finger.
\Citet{fey2025bridging} treat the actuator network optimization as an RL task, where the agent attempts to produce torques that minimize the error between simulated and real trajectories.
They use the method to model friction and hysteresis effects in a robot arm with harmonic drives and demonstrate successful sim-to-real transfer in dynamic whole-body control tasks.
In contrast to their work, our method enables gradient-based optimization, thereby simplifying training.
As a result, our method is capable of modeling complex muscle dynamics, which are highly nonlinear, affected by hysteresis, and subject to configuration-dependent friction along the tendons.

\subsection{Sim-to-real transfer for muscle-actuated systems}
Existing works on sim-to-real transfer with muscle actuation target relatively simple systems and are limited to reaching tasks.
\Citet{tao2025efficient} utilize a combination of system identification with an analytic dynamics model and DR to learn reaching policies for single-joint robotic systems.
\Citet{biyajima2025development} demonstrate a successful sim-to-real transfer with an analytic dynamics model for a one-degree-of-freedom percussion robot.
\Citet{wang2025dynamic} learn a Deep Lagrangian Network~\cite{lutter2019deep} model of PAM dynamics to train a goal-reaching policy for a single muscle with a weight attached.

\Citet{schumacher2024learning} approach muscle actuation from a different direction by emulating muscles in software on a motor-driven quadruped.
Thereby, they simplify the sim-to-real transfer as the muscle dynamics are known exactly.
However, since the muscles are only emulated, this approach sacrifices some advantages of muscle actuation, such as zero-delay compliance. 
\Citet{buchler2022learning} utilize simulations to learn dynamic table tennis policies with a complex muscle-actuated robot.
They circumvent the challenges of simulating the muscle and tendon dynamics by keeping the robot real and simulating only the ball.
Even though~\citet{guist2023hindsight} improve the sample efficiency by simulating multiple balls during each stroke, the training still requires many hours of robot interactions.

To advance sim-to-real learning for muscle-actuated robot arms toward practical robot applications, it is paramount to develop methods for more complex and capable robots. 
Systems with multiple joints introduce significant modeling challenges, such as mechanical coupling between the degrees of freedom (DoFs) and friction dependent on the robot configuration due to tendon routing.
This work tackles these challenges by leveraging the expressiveness of neural networks, thereby unlocking sim-to-real learning for complex muscle-actuated robots.
To the best of our knowledge, we demonstrate the first successful sim-to-real transfer for a 4-DoF muscle-actuated robot.

\section{Sim-to-real pipeline}
\label{sec:method}
\begin{figure}
    \centering
    \adjustbox{width=\linewidth}{\input{figs/overview}}
    \caption{
        Overview of the sim-to-real pipeline. 
        (1)~GenAN training with the differentiable position loss.
        The network is trained to produce torques so that the simulated joint positions match the exploration data.
        (2)~RL training in a parallelized simulation, where the trained actuator network converts the policy's control signals into torques for a torque-based simulator of the arm and external objects.
        (3)~Zero-shot transfer to the real system.
    }
    \label{fig:overview}
\end{figure}
This work presents a novel approach to simulating robots with complex actuator dynamics that leverages the known rigid-body dynamics to learn an actuator model, enabling sim-to-real transfer for these systems.
Our pipeline consists of three phases.
First, we collect a motion dataset with the real robot and train a GenAN to map robot states and control signals to resulting joint torques.
Then, we combine this model with a simulator of the rigid-body dynamics to train an RL policy entirely in simulation.
Lastly, we deploy this policy zero-shot on the real robot.
See \cref{fig:overview} for an overview.

\subsection{Actuator model training}
\label{subsec:actuator_network_training}
We first collect an exploration dataset of 2500 randomized open-loop trajectories, each two seconds in length, for a total of about $\SI{1.4}{\hour}$ of robot data, which we split into $\SI{80}{\percent}$ training data and $\SI{20}{\percent}$ validation data for early stopping.
See \cref{subapp:implementation_data_collection} for details on the exploration motions.
Using this data, we train a GenAN $\boldsymbol{\hat{\tau}}_t = f_{\boldsymbol{\theta}}(\boldsymbol{q}_{t-H:t}, \boldsymbol{u}_{t-H:t})$ that maps from control signals to resulting joint torques $\boldsymbol{\hat{\tau}}_t$.
Similar to~\cite{hwangbo2019learning}, we pass an $H$-step history of joint positions $\boldsymbol{q}_{t-H:t}$ and control signals $\boldsymbol{u}_{t-H:t}$ to the network to model hysteresis effects.
We use a simulator of the arm dynamics with step function $\boldsymbol{q}_{t+1} = \step(\qreal{t}, \qdreal{t}, \tor{t})$ and inverse dynamics function $\boldsymbol{\tau}_t = \invdyn(\qreal{t}, \qdreal{t}, \qddreal{t})$.
Note that this simulator is torque-driven since we do not have a precise analytic actuator model.

\Citet{hwangbo2019learning} use sparse histories $(\boldsymbol{x}_{t}, \boldsymbol{x}_{t-s}, \ldots, \boldsymbol{x}_{t-s \cdot H})$ with stride length $s=4$, mentioning overfitting as a problem of dense histories.
We hypothesize that overfitting occurs since consecutive values are very similar, making it difficult for the network to make proper use of such histories.
To mitigate this issue, we replace the histories $(\boldsymbol{x}_{t}, \boldsymbol{x}_{t-1}, \ldots, \boldsymbol{x}_{t-H})$ in the network input with \emph{delta histories}, i.e., we represent the same sequences by the differences to the current value $(\boldsymbol{x}_{t}, \boldsymbol{x}_{t-1} - \boldsymbol{x}_{t}, \ldots, \boldsymbol{x}_{t-H} - \boldsymbol{x}_{t})$ and standardize all inputs, which amplifies the differences between consecutive values.
\Cref{app:history} indeed shows that short strides work best in our setting.
For the remainder of the paper, we denote the delta histories simply as $\boldsymbol{q}_{t-H:t}$ and $\boldsymbol{u}_{t-H:t}$ for conciseness.

Since we do not have access to torque measurements, we use the joint position data as feedback for learning the actuator model.
Concretely, we sample sequences of joint positions and control signals $\boldsymbol{q}_{t-H:t+1}, \boldsymbol{u}_{t-H:t}$ from the dataset.
On this data, we train the model to produce torques $\torpred{t} = f_{\boldsymbol{\theta}}\mleft(\boldsymbol{q}_{t-H:t}, \boldsymbol{u}_{t-H:t}\mright)$ that minimize the squared distance between the simulated next positions $\qpred{t+1} = \step(\boldsymbol{q}_{t}, \boldsymbol{\dot{q}}_{t}, \boldsymbol{\hat{\tau}}_t)$ and the true next positions from the dataset $\qreal{t+1}$, resulting in the following \emph{position loss}.
\begin{equation}
    \mathcal{L}(\boldsymbol{\theta}) = \left\lVert\qpred{t+1} - \qreal{t+1}\right\rVert^2
    \label{eq:position_loss}
\end{equation}
Differentiating through the simulator enables efficient and stable gradient-based optimization.
As a result, the GenAN training requires relatively little hyperparameter tuning.
Refer to \cref{app:implementation_details_genan} for training details and the hyperparameter values that we use.

In \cref{app:relation_torque_position_loss}, we derive that the position error is related to the torque error in the following way.
\begin{equation}
    \qreal{t+1} - \qpred{t+1} = \Delta t^2 \boldsymbol{M}(\qreal{t})^{-1} (\tor{t} - \torpred{t})
    \label{eq:relation_torque_position_loss}
\end{equation}
$\tor{t} = \invdyn(\qreal{t}, \qdreal{t}, \qddreal{t})$ is the torque computed via inverse dynamics, $\Delta t$ is the simulator time step, and $\boldsymbol{M}(\qreal{t})$ is the mass matrix of the robot in position $\qreal{t}$.
The joint velocities $\qdreal{t}$ and accelerations $\qddreal{t}$ are computed from the position trajectories via finite differences.
This equivalence enables computing gradients of \cref{eq:position_loss} even for simulators that are not natively differentiable.

We also experimented with a multi-step position loss, but we found that it does not yield consistent gains over the single-step variant.
Refer to \cref{app:multi_step_loss} for details on the multi-step loss.

\subsection{Simulated RL environment}
\label{subsec:policy_training}

To obtain a realistic simulation environment for training policies, we deploy the GenAN with a torque-based simulator of the arm dynamics and the objects in the scene.
The arm and objects follow simple rigid body dynamics, which can be simulated accurately with analytic models.
Therefore, we use the learned model only for the complex and hard-to-model muscle and tendon dynamics.
In contrast to learning the entire simulator end-to-end, this scheme allows leveraging prior knowledge about rigid body dynamics and enables changing the scene, e.g., by adding or removing objects, without retraining the network.
Instead of a single network, we use an ensemble of five GenANs, each trained according to \cref{subsec:actuator_network_training}.
Each model in the ensemble is initialized with a different random seed and trained on a different dataset permutation.
The ensemble disagreement constitutes a measure of the model's epistemic uncertainty.
Similar to~\cite{janner2019trust}, we sample a random network from the GenAN ensemble for each simulation step to mitigate policy overfitting to model uncertainty.

\section{Evaluation on a muscle-actuated robot}
\label{sec:experiments}
\columnratio{0.54,0.46}
\begin{paracol}{2}

\switchcolumn[0]

We evaluate on PAMY2~\cite{guist2024safe}, a 4-DoF, PAM-actuated, tendon-driven robot arm.
Each DoF is actuated by an antagonistic muscle pair.
Similar to~\cite{buchler2023learning}, we actuate both muscles in a correlated way with a single control signal $\ctrl{}$.
Increasing $\ctrl{}$ decreases the agonist pressure and simultaneously increases the antagonist pressure.
\Cref{fig:robot} shows the robot and its virtual counterpart simulated in MuJoCo XLA~(MJX)~\cite{todorov2012mujoco}, an efficient GPU-based simulator.

\switchcolumn
\vspace{3.5pt}
\nolinenumbers
\includegraphics[width=0.496\linewidth]{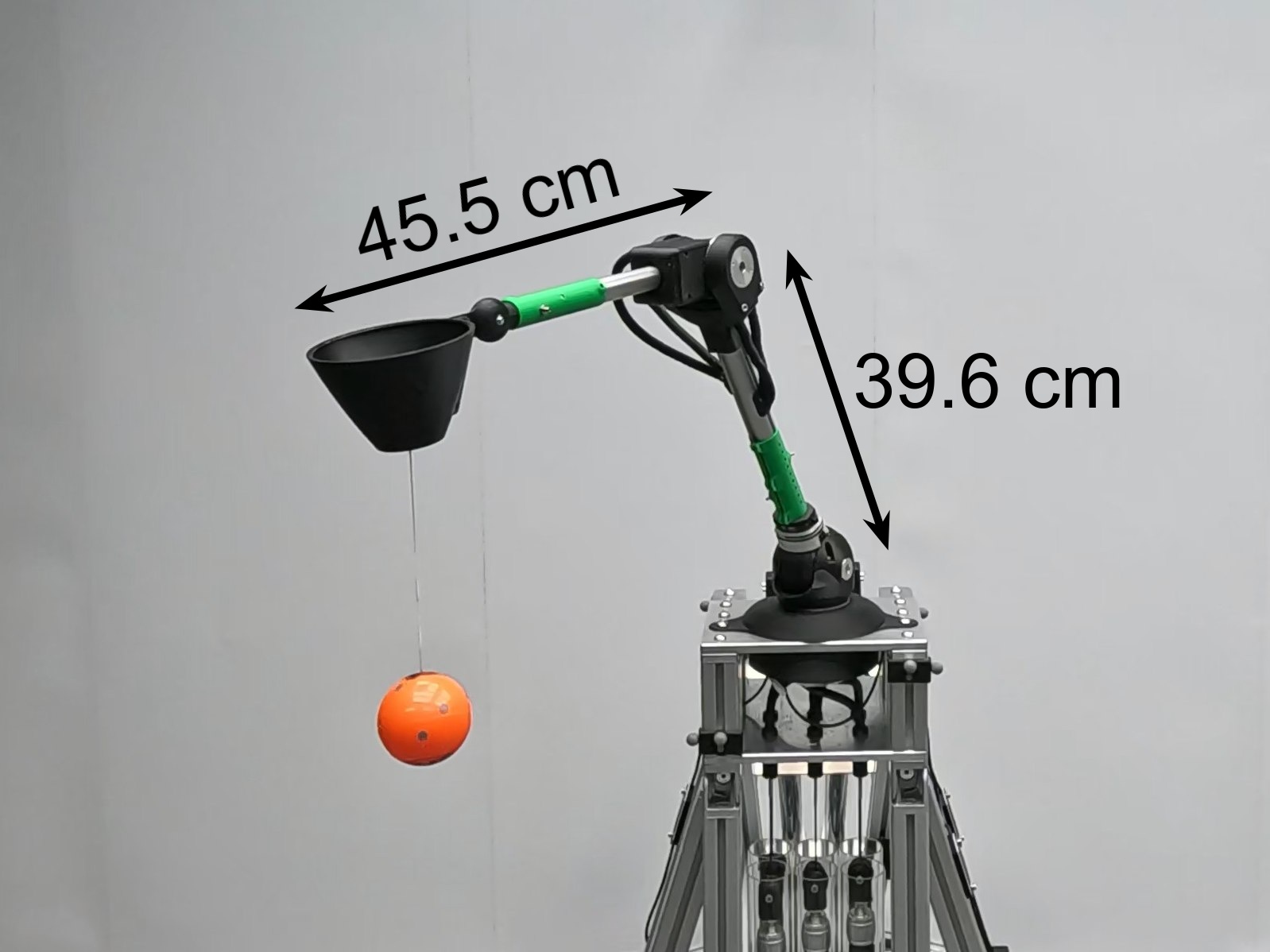}\hfill
\includegraphics[width=0.496\linewidth]{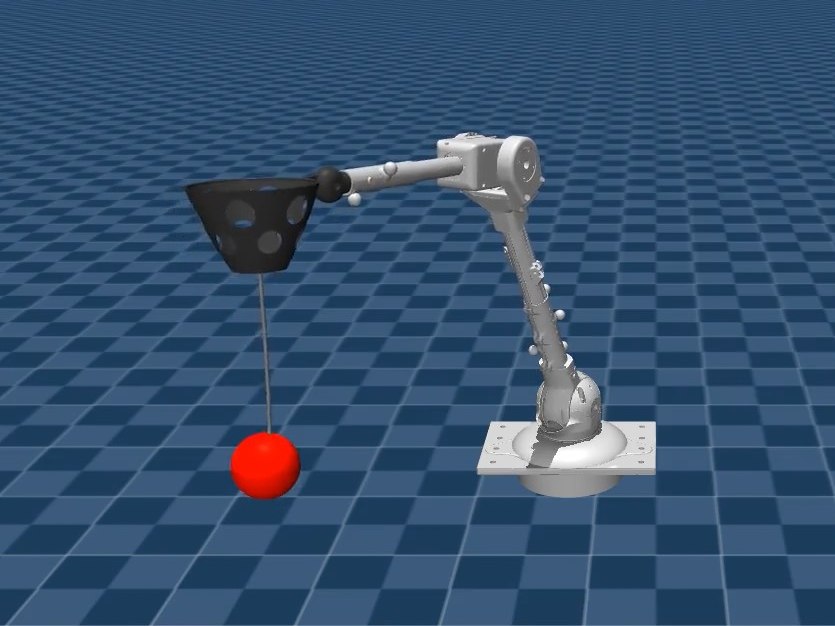}
\captionof{figure}{
    The 4-DoF PAM-actuated PAMY2 (left) with its simulated counterpart (right).
}
\label{fig:robot}
\linenumbers

\end{paracol}

\stepcounter{figure}

\subsection{Actuator model baselines}
We compare the accuracy of GenAN to three actuator model baselines.
\emph{Torque GenAN} matches the architecture of GenAN but is trained with a torque loss instead of the position loss introduced in \cref{eq:position_loss}.
\emph{Unsupervised Actuator Net (UAN)}~\cite{fey2025bridging} is the only existing actuator network technique that does not require torque sensors.
Lastly, we compare to a System Identification (SysID) baseline that constructs an analytical model based on the Chou-Hannaford PAM model~\cite{chou1996measurement} and an explicit tendon friction model.
We use the dataset collected for the GenAN to identify 37 physical model parameters.
See \cref{app:sysid} for details on the model and the SysID procedure.

\subsubsection{Torque GenAN}

For the Torque GenAN baseline, we compute torque labels from the dataset trajectories via inverse dynamics $\boldsymbol{\tau}_t = \invdyn(\boldsymbol{q}_t, \boldsymbol{\dot{q}}_t, \boldsymbol{\ddot{q}}_t)$.
To deal with differences in torque magnitudes across joints, we standardize the torque labels $\boldsymbol{\tau}^{\mathrm{std}}_t = (\boldsymbol{\tau}_t - \boldsymbol{\mu}) / \boldsymbol{\sigma}$, where $\boldsymbol{\mu}$ and $\boldsymbol{\sigma}$ are the torque mean and standard deviation across the training set and the multiplication and division are elementwise.
We train the network with the following loss.
\begin{equation}
    \mathcal{L}_{\mathrm{tor}}(\boldsymbol{\theta}) = \left\lVert f^{\mathrm{std}}_{\boldsymbol{\theta}}(\boldsymbol{q}_{t-H:t}, \boldsymbol{u}_{t-H:t}) - \boldsymbol{\tau}^\mathrm{std}_t \right\rVert^2 \label{eq:torque_loss}
\end{equation}
For inference, we invert the standardization $f_{\boldsymbol{\theta}}(\boldsymbol{q}_{t-H:t}, \boldsymbol{u}_{t-H:t}) = f_{\boldsymbol{\theta}}^{\mathrm{std}}(\boldsymbol{q}_{t-H:t}, \boldsymbol{u}_{t-H:t}) \boldsymbol{\sigma} + \boldsymbol{\mu}$.

\subsubsection{Unsupervised Actuator Net}
\label{subsubsec:uan}
UAN treats the actuator network optimization as an RL task, where the UAN is the policy that gets histories of joint positions and control signals as observations.
Each episode corresponds to one dataset trajectory, and the policy is trained to output torques that reproduce the trajectory in the simulator.
The reward is based on the negative difference to the dataset trajectory, plus a smoothness term.
Since the authors assume a motor-driven system controlled by a PD-controller with known gains, the control signals are desired positions, and the torques commanded by the controller are known.
The agent then learns residuals between the commanded and true torques, caused by friction losses in the gears.
Since there are no torque commands for muscle-actuated robots, we change the method to predict the full joint torques instead.
To ensure a fair comparison, we ran a hyperparameter search for UAN, sweeping over 200 sets of RL hyperparameters, history lengths, network architectures, and reward weights.
Refer to \cref{subapp:implementation_details_uan} for the resulting hyperparameters.

\subsection{Simulation accuracy}
\label{subsec:actuator_net_accuracy}

To test whether the models faithfully reproduce PAMY2's complex actuation dynamics, we collected a test set of 800 robot trajectories with the exploration policy (\cref{subapp:implementation_data_collection}).
For each trajectory, we initialize the simulator to the same state, replay the recorded command sequence, and measure the deviation from the real joint positions.
We assess the accuracy of the simulation for a single step and the robustness to error accumulation during multi-step rollouts, where the network inputs depend on previous predictions.
\Cref{fig:accuracy_torque_vs_position} compares the position error for GenAN to the three baselines.
For both single-step and multi-step predictions, GenAN produces the most accurate trajectories, yielding $\SI{6}{\percent}$ and $\SI{32}{\percent}$ lower errors, respectively, than Torque GenAN.
The reason for the performance differences between GenAN and this baseline can be explained by \cref{eq:relation_torque_position_loss}.
While the equation shows that an ideal actuator network with zero torque error would result in zero position error, it also reveals that errors from suboptimal torque predictions are multiplied by the inverse of the mass matrix.
Since the mass matrix can have significant off-diagonal elements, torque errors of equal magnitude can induce substantially different position errors depending on their direction and may either reinforce or cancel one another.
Unlike the torque loss, the position loss directly accounts for this effect and scores torque predictions according to their impact on the resulting motion.

\begin{figure}[ht]
    \begin{subfigure}[t]{0.495\linewidth}
        \centering
\begin{tikzpicture}

\definecolor{darkgray141160203}{RGB}{141,160,203}
\definecolor{darkgray176}{RGB}{176,176,176}
\definecolor{dimgray}{RGB}{105,105,105}
\definecolor{gray}{RGB}{128,128,128}
\definecolor{gray16196136}{RGB}{161,96,136}
\definecolor{mediumaquamarine102194165}{RGB}{102,194,165}
\definecolor{orchid231138195}{RGB}{231,138,195}
\definecolor{salmon25214198}{RGB}{252,141,98}
\definecolor{seagreen71135115}{RGB}{71,135,115}
\definecolor{sienna1769868}{RGB}{176,98,68}
\definecolor{slategray98111142}{RGB}{98,111,142}
\definecolor{yellow}{RGB}{255,255,0}

\begin{axis}[
golden width landscape=0.75\linewidth,scale only axis,tick align=outside,
tick pos=left,
x grid style={darkgray176},
xmin=-0.8, xmax=3.7,
xtick style={color=black,draw=none},
xtick={0,1,2,3},
xticklabel style={align=center},
xticklabels={GenAN,Torque\\GenAN,UAN,SysID},
y grid style={darkgray176},
ylabel={position error (°)},
ymajorgrids,
ymin=0, ymax=0.0357139953073805,
ytick style={color=black}
]
\draw[draw=seagreen71135115,fill=mediumaquamarine102194165,very thick] (axis cs:-0.35,0) rectangle (axis cs:0.35,0.00676372656021279);
\draw[draw=sienna1769868,fill=salmon25214198,very thick] (axis cs:0.65,0) rectangle (axis cs:1.35,0.00721852944703384);
\draw[draw=slategray98111142,fill=darkgray141160203,very thick] (axis cs:1.65,0) rectangle (axis cs:2.35,0.0289861555688861);
\draw[draw=gray16196136,fill=orchid231138195,very thick] (axis cs:2.65,0) rectangle (axis cs:3.35,0.0327971943524494);
\path [draw=dimgray, very thick]
(axis cs:0,0.0064799745263861)
--(axis cs:0,0.00707070156742481);

\path [draw=dimgray, very thick]
(axis cs:1,0.00692862806980977)
--(axis cs:1,0.007524725104891);

\path [draw=dimgray, very thick]
(axis cs:2,0.0276675635286874)
--(axis cs:2,0.0303469181171918);

\path [draw=dimgray, very thick]
(axis cs:3,0.0316353903038989)
--(axis cs:3,0.0340133288641719);

\addplot [semithick, dimgray, mark=-, mark size=3, mark options={solid,draw=dimgray}, only marks]
table {%
0 0.0064799745263861
1 0.00692862806980977
2 0.0276675635286874
3 0.0316353903038989
};
\addplot [semithick, dimgray, mark=-, mark size=3, mark options={solid,draw=dimgray}, only marks]
table {%
0 0.00707070156742481
1 0.007524725104891
2 0.0303469181171918
3 0.0340133288641719
};
\draw[gray, dashed, thick] (axis cs:-0.45,0) rectangle (axis cs:1.45,0.008277197615380096);
\coordinate (yellowtopleft) at (axis cs:-0.45,0.008277197615380096);
\coordinate (yellowtopright) at (axis cs:1.45,0.008277197615380096);
\end{axis}

\fill[white] (\insetxpos,\insetypos) rectangle (\insetxpos+\insetwidth,\insetypos+\insetheight);
\begin{axis}[
width=\insetwidth,height=\insetheight,scale only axis,at={(\insetxpos,\insetypos)},anchor=south west,axis line style={gray,dotted,thick},xtick=\empty,xticklabels=\empty,
tick align=outside,
tick pos=left,
xmin=-0.6, xmax=1.6,
y grid style={gray,dotted},
ymajorgrids,scaled y ticks=false,yticklabel={\pgfmathparse{\tick*100}\pgfmathprintnumber[fixed,precision=1]{\pgfmathresult}},
ymin=0, ymax=0.0078,
ytick style={color=black}
]
\draw[draw=seagreen71135115,fill=mediumaquamarine102194165,thick] (axis cs:-0.3,0) rectangle (axis cs:0.3,0.00676372656021279);
\draw[draw=sienna1769868,fill=salmon25214198,thick] (axis cs:0.7,0) rectangle (axis cs:1.3,0.00721852944703384);
\path [draw=dimgray, thick]
(axis cs:0,0.0064799745263861)
--(axis cs:0,0.00707070156742481);

\path [draw=dimgray, thick]
(axis cs:1,0.00692862806980977)
--(axis cs:1,0.007524725104891);

\addplot [semithick, dimgray, mark=-, mark size=4, mark options={solid,draw=dimgray}, only marks]
table {%
0 0.0064799745263861
1 0.00692862806980977
};
\addplot [semithick, dimgray, mark=-, mark size=4, mark options={solid,draw=dimgray}, only marks]
table {%
0 0.00707070156742481
1 0.007524725104891
};
\end{axis}

\draw[gray, dotted, thick] (\insetxpos,\insetypos) rectangle (\insetxpos+\insetwidth,\insetypos+\insetheight);
\draw[gray, dotted, thick] (yellowtopleft) -- (\insetxpos,\insetypos);
\draw[gray, dotted, thick] (yellowtopright) -- (\insetxpos+\insetwidth,\insetypos);
\end{tikzpicture}
        \caption{Error after 1 step / $\SI{2}{\ms}$}
        \label{subfig:accuracy_torque_vs_position_1step}
    \end{subfigure}
    \begin{subfigure}[t]{0.495\linewidth}
        \centering
\begin{tikzpicture}

\definecolor{darkgray141160203}{RGB}{141,160,203}
\definecolor{darkgray176}{RGB}{176,176,176}
\definecolor{dimgray}{RGB}{105,105,105}
\definecolor{gray}{RGB}{128,128,128}
\definecolor{gray16196136}{RGB}{161,96,136}
\definecolor{mediumaquamarine102194165}{RGB}{102,194,165}
\definecolor{orchid231138195}{RGB}{231,138,195}
\definecolor{salmon25214198}{RGB}{252,141,98}
\definecolor{seagreen71135115}{RGB}{71,135,115}
\definecolor{sienna1769868}{RGB}{176,98,68}
\definecolor{slategray98111142}{RGB}{98,111,142}
\definecolor{yellow}{RGB}{255,255,0}

\begin{axis}[
golden width landscape=0.75\linewidth,scale only axis,tick align=outside,
tick pos=left,
x grid style={darkgray176},
xmin=-0.8, xmax=3.7,
xtick style={color=black,draw=none},
xtick={0,1,2,3},
xticklabel style={align=center},
xticklabels={GenAN,Torque\\GenAN,UAN,SysID},
y grid style={darkgray176},
ylabel={},
ymajorgrids,
ymin=0, ymax=32.4634305110225,
ytick style={color=black}
]
\draw[draw=seagreen71135115,fill=mediumaquamarine102194165,very thick] (axis cs:-0.35,0) rectangle (axis cs:0.35,4.4454567958345);
\draw[draw=sienna1769868,fill=salmon25214198,very thick] (axis cs:0.65,0) rectangle (axis cs:1.35,6.56454731838052);
\draw[draw=slategray98111142,fill=darkgray141160203,very thick] (axis cs:1.65,0) rectangle (axis cs:2.35,29.3850789038589);
\draw[draw=gray16196136,fill=orchid231138195,very thick] (axis cs:2.65,0) rectangle (axis cs:3.35,11.2415129492459);
(axis cs:0,4.30914694832618)
--(axis cs:0,4.591875440919);

\path [draw=dimgray, very thick]
(axis cs:1,5.96143381923372)
--(axis cs:1,7.39838189973941);

\path [draw=dimgray, very thick]
(axis cs:2,27.9466444233771)
--(axis cs:2,30.9175528676405);

\path [draw=dimgray, very thick]
(axis cs:3,10.8745905977596)
--(axis cs:3,11.6259095630069);

\addplot [semithick, dimgray, mark=-, mark size=3, mark options={solid,draw=dimgray}, only marks]
table {%
0 4.30914694832618
1 5.96143381923372
2 27.9466444233771
3 10.8745905977596
};
\addplot [semithick, dimgray, mark=-, mark size=3, mark options={solid,draw=dimgray}, only marks]
table {%
0 4.591875440919
1 7.39838189973941
2 30.9175528676405
3 11.6259095630069
};
\draw[gray, dashed, thick] (axis cs:-0.45,0) rectangle (axis cs:1.45,8.138220089713347);
\coordinate (yellowtopleft) at (axis cs:-0.45,8.138220089713347);
\coordinate (yellowtopright) at (axis cs:1.45,8.138220089713347);
\end{axis}

\fill[white] (\insetxpos,\insetypos) rectangle (\insetxpos+\insetwidth,\insetypos+\insetheight);
\begin{axis}[
width=\insetwidth,height=\insetheight,scale only axis,at={(\insetxpos,\insetypos)},anchor=south west,axis line style={gray,dotted,thick},xtick=\empty,xticklabels=\empty,
tick align=outside,
tick pos=left,
xmin=-0.6, xmax=1.6,
y grid style={gray,dotted},
ymajorgrids,scaled y ticks=false,yticklabel style={/pgf/number format/fixed,/pgf/number format/precision=1},
ymin=0, ymax=7.8,
ytick style={color=black}
]
\draw[draw=seagreen71135115,fill=mediumaquamarine102194165,thick] (axis cs:-0.3,0) rectangle (axis cs:0.3,4.4454567958345);
\draw[draw=sienna1769868,fill=salmon25214198,thick] (axis cs:0.7,0) rectangle (axis cs:1.3,6.56454731838052);
\path [draw=dimgray, thick]
(axis cs:0,4.30914694832618)
--(axis cs:0,4.591875440919);

\path [draw=dimgray, thick]
(axis cs:1,5.96143381923372)
--(axis cs:1,7.39838189973941);

\addplot [semithick, dimgray, mark=-, mark size=4, mark options={solid,draw=dimgray}, only marks]
table {%
0 4.30914694832618
1 5.96143381923372
};
\addplot [semithick, dimgray, mark=-, mark size=4, mark options={solid,draw=dimgray}, only marks]
table {%
0 4.591875440919
1 7.39838189973941
};
\end{axis}

\draw[gray, dotted, thick] (\insetxpos,\insetypos) rectangle (\insetxpos+\insetwidth,\insetypos+\insetheight);
\draw[gray, dotted, thick] (yellowtopleft) -- (\insetxpos,\insetypos);
\draw[gray, dotted, thick] (yellowtopright) -- (\insetxpos+\insetwidth,\insetypos);
\end{tikzpicture}
        \caption{Error after 500 steps / $\SI{1}{\s}$}
        \label{subfig:accuracy_torque_vs_position_500steps}
    \end{subfigure}
    \caption{
        Simulation accuracy of the GenAN in comparison to the Torque GenAN baseline, UAN~\cite{fey2025bridging}, and a SysID baseline.
        For 800 real test trajectories, we reset the simulator to the same initial configuration and apply the same command sequence.
        The plots show the mean absolute error between simulated and real trajectories after (a) a single step (2 milliseconds) and (b) 500 steps (1 second).
        The error bars visualize the 95\% confidence intervals, obtained via bootstrapping.
        GenAN yields the most accurate simulation, vastly outperforming UAN and the SysID baseline.
    }
    \label{fig:accuracy_torque_vs_position}
\end{figure}

UAN fails to capture the dynamics, resulting in substantially higher errors.
We believe that UAN is unable to model PAMY2's actuator dynamics accurately since the RL agent gets incomplete information in this setting.
As the agent observes only the controls and the positions resulting from its previous predictions, but not the true positions, it cannot reliably infer rewards or values, making the task partially observable.
Note that we cannot provide the true positions, as they are not available during deployment.
In the UAN paper~\cite{fey2025bridging}, the controls are desired positions for a PD-controller.
With an accurate controller, the true positions are typically close to the targets, making the desired positions a decent proxy for the true positions and mitigating the partial-observability issue.

The SysID baseline is also substantially less accurate than GenAN, particularly in the one-step predictions.
Even with 37 parameters identified from data, the analytical model is not expressive enough to capture the complex muscle and tendon dynamics faithfully.

\subsection{Task descriptions}
\label{subsec:rl_task}
To validate the efficacy of the GenAN-augmented simulator for training policies for PAMY2, we learn three tasks: \emph{reacher}, \emph{ball-in-a-cup}~\cite{kober2008policy}, and \emph{table tennis}~\cite{guist2024safe,buchler2022learning}.
The reacher task involves only the robot and, therefore, directly evaluates the precision of the learned motions independent of external error sources, such as object tracking noise.
Conversely, the ball-in-a-cup and table tennis tasks include interactions with external objects and require highly dynamic yet precise motions.
We train policies using the skrl~\cite{serrano2023skrl} implementation of Proximal Policy Optimization~(PPO)~\cite{schulman2017proximal} in a GPU-accelerated, massively parallelized simulation.

The observations $\boldsymbol{o}_t = (\boldsymbol{q}_t,\boldsymbol{\dot{q}}_t, \boldsymbol{u}_{t-1}, \boldsymbol{o}_t^\text{task})$ include the current position $\boldsymbol{q}_t$ and velocity $\boldsymbol{\dot{q}}_t$, the last command $\boldsymbol{u}_{t-1}$, and task-specific observations $\boldsymbol{o}_t^\mathrm{task}$.
The actions are changes to the control signal $\boldsymbol{a}_t = \Delta \boldsymbol{u}_t$ and the reward $r_t = r^\mathrm{task}_t + c^\mathrm{act} r^\mathrm{act}_t + c^\mathrm{disag} r^\mathrm{disag}_t + c^\mathrm{lim} r^\mathrm{lim}_t$ includes the main task reward $r^\mathrm{task}_t$, an action penalty $r^\mathrm{act}_t = -\lVert\boldsymbol{a}_t\rVert^2$, and a disagreement penalty 
$r^\mathrm{disag}_t = -\sum_{j=1}^4 \boldsymbol{\sigma}^\tau_j$ with $\boldsymbol{\sigma}^\tau = \Sd_i\left[f_{\boldsymbol{\theta}_i}(\boldsymbol{q}_{t-H:t}, \boldsymbol{u}_{t-H:t})\right]$ that penalizes uncertainty in the GenAN ensemble, measured as the standard deviation across the outputs of the ensemble members.
$r^\mathrm{lim}_t$ applies a penalty if the agent gets too close to the joint limits.
We use the same weighting constants across all tasks.

In the \emph{reacher} task, the robot is initialized with a random control signal and has to reach uniformly sampled target joint positions $\boldsymbol{g}$.
For \emph{ball-in-a-cup}, the robot has to swing a ball on a string into a cup at its end effector.
For the \emph{table tennis} task, we use a similar setup as in \cite{guist2024safe,buchler2022learning}, where the AIMY ball launcher~\cite{dittrich2023aimy} launches a ball that the robot has to return to the opponent's side of the table.
Refer to \cref{fig:ball_in_a_cup_rollout} for visualizations of the tasks and to \cref{subapp:implementation_rl_tasks} for detailed task descriptions, including the task-specific observations and rewards.

\subsection{Transfer to the real robot}
\label{subsec:transfer}

After training entirely in simulation, we deploy the policies on the real robot.
We consider a reacher episode a success if the average distance to the goal is less than \ang{2} at the end of the episode.
For ball-in-a-cup, we define success as catching the ball, and for table tennis, success means returning the ball to the opponent's side of the table.
\Cref{fig:ball_in_a_cup_rollout} shows successful ball-in-a-cup and table tennis rollouts.
We compare the performance of policies trained with GenAN to four baselines: training with the Torque GenAN, removing the penalty on the GenAN disagreement, replacing the GenAN ensemble with a single GenAN, and reducing the action penalty weight from $c^\mathrm{act}=1250$ to $250$.

\begin{figure}[ht]
    \begin{tikzpicture}

\definecolor{darkgray141160203}{RGB}{141,160,203}
\definecolor{darkgray176}{RGB}{176,176,176}
\definecolor{dimgray}{RGB}{105,105,105}
\definecolor{gray16196136}{RGB}{161,96,136}
\definecolor{lightgray204}{RGB}{204,204,204}
\definecolor{mediumaquamarine102194165}{RGB}{102,194,165}
\definecolor{olivedrab11615158}{RGB}{116,151,58}
\definecolor{orchid231138195}{RGB}{231,138,195}
\definecolor{salmon25214198}{RGB}{252,141,98}
\definecolor{seagreen71135115}{RGB}{71,135,115}
\definecolor{sienna1769868}{RGB}{176,98,68}
\definecolor{slategray98111142}{RGB}{98,111,142}
\definecolor{yellowgreen16621684}{RGB}{166,216,84}

\begin{axis}[
  axis lines=left,
  axis line style={-},
  width=0.8\linewidth,height=0.37\linewidth,
  legend columns=1,
  legend cell align={left},
  legend style={
  fill opacity=0.8,
  draw opacity=1,
  text opacity=1,
  at={(1.01,0.5)},
  anchor=west,
  draw=none,
  inner sep=0pt,
  outer sep=0pt,
  font=\footnotesize
},
tick align=outside,
tick pos=left,
x grid style={darkgray176},
xmin=0.545, xmax=16.055,
xtick style={color=black,draw=none},
xtick={1.6,2.45,3.3,4.15,5,6.6,7.45,8.3,9.15,10,11.6,12.45,13.3,14.15,15},
xticklabels={,,reacher,,,,,ball-in-a-cup,,,,,table tennis,,},
xticklabel style={color=black,yshift=0pt,inner sep=0pt,text depth=0pt},
y grid style={darkgray176},
ylabel={success rate (\%)},
ylabel style={yshift=-1pt,inner sep=0pt,text depth=0pt},
ytick style={color=black,draw=none},
yticklabel style={color=black,xshift=4pt},
ymajorgrids,
ymin=0, ymax=100.2,
]
\draw[draw=seagreen71135115,fill=mediumaquamarine102194165,very thick] (axis cs:1.25,0) rectangle (axis cs:1.95,90);
\draw[draw=sienna1769868,fill=salmon25214198,very thick] (axis cs:2.1,0) rectangle (axis cs:2.8,97);
\draw[draw=slategray98111142,fill=darkgray141160203,very thick] (axis cs:2.95,0) rectangle (axis cs:3.65,86);
\draw[draw=gray16196136,fill=orchid231138195,very thick] (axis cs:3.8,0) rectangle (axis cs:4.5,93);
\draw[draw=olivedrab11615158,fill=yellowgreen16621684,very thick] (axis cs:4.65,0) rectangle (axis cs:5.35,70);
\draw[draw=seagreen71135115,fill=mediumaquamarine102194165,very thick] (axis cs:6.25,0) rectangle (axis cs:6.95,75);
\draw[draw=sienna1769868,fill=salmon25214198,very thick] (axis cs:7.1,0) rectangle (axis cs:7.8,76);
\draw[draw=slategray98111142,fill=darkgray141160203,very thick] (axis cs:7.95,0) rectangle (axis cs:8.65,56);
\draw[draw=gray16196136,fill=orchid231138195,very thick] (axis cs:8.8,0) rectangle (axis cs:9.5,74);
\draw[draw=olivedrab11615158,fill=yellowgreen16621684,very thick] (axis cs:9.65,0) rectangle (axis cs:10.35,55);
\draw[draw=seagreen71135115,fill=mediumaquamarine102194165,very thick] (axis cs:11.25,0) rectangle (axis cs:11.95,96);
\draw[draw=sienna1769868,fill=salmon25214198,very thick] (axis cs:12.1,0) rectangle (axis cs:12.8,75);
\draw[draw=slategray98111142,fill=darkgray141160203,very thick] (axis cs:12.95,0) rectangle (axis cs:13.65,82);
\draw[draw=gray16196136,fill=orchid231138195,very thick] (axis cs:13.8,0) rectangle (axis cs:14.5,83);
\draw[draw=olivedrab11615158,fill=yellowgreen16621684,very thick] (axis cs:14.65,0) rectangle (axis cs:15.35,54);
\path [draw=dimgray, very thick]
(axis cs:1.6,82.5634338495086)
--(axis cs:1.6,94.4770862939325);

\path [draw=dimgray, very thick]
(axis cs:2.45,91.5480635709472)
--(axis cs:2.45,98.9745475975961);

\path [draw=dimgray, very thick]
(axis cs:3.3,77.8628117890362)
--(axis cs:3.3,91.4736563400609);

\path [draw=dimgray, very thick]
(axis cs:4.15,86.2504852609265)
--(axis cs:4.15,96.5680738932727);

\path [draw=dimgray, very thick]
(axis cs:5,60.4151453666533)
--(axis cs:5,78.1051147050672);

\path [draw=dimgray, very thick]
(axis cs:6.6,65.6955364519384)
--(axis cs:6.6,82.4547886377123);

\path [draw=dimgray, very thick]
(axis cs:7.45,66.7676636501855)
--(axis cs:7.45,83.3086744430512);

\path [draw=dimgray, very thick]
(axis cs:8.3,46.228104651677)
--(axis cs:8.3,65.3279733698392);

\path [draw=dimgray, very thick]
(axis cs:9.15,64.6290105508128)
--(axis cs:9.15,81.5953015352519);

\path [draw=dimgray, very thick]
(axis cs:10,45.2446029974421)
--(axis cs:10,64.385462020488);

\path [draw=dimgray, very thick]
(axis cs:11.6,90.1629285641121)
--(axis cs:11.6,98.4336696008452);

\path [draw=dimgray, very thick]
(axis cs:12.45,65.6955364519384)
--(axis cs:12.45,82.4547886377123);

\path [draw=dimgray, very thick]
(axis cs:13.3,73.3326416136993)
--(axis cs:13.3,88.2997745010536);

\path [draw=dimgray, very thick]
(axis cs:14.15,74.451995)
--(axis cs:14.15,89.106434);

\path [draw=dimgray, very thick]
(axis cs:15,44.26486032)
--(axis cs:15,63.43919169);

\addlegendimage{area legend, fill=mediumaquamarine102194165, draw=seagreen71135115, very thick}
\addlegendentry{GenAN}

\addlegendimage{area legend, fill=salmon25214198, draw=sienna1769868, very thick}
\addlegendentry{Torque GenAN}

\addlegendimage{area legend, fill=darkgray141160203, draw=slategray98111142, very thick}
\addlegendentry{No disag. penalty}

\addlegendimage{area legend,fill=orchid231138195,draw=gray16196136,very thick}
\addlegendentry{No ensemble}

\addlegendimage{area legend,fill=yellowgreen16621684,draw=olivedrab11615158,very thick}
\addlegendentry{Low action penalty}

\addplot [semithick, dimgray, mark=-, mark size=5, mark options={solid}, only marks]
table {%
1.6 82.5634338495086
2.45 91.5480635709472
3.3 77.8628117890362
4.15 86.2504852609265
5 60.4151453666533
6.6 65.6955364519384
7.45 66.7676636501855
8.3 46.228104651677
9.15 64.6290105508128
10 45.2446029974421
11.6 90.1629285641121
12.45 65.6955364519384
13.3 73.3326416136993
14.15 74.451995
15 44.26486032
};
\addplot [semithick, dimgray, mark=-, mark size=5, mark options={solid}, only marks]
table {%
1.6 94.4770862939325
2.45 98.9745475975961
3.3 91.4736563400609
4.15 96.5680738932727
5 78.1051147050672
6.6 82.4547886377123
7.45 83.3086744430512
8.3 65.3279733698392
9.15 81.5953015352519
10 64.385462020488
11.6 98.4336696008452
12.45 82.4547886377123
13.3 88.2997745010536
14.15 89.106434
15 63.43919169
};
\draw (axis cs:1.6,6) node[
  scale=0.9,
  text=white,
  rotate=0.0
]{\bfseries 90};
\draw (axis cs:2.45,6) node[
  scale=0.9,
  text=white,
  rotate=0.0
]{\bfseries 97};
\draw (axis cs:3.3,6) node[
  scale=0.9,
  text=white,
  rotate=0.0
]{\bfseries 86};
\draw (axis cs:4.15,6) node[
  scale=0.9,
  text=white,
  rotate=0.0
]{\bfseries 93};
\draw (axis cs:5,6) node[
  scale=0.9,
  text=white,
  rotate=0.0
]{\bfseries 70};
\draw (axis cs:6.6,6) node[
  scale=0.9,
  text=white,
  rotate=0.0
]{\bfseries 75};
\draw (axis cs:7.45,6) node[
  scale=0.9,
  text=white,
  rotate=0.0
]{\bfseries 76};
\draw (axis cs:8.3,6) node[
  scale=0.9,
  text=white,
  rotate=0.0
]{\bfseries 56};
\draw (axis cs:9.15,6) node[
  scale=0.9,
  text=white,
  rotate=0.0
]{\bfseries 74};
\draw (axis cs:10,6) node[
  scale=0.9,
  text=white,
  rotate=0.0
]{\bfseries 55};
\draw (axis cs:11.6,6) node[
  scale=0.9,
  text=white,
  rotate=0.0
]{\bfseries 96};
\draw (axis cs:12.45,6) node[
  scale=0.9,
  text=white,
  rotate=0.0
]{\bfseries 75};
\draw (axis cs:13.3,6) node[
  scale=0.9,
  text=white,
  rotate=0.0
]{\bfseries 82};
\draw (axis cs:14.15,6) node[
  scale=0.9,
  text=white,
  rotate=0.0
]{\bfseries 83};
\draw (axis cs:15,6) node[
  scale=0.9,
  text=white,
  rotate=0.0
]{\bfseries 54};
\end{axis}
\end{tikzpicture}
    \caption{
        Success rates for the reacher, ball-in-a-cup, and table tennis policies on PAMY2 (higher is better).
        Results are computed for 100 trials.
        The error bars show the 95\% confidence intervals across trials computed with the Wilson score interval.
    }
    \label{fig:success_rates}
\end{figure}

\Cref{fig:success_rates} shows that our method achieves robust zero-shot transfer on all three tasks, even though they require highly dynamic yet accurate motions and precise control is particularly challenging for PAM-actuated systems~\cite{hao2017modeling,minh2010cascade,qin2022active}.
To the best of our knowledge, this result is the first successful sim-to-real transfer with a muscle-actuated robot for tasks of this complexity, underscoring the utility of learned actuator models for these systems.
Since the GenAN is trained on trajectories of the robot without the ball, the successful ball-in-a-cup transfer also indicates robustness to unseen dynamic loads to some extent.
The $\SI{54}{\gram}$ ball adds substantial mass to the $\SI{107}{\gram}$ end effector and generates considerable unseen forces during fast swings.
Although~\cref{fig:accuracy_torque_vs_position} shows that GenAN results in the most accurate simulation, the agent trained with this model outperforms the Torque GenAN baseline agent only in the table tennis task, achieving comparable performance in the other tasks.
We believe that this is the case since table tennis requires the highest accuracy, as even slight errors in the racket position can have drastic effects on the ball landing position.

For all tasks, reducing the action penalty degrades the performance since this setting yields more jittery policies.
Because the GenAN training trajectories (see \cref{subapp:implementation_data_collection}) are relatively smooth, these motions are out-of-distribution for the GenAN, making the simulation less realistic.
Smooth policies are desirable for real robots in any case, as they consume less energy and reduce wear.
Removing the disagreement penalty or the GenAN ensemble altogether degrades the table tennis performance, but the results are mixed for reacher and ball-in-a-cup.
We believe that this is the case because the relatively large dataset covers the state space well, resulting in low epistemic uncertainty.
In \cref{subsec:an_training_data}, we show that the benefits of the GenAN ensemble are clearer in low-data regimes.
Refer to \cref{app:transfer_results} for further evaluations of the transfer, including a comparison to the results of \citet{guist2024safe}, who train their table tennis policy on the real robot.
\Cref{fig:table_tennis_successes_0685} shows that our method matches their real-robot performance while requiring substantially less real-robot data and learning the policy purely in simulation.

\subsection{Influence of the GenAN training dataset size}
\label{subsec:an_training_data}

In this section, we investigate the impact of the GenAN training set size on the simulation accuracy and policy transfer.
\Cref{subfig:an_accuracy_dataset_size} shows the position error of the GenAN-augmented simulation on the test set from~\cref{subsec:actuator_net_accuracy} for different training set sizes.
The errors for the single-step and multi-step predictions decrease for larger datasets, but the decrease stops at around 1500 trajectories, signifying diminishing returns from additional data.
To investigate the policy transfer in different data regimes, we use these models to train reacher policies.
\Cref{subfig:success_rates_reacher_dataset_size} visualizes the transfer success rates in relation to the GenAN training set size, showing that we can reduce the dataset to 1000 trajectories (33 minutes of robot data) without sacrificing performance.
This insight aligns with our earlier finding that the effect of additional data diminishes for large datasets.
Transfer performance degrades only with further dataset reduction.
Furthermore, the graphs highlight that the performance degradation is less severe for the ensemble configurations, suggesting that the ensemble effectively mitigates the effects of epistemic model uncertainty when data is scarce.

\begin{figure}[t]
    \begin{subfigure}[t]{0.49\linewidth}
    \centering
    \mbox{
\begin{tikzpicture}

\definecolor{darkgray176}{RGB}{176,176,176}
\definecolor{darkorange25512714}{RGB}{255,127,14}
\definecolor{forestgreen4416044}{RGB}{44,160,44}

\begin{axis}[
xtick={50, 100, 200, 500, 1000, 1500, 2500},
height=\datasetsizeheight,
font=\datasetsizefontsize,
ticklabel style={font=\datasetsizefontsize},
log basis x={10},
scale only axis,
tick align=outside,
tick pos=left,
width=0.68\linewidth,
x grid style={darkgray176},
xlabel={dataset size (\# trajectories)},
xlabel style={inner sep=0pt},
xmajorgrids,
xmin=41.1170079713445, xmax=3040.10447664664,
xmode=log,
xtick style={color=black},
xticklabel style={
    rotate=45.0,
    anchor=north east,
    xshift=2pt,
    yshift=2pt
},
yticklabel style={color=darkorange25512714},
y grid style={darkgray176},
ylabel=\textcolor{darkorange25512714}{pos.\ error after 1 step (°)},
ylabel style={yshift=-3pt},
ymajorgrids,
ymin=0, ymax=0.00950781438358904,
ytick style={color=black}
]
\addplot [line width=1pt, darkorange25512714, mark=*, mark size=2, mark options={solid}]
table {%
50 0.00887704061702094
100 0.0076940079968234
200 0.00743608998190265
500 0.00530947114681319
1000 0.00463300686020234
1500 0.00416212526451298
2000 0.00374750851938028
2500 0.00364963596066373
};
\path [draw=darkorange25512714, draw opacity=0.8, thick]
(axis cs:50,0.00856290658395053)
--(axis cs:50,0.00922210140504604);

\path [draw=darkorange25512714, draw opacity=0.8, thick]
(axis cs:100,0.0074107341585601)
--(axis cs:100,0.00800216244814154);

\path [draw=darkorange25512714, draw opacity=0.8, thick]
(axis cs:200,0.00718370890727995)
--(axis cs:200,0.00770916171834755);

\path [draw=darkorange25512714, draw opacity=0.8, thick]
(axis cs:500,0.0051111098309192)
--(axis cs:500,0.00552321889718472);

\path [draw=darkorange25512714, draw opacity=0.8, thick]
(axis cs:1000,0.00446946887723956)
--(axis cs:1000,0.00480687187922772);

\path [draw=darkorange25512714, draw opacity=0.8, thick]
(axis cs:1500,0.00401121009693328)
--(axis cs:1500,0.00432481863887308);

\path [draw=darkorange25512714, draw opacity=0.8, thick]
(axis cs:2000,0.00359991124789481)
--(axis cs:2000,0.00390280236157484);

\path [draw=darkorange25512714, draw opacity=0.8, thick]
(axis cs:2500,0.00350784183418608)
--(axis cs:2500,0.00379710033319115);

\addplot [semithick, darkorange25512714, opacity=0.8, mark=-, mark size=4, mark options={solid}, only marks]
table {%
50 0.00856290658395053
100 0.0074107341585601
200 0.00718370890727995
500 0.0051111098309192
1000 0.00446946887723956
1500 0.00401121009693328
2000 0.00359991124789481
2500 0.00350784183418608
};
\addplot [semithick, darkorange25512714, opacity=0.8, mark=-, mark size=4, mark options={solid}, only marks]
table {%
50 0.00922210140504604
100 0.00800216244814154
200 0.00770916171834755
500 0.00552321889718472
1000 0.00480687187922772
1500 0.00432481863887308
2000 0.00390280236157484
2500 0.00379710033319115
};
\end{axis}

\begin{axis}[
xtick=\empty,
axis y line=right,
axis line style={-},
font=\datasetsizefontsize,
height=\datasetsizeheight,
log basis x={10},
scale only axis,
tick align=outside,
width=0.68\linewidth,
x grid style={darkgray176},
xmin=41.1170079713445, xmax=3040.10447664664,
xmode=log,
xtick pos=left,
xtick style={color=black},
y grid style={darkgray176},
ylabel=\textcolor{forestgreen4416044}{pos.\ err.\ after 500 steps (°)},
ylabel style={yshift=3pt},
ymin=0, ymax=11.1498620400416,
ytick pos=right,
ytick style={color=black},
yticklabel style={anchor=west,color=forestgreen4416044}
]
\addplot [line width=1pt, forestgreen4416044, mark=square*, mark size=2, mark options={solid}]
table {%
50 10.360386956689
100 9.14152262382695
200 6.55953780151114
500 5.24288565030196
1000 4.95648557311949
1500 4.60999093029828
2000 4.53720971516307
2500 4.54288389481606
};
\path [draw=forestgreen4416044, draw opacity=0.8, semithick]
(axis cs:50,9.94795576745169)
--(axis cs:50,10.8284743051533);

\path [draw=forestgreen4416044, draw opacity=0.8, semithick]
(axis cs:100,8.76131293540414)
--(axis cs:100,9.57131412834658);

\path [draw=forestgreen4416044, draw opacity=0.8, semithick]
(axis cs:200,6.27309359446946)
--(axis cs:200,6.91689207467482);

\path [draw=forestgreen4416044, draw opacity=0.8, semithick]
(axis cs:500,5.07011122067218)
--(axis cs:500,5.42738170171092);

\path [draw=forestgreen4416044, draw opacity=0.8, semithick]
(axis cs:1000,4.79628871247621)
--(axis cs:1000,5.12128876173953);

\path [draw=forestgreen4416044, draw opacity=0.8, semithick]
(axis cs:1500,4.46056460282078)
--(axis cs:1500,4.76664791550226);

\path [draw=forestgreen4416044, draw opacity=0.8, semithick]
(axis cs:2000,4.40071960738744)
--(axis cs:2000,4.68767638229005);

\path [draw=forestgreen4416044, draw opacity=0.8, semithick]
(axis cs:2500,4.40174323126967)
--(axis cs:2500,4.69161375049304);

\addplot [semithick, forestgreen4416044, opacity=0.8, mark=-, mark size=4, mark options={solid}, only marks]
table {%
50 9.94795576745169
100 8.76131293540414
200 6.27309359446946
500 5.07011122067218
1000 4.79628871247621
1500 4.46056460282078
2000 4.40071960738744
2500 4.40174323126967
};
\addplot [semithick, forestgreen4416044, opacity=0.8, mark=-, mark size=4, mark options={solid}, only marks]
table {%
50 10.8284743051533
100 9.57131412834658
200 6.91689207467482
500 5.42738170171092
1000 5.12128876173953
1500 4.76664791550226
2000 4.68767638229005
2500 4.69161375049304
};
\end{axis}

\end{tikzpicture}}
    \caption{ 
        Mean absolute position error for GenANs trained on datasets of different sizes (lower is better).
    }
    \label{subfig:an_accuracy_dataset_size}
    \end{subfigure}
    \hfill
    \begin{subfigure}[t]{0.49\linewidth}
    \mbox{
\begin{tikzpicture}

\definecolor{crimson2143940}{RGB}{214,39,40}
\definecolor{darkgray176}{RGB}{176,176,176}
\definecolor{lightgray204}{RGB}{204,204,204}
\definecolor{steelblue31119180}{RGB}{31,119,180}

\begin{axis}[
width=0.68\linewidth,height=\datasetsizeheight,scale only axis,legend cell align={left},
legend style={
  fill opacity=0.8,
  draw opacity=1,
  text opacity=1,
  at={(0.98,0.03)},
  anchor=south east,
  draw=lightgray204
},
font=\datasetsizefontsize,
log basis x={10},
tick align=outside,
tick pos=left,
x grid style={darkgray176},
xlabel={dataset size (\(\displaystyle \#\) trajectories)},
xlabel style={inner sep=0pt},
ticklabel style={font=\datasetsizefontsize},
xmajorgrids,
xmin=41.1170079713445, xmax=3040.10447664664,
xmode=log,
xtick style={color=black},
xtick={50, 100, 200, 500, 1000, 1500, 2500},
xticklabel style={rotate=45.0,anchor=north east,xshift=2pt,yshift=2pt},
y grid style={darkgray176},
ylabel={success rate (\%)},
ymajorgrids,
ymin=0, ymax=100,
ytick style={color=black}
]
\addplot [line width=1pt, steelblue31119180, mark=*, mark size=2, mark options={solid}]
table {%
50 29
100 34
200 53
500 78
1000 89
1500 84
2000 90
2500 90
};
\addlegendentry{Ensemble}
\addplot [line width=1pt, crimson2143940, mark=square*, mark size=2, mark options={solid}]
table {%
50 4
100 9
200 21
500 63
1000 85
1500 77
2000 94
2500 93
};
\addlegendentry{No ensemble}
\path [draw=steelblue31119180, draw opacity=0.8, thick]
(axis cs:50,21.0148357491223)
--(axis cs:50,38.5388911755711);

\path [draw=steelblue31119180, draw opacity=0.8, thick]
(axis cs:100,25.4615207973482)
--(axis cs:100,43.7222711452754);

\path [draw=steelblue31119180, draw opacity=0.8, thick]
(axis cs:200,43.2888569700994)
--(axis cs:200,62.4891820406587);

\path [draw=steelblue31119180, draw opacity=0.8, thick]
(axis cs:500,68.9296464850142)
--(axis cs:500,84.9987176153946);

\path [draw=steelblue31119180, draw opacity=0.8, thick]
(axis cs:1000,81.3687034969197)
--(axis cs:1000,93.7458036429354);

\path [draw=steelblue31119180, draw opacity=0.8, thick]
(axis cs:1500,75.5797306107298)
--(axis cs:1500,89.9047115111952);

\path [draw=steelblue31119180, draw opacity=0.8, thick]
(axis cs:2000,82.5634338495087)
--(axis cs:2000,94.4770862939325);

\path [draw=steelblue31119180, draw opacity=0.8, thick]
(axis cs:2500,82.5634338495087)
--(axis cs:2500,94.4770862939325);

\addplot [semithick, steelblue31119180, opacity=0.8, mark=-, mark size=4, mark options={solid}, only marks, forget plot]
table {%
50 21.0148357491223
100 25.4615207973482
200 43.2888569700994
500 68.9296464850142
1000 81.3687034969197
1500 75.5797306107298
2000 82.5634338495087
2500 82.5634338495087
};
\addplot [semithick, steelblue31119180, opacity=0.8, mark=-, mark size=4, mark options={solid}, only marks, forget plot]
table {%
50 38.5388911755711
100 43.7222711452754
200 62.4891820406587
500 84.9987176153946
1000 93.7458036429354
1500 89.9047115111952
2000 94.4770862939325
2500 94.4770862939325
};
\path [draw=crimson2143940, draw opacity=0.8, thick]
(axis cs:50,1.56633039915476)
--(axis cs:50,9.83707143588792);

\path [draw=crimson2143940, draw opacity=0.8, thick]
(axis cs:100,4.80725400025652)
--(axis cs:100,16.2262128527163);

\path [draw=crimson2143940, draw opacity=0.8, thick]
(axis cs:200,14.1656540619153)
--(axis cs:200,29.9799688340899);

\path [draw=crimson2143940, draw opacity=0.8, thick]
(axis cs:500,53.2205295809429)
--(axis cs:500,71.8176394656755);

\path [draw=crimson2143940, draw opacity=0.8, thick]
(axis cs:1000,76.7164404091676)
--(axis cs:1000,90.6940147163434);

\path [draw=crimson2143940, draw opacity=0.8, thick]
(axis cs:1500,67.8456169771262)
--(axis cs:1500,84.1567341196965);

\path [draw=crimson2143940, draw opacity=0.8, thick]
(axis cs:2000,87.5231845541041)
--(axis cs:2000,97.2213876036812);

\path [draw=crimson2143940, draw opacity=0.8, thick]
(axis cs:2500,86.2504852609265)
--(axis cs:2500,96.5680738932727);

\addplot [semithick, crimson2143940, opacity=0.8, mark=-, mark size=4, mark options={solid}, only marks, forget plot]
table {%
50 1.56633039915476
100 4.80725400025652
200 14.1656540619153
500 53.2205295809429
1000 76.7164404091676
1500 67.8456169771262
2000 87.5231845541041
2500 86.2504852609265
};
\addplot [semithick, crimson2143940, opacity=0.8, mark=-, mark size=4, mark options={solid}, only marks, forget plot]
table {%
50 9.83707143588792
100 16.2262128527163
200 29.9799688340899
500 71.8176394656755
1000 90.6940147163434
1500 84.1567341196965
2000 97.2213876036812
2500 96.5680738932727
};
\end{axis}

\end{tikzpicture}}
    \caption{
        Reacher transfer success rate when using GenANs trained with different dataset sizes (higher is better).
    }
    \label{subfig:success_rates_reacher_dataset_size}
    \end{subfigure}
    \caption{Influence of the GenAN training dataset size on (a) the simulation position error and (b) the transfer success rate for the reacher task.
    The error bars visualize the 95\% confidence intervals.
    Smaller training sets reduce the simulation accuracy both after a single step ($\SI{2}{\ms}$; orange plot) and after 500 steps ($\SI{1}{\s}$; green plot).
    With a single GenAN, instead of the ensemble, policy performance degrades more severely for smaller datasets.}
    \label{fig:dataset_size}
\end{figure}

\section{Limitations}
\label{sec:limitations}
Training GenANs requires a dataset with good workspace coverage.
Collecting suitable data might necessitate tuning parameters of the exploration policy (\cref{subapp:implementation_data_collection}).
Scaling the method to systems with more DoFs exacerbates the challenge of achieving good coverage due to the curse of dimensionality.
Therefore, in future work, we plan to integrate an active exploration approach that automatically discovers the most informative data to guide exploration~\cite{pathak2019self,shyam2019model,schneider2022active}.

Although the successful ball-in-a-cup transfer indicates that our pipeline is robust to changes in the end-effector load, substantially larger changes may still degrade the GenAN accuracy.
Vastly different loads might significantly alter the relationship between commands and resulting positions, potentially producing command-position combinations not represented in the training data.
Achieving robustness to very large load changes, therefore, might require adding different weights to the end effector already during the data collection.
Environment contacts can induce a similar distribution shift, suggesting that contact-rich tasks may require representative contacts in the training data.
To learn the actuator model on this data would require a very accurate contact model to disentangle actuator torques from the effects of the external contact forces.

In our experiments, we observed that PAMY2's dynamics change slowly over time due to tendon elongation, wear, and slight deformations of the 3D-printed parts, which likely also occur in other complex robots.
Currently, these changes necessitate collecting new data and retraining the GenAN at regular intervals when using the robot for a long time.
In future work, we plan to develop an adaptive model that addresses this limitation while also enabling the transfer between robot instances.

\section{Conclusion}
\label{sec:conclusion}
In this paper, we showed the utility of learned actuator models for muscle- and tendon-actuated robot arms, demonstrating that these models overcome longstanding modeling challenges that have impeded sim-to-real learning for such complex robots.
Our Generalized Actuator Network requires only joint position measurements, thereby eliminating the need for torque sensors and making the method applicable to a wide range of robots with different actuation types.
By integrating our actuator model with a standard robotics simulator, we enable training RL policies for zero-shot transfer to physical robots.
We validated our method on PAMY2, a complex 4-DoF tendon-driven and PAM-actuated robot, by learning three tasks that require dynamic yet precise motions: reacher, ball-in-a-cup, and table tennis.
Notably, these results constitute the first sim-to-real transfer for a multi-joint muscle-actuated robot.
We further demonstrated that ensembles of actuator models provide an effective means to handle epistemic uncertainty when training on limited data, establishing GenAN as an efficient policy learning framework for robots with complex actuation.

\acknowledgments{We thank Felix Grüninger, Heiko Ott, and Thomas Steinbrenner for support with the robot hardware and 3D printing; Gökce Ergün and Senya Polikovsky for help with the ball tracking; and Leyla Gurbanova for implementing an early version of the ball-in-a-cup task.
This work was supported by the Max Planck Institute for Intelligent Systems.
Jan Schneider was supported by the Konrad Zuse School of Excellence in Learning and Intelligent Systems (ELIZA), sponsored by the German Federal Ministry of Education and Research.}


\bibliography{bibliography}

\appendix

\crefalias{section}{appendix}
\crefalias{subsection}{appendix}
\crefalias{subsubsection}{appendix}
\section{Implementation details and hyperparameters}
\label{app:implementation_details}
This section describes implementation details and hyperparameters for the data collection, the GenAN training, the UAN implementation, the RL tasks, and the policy optimization.

\newcolumntype{A}{>{\raggedright\arraybackslash}p{0.28\linewidth}}
\newcolumntype{B}{>{\centering\arraybackslash}p{0.4\linewidth}}

\subsection{Data collection}
\label{subapp:implementation_data_collection}

\begin{figure}[ht]
    \begin{subfigure}[t]{0.48\linewidth}
        \input{plots/exploration_controls}
        \caption{Control signals}
        \label{fig:exploration_controls}
    \end{subfigure}
    \begin{subfigure}[t]{0.48\linewidth}
        \input{plots/exploration_positions}
        \caption{Resulting position trajectories}
        \label{fig:exploration_positions}
    \end{subfigure}
    \caption{
        Five exploration trajectories collected by the open-loop exploration policy.
        Each row corresponds to one DoF of the robot.
    }
    \label{fig:exploration}
\end{figure}

To collect the data for training the GenAN, we use the following open-loop exploration scheme.
Each exploration trajectory is two seconds long and generated by sampling control signals every 0.5 seconds and fitting a cubic spline between these commands.
As an additional randomization step, we sample a noise level $\sigma \sim \mathcal{U}(0.0005, 0.003)$ for each trajectory and add zero-mean Gaussian noise at every time step $n_t \sim \mathcal{N}(0, \sigma^2)$.
Afterward, we clip the control signal to the range of valid control signals $[-1, 1]$.
This scheme yields relatively smooth but diverse exploration trajectories that span the robot's workspace and exhibit different velocity profiles.
To prevent damage to the robot during the exploration motions, we set the minimum and maximum pressures of the PAMs so that the robot does not reach its joint limits, even at the control limits.
Furthermore, we utilize a soft limit on the joint velocities at $\SI{250}{\degree\per\second}$, $\SI{200}{\degree\per\second}$, $\SI{400}{\degree\per\second}$, and $\SI{400}{\degree\per\second}$, respectively, for the four joints.
If a joint exceeds its limit, we divide the commanded control increments $\Delta \ctrl{}$ by a factor of \num{10} to slow down the motion until the velocity falls below $\SI{80}{\percent}$ of the velocity limit.
\Cref{fig:exploration} shows five example trajectories collected by the exploration policy.

At each step, we record the current joint position $\qreal{t}$ and control signal $\ctrl{t}$.
We compute joint velocities $\qdreal{t}$ and accelerations $\qddreal{t}$ from the positions via backward differences and central differences, respectively.
See \cref{app:computing_velocities} for an explanation of this choice.

\subsection{GenAN training and evaluation}
\label{app:implementation_details_genan}

We split the dataset collected according to \cref{subapp:implementation_data_collection} into $\SI{80}{\%}$ training and $\SI{20}{\%}$ validation data by assigning trajectories randomly to the two splits.
Splitting at the step level instead would mean that the network is trained on data points that are potentially very similar to the validation data points due to the temporal correlation within the trajectory.
We train the GenAN for 150 epochs, which typically takes around 25 minutes on an Nvidia A100 GPU. 
After the training, we select the model with the lowest validation loss. 

The test set of 800 trajectories was collected separately after the transfer experiments of \cref{subsec:transfer}.
Due to the gradual dynamics changes mentioned in \cref{sec:limitations}, the performance on the test set is potentially a conservative estimate of the GenAN accuracy directly after training.
The hundreds of interactions with the real system during the transfer experiments could already have led to slight dynamics changes in the robot, which would increase the position errors measured in \cref{subsec:actuator_net_accuracy,app:history}.

\Cref{tab:hyperparameters_genan} lists the complete GenAN hyperparameter configuration used in our experiments.

\begin{table}[ht]
    \caption{Hyperparameters for the GenAN training.}
    \label{tab:hyperparameters_genan}
    \centering
    \begin{tabularx}{0.6\linewidth}{@{} X >{\raggedright\arraybackslash}p{0.2\linewidth} >{\centering\arraybackslash}p{0.2\linewidth} @{}}
    \toprule
    & \textbf{Hyperparameter} & \textbf{Value} \\
    \midrule
    \multirow{2}{*}{Input}
        & History length & 3 \\
        & History stride & 1 \\
    \midrule
    \multirow{4}{*}{Architecture}
        & Hidden layers & 2 \\
        & Neurons per layer & 512 \\
        & Activation function & tanh \\
        & Ensemble size & 5 \\
    \midrule
    \multirow{3}{*}{Training}
        & Optimizer & Adam \\
        & Learning rate & \num{1e-4} \\
        & Batch size & 512 \\
    \bottomrule
    \end{tabularx}
\end{table}

\subsection{UAN training}
\label{subapp:implementation_details_uan}

As mentioned in \cref{subsubsec:uan}, the main difference of our UAN adaptation to the original implementation in \cite{fey2025bridging} is that we predict the full joint torques, rather than residuals.
We also modify the network architecture to account for coupling between joints. 
The original UAN always receives only the state, control signal, and history of one joint and predicts its scalar torque.
This model is then applied independently to each joint.
In contrast, our network jointly processes the inputs of all joints and outputs the complete torque vector $\tor{t}$.
The reward function follows the original formulation
\begin{equation*}
    r_t = c^\mathrm{pos} \exp(-c^\mathrm{exp} \lVert \qreal{t+1} - \qpred{t+1}\rVert_2^2) + c^\mathrm{smooth} \exp(-0.125 \lVert \mathbf{a}_t - \mathbf{a}_{t-1}\rVert_1^2),
\end{equation*}
where $\qpred{t+1} = \step(\qpred{t}, \boldsymbol{\dot{\hat{q}}}_{t}, \torpred{t})$ for all $t \in \{H, \ldots, T - 1\}$ are the simulated positions for the torque predictions of the UAN $\torpred{t} = f_{\boldsymbol{\theta}}\mleft(\qpred{t-H:t}, \ctrl{t-H:t}\mright)$, and $\qpred{t'} = \qreal{t'}$ for all $t' \in \{0, \ldots, H\}$.
$\qreal{t}$ are the $t$-th joint positions in the dataset trajectory.

\Cref{tab:hyperparameters_uan} displays the hyperparameters that we use for the UAN baseline.
The values are the result of a hyperparameter sweep over 200 UAN training configurations.

\begin{table}[ht]
    \centering
    \caption{Hyperparameters for the UAN training. The values are the result of a hyperparameter sweep over 200 training configurations.}
    \label{tab:hyperparameters_uan}
    \begin{tabularx}{0.85\linewidth}{@{} X A B @{}}
        \toprule
        & \multicolumn{1}{c}{\textbf{Hyperparameter}} & \textbf{Value} \\
        \midrule
        \multirow{7}{*}{Environment}
            & Parallel instances & 1024 \\
            & Episode length & $\SI{2}{\s}$ \\
            & $c^\mathrm{exp}$ & 17.56 \\
            & $c^\mathrm{pos}$ & 5.769 \\
            & $c^\mathrm{smooth}$ & 0.6789 \\
            & History length & 1 \\
            & History stride & 1 \\
        \midrule
        \multirow{14}{*}{PPO (skrl)}
            & discount\_factor & 0.9783 \\
            & lambda & 0.95 \\
            & learning\_rate & \num{2.166e-05} \\
            & entropy\_loss\_scale & 0.025 \\
            & ratio\_clip & 0.2676 \\
            & rollouts & 1024 \\
            & mini\_batches & 16 \\
            & learning\_epochs & 5 \\
            & observation\_preprocessor & RunningStandardScaler \\
            & value\_preprocessor & RunningStandardScaler \\
            & grad\_norm\_clip & 1.0 \\
            & value\_clip & 0.2 \\
            & value\_loss\_scale & 1.0 \\
            & kl\_threshold & 0.008 \\
        \midrule
        \multirow{3}{*}{Policy}
            & Hidden layers & 3 \\
            & Neurons per layer & 256 \\
            & Activation function & tanh \\
        \bottomrule
    \end{tabularx}
\end{table}

\subsection{RL tasks}
\label{subapp:implementation_rl_tasks}

For all tasks, we run the simulator at $\SI{500}{\Hz}$ for numerical stability but query the RL agent at $\SI{100}{\Hz}$ by repeating each action for 5 steps.
We found that this lower policy frequency makes policy learning more efficient and robust.
To obtain diverse yet stable and realistic initial conditions in the simulator, we sample initial control signals $\ctrl{\mathrm{init}} \sim \mathcal{U}(\boldsymbol{u}_\mathrm{init}^\mathrm{min}, \boldsymbol{u}_\mathrm{init}^\mathrm{max})$.
We then set the robot joint position to the intermediate angles $(0, \SI{45}{\degree}, \SI{45}{\degree}, 0)$ and apply $\ctrl{\mathrm{init}}$ for 500 steps, i.e., 1 second in simulation time.
In contrast to directly sampling random initial positions, this scheme ensures that the combination of controls and positions corresponds to a stable configuration on the real system.
The RL episode starts after these initial 500 steps.
During the transfer experiments, we follow a similar procedure.
We sample initial controls from the same distribution to test the policy in diverse initial conditions.
The only difference is that we ramp the controls linearly to $\ctrl{\mathrm{init}}$ over 2 seconds to avoid unnecessarily aggressive motions during the reset.

The policy actions are desired changes to the control signal $\boldsymbol{a}_t = \Delta \boldsymbol{u}_t$.
We squash the policy output to $[-1, 1]$ using the $\tanh$ function.
Afterward, the result is scaled by $\Delta u^\mathrm{max} = 0.01$ to prevent the policy from executing overly aggressive motions, resulting in the following mapping from policy output $\boldsymbol{\hat{a}}_t$ to control signal change $\Delta \ctrl{t}$.
\begin{equation}
    \Delta \ctrl{t} = \Delta u^\mathrm{max} \tanh(\boldsymbol{\hat{a}}_t)
\end{equation}

For the reward weighting constants, we use $c^\mathrm{act} = 1250$, $c^\mathrm{disag} = 0.025$, and $c^\mathrm{lim} = 1$ for all tasks.

\subsubsection{Reacher}
\label{subapp:implementation_reacher}

For the reacher task, we sample the goal positions for each joint according to $\boldsymbol{g} \sim \mathcal{U}(\boldsymbol{g}^\mathrm{min}, \boldsymbol{g}^\mathrm{max})$.
In addition to the general observations described in \cref{subsec:rl_task}, the agent gets the goal position as task-specific observation $\boldsymbol{o}_t^\mathrm{task} = \boldsymbol{g}$.
The task-specific reward is the negative joint-space distance to the target.
\begin{equation}
    r^\mathrm{task}_t = -\lVert \boldsymbol{q}_t - \boldsymbol{g}\rVert
    \label{eq:reward_reacher}
\end{equation}
Note that the distance reward is applied every step and therefore encourages the agent to move quickly to the goal.

The complete configuration of environment and agent hyperparameters that we use for the reacher task is given in \cref{tab:hyperparameters_reacher}.

\renewcommand{\arraystretch}{1.2}
\begin{table}[ht]
    \centering
    \caption{Hyperparameters for the reacher task.}
    \label{tab:hyperparameters_reacher}
    \begin{tabularx}{0.85\linewidth}{@{} X A B @{}}
        \toprule
        & \multicolumn{1}{c}{\textbf{Hyperparameter}} & \textbf{Value} \\
        \midrule
        \multirow{10}{*}{Environment}
            & Parallel instances & 1024 \\
            & Episode length & $\SI{2}{\s}$ \\
            & Action repeat & 5 \\
            & $\Delta u^\mathrm{max}$ & 0.01 \\
            & $\boldsymbol{u}_\mathrm{init}^\mathrm{min}$ & $(-0.5, -0.6, -0.6, -0.5)$ \\
            & $\boldsymbol{u}_\mathrm{init}^\mathrm{max}$ & $(0.5, 0.0, 0.4, 0.5)$ \\
            & $\boldsymbol{q}^\mathrm{min}$ & $(\SI{-90}{\degree}, \SI{-75}{\degree}, \SI{-85}{\degree}, \SI{-85}{\degree})$ \\
            & $\boldsymbol{q}^\mathrm{max}$ & $(\SI{90}{\degree}, \SI{85}{\degree}, \SI{85}{\degree}, \SI{85}{\degree})$ \\
            & $\boldsymbol{g}^\mathrm{min}$ & $(-\SI{50}{\degree}, \SI{20}{\degree}, -\SI{50}{\degree}, -\SI{50}{\degree})$ \\
            & $\boldsymbol{g}^\mathrm{max}$ & $(\SI{50}{\degree}, \SI{60}{\degree}, \SI{50}{\degree}, \SI{50}{\degree})$ \\
        \midrule
        \multirow{14}{*}{PPO (skrl)}
            & discount\_factor & 0.9801 \\
            & lambda & 0.95 \\
            & learning\_rate & \num{3.949e-05} \\
            & entropy\_loss\_scale & 0.025 \\
            & ratio\_clip & 0.1521 \\
            & rollouts & 64 \\
            & mini\_batches & 32 \\
            & learning\_epochs & 10 \\
            & observation\_preprocessor & RunningStandardScaler \\
            & value\_preprocessor & RunningStandardScaler \\
            & grad\_norm\_clip & 1.0 \\
            & value\_clip & 0.2 \\
            & value\_loss\_scale & 1.0 \\
            & kl\_threshold & 0.008 \\
        \midrule
        \multirow{3}{*}{Policy}
            & Hidden layers & 4 \\
            & Neurons per layer & 64 \\
            & Activation function & LeakyReLU \\
        \bottomrule
    \end{tabularx}
\end{table}

\subsubsection{Ball-in-a-cup}
\label{subapp:implementation_ball_in_cup}

To obtain the real ball position and velocity during the transfer, we use a Vicon object tracking system.
Placing 15 flat reflective markers in an irregular pattern on the ball yields reasonably reliable tracking.
We observe two issues with this solution: the reflectiveness of the ball's material and occlusions when the ball is near or inside the cup.
The tracking software occasionally misidentifies reflections on the ball as markers, resulting in small errors in the ball position measurements, while the occlusions cause missing detections.
To make the agent more robust to these errors, we inject zero-mean Gaussian noise with $\sigma^b = \SI{0.5}{\cm}$ into the ball positions during training in simulation, and with a probability of 5 percent, we omit the ball position entirely.
Both in simulation and on the real system, we maintain a buffer of the last 5 ball positions and provide the last successful position measurement to the agent, as well as the finite-difference ball velocity averaged over the buffer.

In the simulated ball-in-a-cup environment, the ball is initialized uniformly on a sphere centered at the string attachment point, with radius equal to the string length.
We chose this initialization over initializing the ball only below the cup to increase the diversity of the initial conditions.
In the real environment, however, the ball naturally always starts below the cup.
There is no domain randomization on the parameters governing the ball and string dynamics.
Applying domain randomization here could further increase the robustness of the policy to the differences between the simulated and real environments and could, therefore, be explored in future work.

The task-specific observations contain the ball position and velocity in cup frame $\boldsymbol{o}_t^\mathrm{task} = (\boldsymbol{x}^b_t, \boldsymbol{\dot{x}}^b_t)$.
The task reward is defined as $r^\mathrm{task}_t = r^\mathrm{cup}_t + c^\mathrm{vel} r^\mathrm{vel}_t$, where 
$$
r^\mathrm{cup}_t = 
\begin{cases}
10, &\qquad\text{if the ball is in the cup} \\
0,  &\qquad\text{otherwise}
\end{cases}
$$
is a sparse success reward, and $r^\mathrm{vel}_t = -\lVert\boldsymbol{\dot{q}}_t\rVert^2$ is a joint velocity penalty that discourages overly aggressive motions with $c^\mathrm{vel} = 0.0025$.

The complete configuration of environment and agent hyperparameters for the ball-in-a-cup task is given in \cref{tab:hyperparameters_ball_in_cup}.

\begin{table}[ht]
    \centering
    \caption{Hyperparameters for the ball-in-a-cup task.}
    \label{tab:hyperparameters_ball_in_cup}
    \begin{tabularx}{0.85\linewidth}{@{} X A B @{}}
        \toprule
        & \multicolumn{1}{c}{\textbf{Hyperparameter}} & \textbf{Value} \\
        \midrule
        \multirow{12}{*}{Environment}
            & Parallel instances & 1024 \\
            & Episode length & $\SI{2}{\s}$ \\
            & Action repeat & 5 \\
            & $\Delta u^\mathrm{max}$ & 0.01 \\
            & $\boldsymbol{u}_\mathrm{init}^\mathrm{min}$ & $(-0.5, -0.6, -0.6, -0.5)$ \\
            & $\boldsymbol{u}_\mathrm{init}^\mathrm{max}$ & $(0.5, 0.0, 0.4, 0.5)$ \\
            & $\boldsymbol{q}^\mathrm{min}$ & $(\SI{-90}{\degree}, \SI{-75}{\degree}, \SI{-85}{\degree}, \SI{-85}{\degree})$ \\
            & $\boldsymbol{q}^\mathrm{max}$ & $(\SI{90}{\degree}, \SI{85}{\degree}, \SI{85}{\degree}, \SI{85}{\degree})$ \\
            & String length & $\SI{20}{\cm}$ \\
            & Sim. ball noise std $\sigma^b$ & $\SI{0.5}{\cm}$ \\
            & Sim. ball dropout rate & $\SI{5}{\%}$ \\
            & Ball pos. buffer size & 5 \\
        \midrule
        \multirow{14}{*}{PPO (skrl)}
            & discount\_factor & 0.9835 \\
            & lambda & 0.95 \\
            & learning\_rate & \num{1.201e-04} \\
            & entropy\_loss\_scale & 0.005 \\
            & ratio\_clip & 0.05882 \\
            & rollouts & 1024 \\
            & mini\_batches & 128 \\
            & learning\_epochs & 10 \\
            & observation\_preprocessor & RunningStandardScaler \\
            & value\_preprocessor & RunningStandardScaler \\
            & grad\_norm\_clip & 1.0 \\
            & value\_clip & 0.2 \\
            & value\_loss\_scale & 1.0 \\
            & kl\_threshold & 0.008 \\
        \midrule
        \multirow{3}{*}{Policy}
            & Hidden layers & 3 \\
            & Neurons per layer & 128 \\
            & Activation function & ELU \\
        \bottomrule
    \end{tabularx}
\end{table}

\subsubsection{Table tennis}
\label{subapp:implementation_table_tennis}

The table tennis task largely follows the design of~\citet{buchler2022learning}.
Each episode contains a single ball shot by the ball launcher AIMY~\cite{dittrich2023aimy} and the robot's attempt to return the ball to the other side.
As in~\cite{buchler2022learning}, we keep the ball launcher settings fixed across episodes.
Nevertheless, the resulting ball trajectories vary considerably due to imperfections in the balls, ball launcher wheels, and other components.

To track the table tennis balls in the real environment, we use the visual ball-tracking system developed by \citet{gomez2019reliable}.
We additionally deploy the extended Kalman filter (EKF) implemented by \citet{ma2022learning}.
Similar to~\cite{buchler2022learning}, the simulated training environment makes use of a dataset of 100 previously recorded ball trajectories.
Instead of simulating the entire ball trajectory, we sample a trajectory from the dataset and replay it in the simulator until the ball makes contact with the racket.
During racket contact, we deploy the custom rebound model from~\cite{mulling2011biomimetic,buchler2022learning}.
After contact, we simulate the flight path with MJX.
Using recorded trajectories ensures that the RL agent learns to handle the noise of the ball-tracking system, facilitating a more robust transfer.
The recorded ball trajectories are also filtered by the EKF.

The task-specific observations $\boldsymbol{o}_t^\mathrm{task} = (\boldsymbol{x}^b_t, \boldsymbol{\dot{x}}^b_t, \boldsymbol{x}^r_t, \boldsymbol{\dot{x}}^r_t, h_t)$ contain the ball position and velocity in the world frame $\boldsymbol{x}^b_t, \boldsymbol{\dot{x}}^b_t$, the racket position and velocity $\boldsymbol{x}^r_t, \boldsymbol{\dot{x}}^r_t$, and $h_t$, a binary indicator of whether the ball has already been hit in this episode.
The task reward is defined as $r^\mathrm{task}_t = c^\mathrm{hit} r^\mathrm{hit}_t + c^\mathrm{tt} r^\mathrm{tt}_t$, where 
$$
r^\mathrm{hit}_t = 
\begin{cases}
1, &\qquad\text{if}~h_t~\text{and}~\neg h_{t-1} \\
0,  &\qquad\text{otherwise}
\end{cases}
$$
is a sparse reward that is $1$ only for the first ball hit in the episode and
$$
r^\mathrm{tt}_t = 
\begin{cases}
    \max(1 - 2.96 \lVert \mathbf{b}^\mathrm{land} - \mathbf{b}^\mathrm{des} \rVert^\frac{3}{4}, 0), &\qquad\text{if}~h_{t} \\
    0, &\qquad\text{otherwise}
\end{cases}
$$
is the landing point reward from~\cite{buchler2022learning}.
For the weighting constants, we use $c^\mathrm{hit} = 800$ and $c^\mathrm{tt} = 5000$.
Here, $\mathbf{b}^\mathrm{land}$ is the ball landing point in the table frame, and $\mathbf{b}^\mathrm{des}$ is the target landing point in the center of the opponent's side of the table.
In contrast to~\cite{buchler2022learning}, we use a sparse hit reward, as we found that additional reward shaping is not necessary in our setting.
A further implementation difference is that we terminate each episode after a fixed length of $\SI{1.09}{\s}$ to keep the parallelized environments synchronized for ease of implementation.

The complete configuration of environment and agent hyperparameters for the table tennis task is given in \cref{tab:hyperparameters_table_tennis}.

\begin{table}[ht]
    \centering
    \caption{Hyperparameters for the table tennis task.}
    \label{tab:hyperparameters_table_tennis}
    \begin{tabularx}{0.85\linewidth}{@{} X A B @{}}
        \toprule
        & \multicolumn{1}{c}{\textbf{Hyperparameter}} & \textbf{Value} \\
        \midrule
        \multirow{12}{*}{Environment}
            & Parallel instances & 1024 \\
            & Episode length & $\SI{1.09}{\s}$ \\
            & Action repeat & 5 \\
            & $\Delta u^\mathrm{max}$ & 0.01 \\
            & $\boldsymbol{u}_\mathrm{init}^\mathrm{min}$ & $(-0.2, -0.5, 0.0, -0.2)$ \\
            & $\boldsymbol{u}_\mathrm{init}^\mathrm{max}$ & $(0.2, 0.0, 0.5, 0.2)$ \\
            & $\boldsymbol{q}^\mathrm{min}$ & $(\SI{-90}{\degree}, \SI{-75}{\degree}, \SI{-85}{\degree}, \SI{-85}{\degree})$ \\
            & $\boldsymbol{q}^\mathrm{max}$ & $(\SI{90}{\degree}, \SI{85}{\degree}, \SI{85}{\degree}, \SI{85}{\degree})$ \\
            & EKF measurement noise & \num{1e-03} \\
            & EKF transition noise (pos.) & $10^{-4}$ \\
            & EKF transition noise (vel.) & $10^{-2}$ \\
            & EKF transition noise (spin) & $10^{-4}$ \\
        \midrule
        \multirow{14}{*}{PPO (skrl)}
            & discount\_factor & 0.9815 \\
            & lambda & 0.95 \\
            & learning\_rate & \num{9.699e-05} \\
            & entropy\_loss\_scale & 0.025 \\
            & ratio\_clip & 0.1029 \\
            & rollouts & 128 \\
            & mini\_batches & 16 \\
            & learning\_epochs & 10 \\
            & observation\_preprocessor & RunningStandardScaler \\
            & value\_preprocessor & RunningStandardScaler \\
            & grad\_norm\_clip & 1.0 \\
            & value\_clip & 0.2 \\
            & value\_loss\_scale & 1.0 \\
            & kl\_threshold & 0.008 \\
        \midrule
        \multirow{3}{*}{Policy}
            & Hidden layers & 4 \\
            & Neurons per layer & 128 \\
            & Activation function & tanh \\
        \bottomrule
    \end{tabularx}
\end{table}

\section{Computing velocities and accelerations}
\label{app:computing_velocities}

Let $\qreal{t}$ be the real robot trajectory and $\qsim{t}$ the simulated trajectory for $t \in \{0, \ldots, T\}$.
To compute the labels for the torque loss of \cref{eq:torque_loss}, we compute the torques that reproduce the real trajectory in the simulator.
We initialize the simulator with the first two positions of the trajectory by setting $\qsim{0} = \qreal{0}$ and $\qsim{1} = \qreal{1}$ and compute the torques $\tor{t}$ that result in $\qsim{t+1} = \qreal{t+1}$ for $t \in \{1, \ldots, T-1\}$.

MuJoCo and many other dynamics simulators use a symplectic Euler integrator, which combines an explicit integration step for the velocities
\begin{align}
     \qdsim{t+1} &= \qdsim{t} + \dt\,\qddsim{t} \label{eq:int_dotqtp1}
\end{align}
with an implicit step for the positions
\begin{align}
    \qsim{t+1} &= \qsim{t} + \dt\,\qdsim{t+1}. \label{eq:int_qtp1}
\end{align}

Furthermore, the inverse dynamics equation of a robot manipulator is given by 
\begin{align}
    \boldsymbol{\tau}_t &= \M{\qsim{t}} \qddsim{t} + \cor{\qsim{t}}{\qdsim{t}} + \grav{\qsim{t}} \label{eq:inv_dyn}\\
    &= \invdyn\mleft(\qsim{t},\qdsim{t},\qddsim{t}\mright),
\end{align}
where $\boldsymbol{M}(\qsim{t})$ is the mass matrix, $\cor{\qsim{t}}{\qdsim{t}}$ are the centrifugal and Coriolis forces, and $\grav{\qsim{t}}$ is the gravity vector.

\noindent By rearranging, we obtain the forward dynamics equation
\begin{align}
    \qddsim{t} &= \Minv{\qsim{t}} \left(\tor{t}- \cor{\qsim{t}}{\qdsim{t}} - \grav{\qsim{t}}\right). \label{eq:fwd_dyn}
\end{align}

Assume an arbitrary $t \in \{1, \ldots, T-1\}$, $\qsim{t-1} = \qreal{t-1}$, and $\qsim{t} = \qreal{t}$.
We show how to compute the torque $\tor{t}$, so that $\qsim{t+1} = \qreal{t+1}$.
Using \cref{eq:int_dotqtp1,eq:int_qtp1,eq:fwd_dyn}, we obtain

\begin{align}
    \qreal{t+1} &= \qsim{t+1} \\
    &= \qsim{t} + \dt\,\left(\qdsim{t} + \dt\,\qddsim{t}\right) \\
    \begin{split}
        &=\qsim{t} + \dt\,\Big(\qdsim{t} + \dt \Minv{\qsim{t}}\Big(\tor{t} \\
        &\qquad\qquad\qquad- \cor{\qsim{t}}{\qdsim{t}} - \grav{\qsim{t}}\Big)\Big)
    \end{split} \\
    \begin{split}
        &=\qsim{t} + \dt\,\qdsim{t} + \dt^2 \Minv{\qsim{t}}\Big(\tor{t} \\
        &\qquad\qquad\qquad- \cor{\qsim{t}}{\qdsim{t}} - \grav{\qsim{t}}\Big).
    \end{split}
\end{align}

\noindent Rearrange the equation to

\begin{align}
    \begin{split}
    \tor{t} &= \frac{1}{\dt^2} \M{\qsim{t}}\mleft(\qreal{t+1} - \qsim{t} - \dt\qdsim{t}\mright) \\
    &\qquad+ \cor{\qsim{t}}{\qdsim{t}} + \grav{\qsim{t}}.
    \end{split}
\end{align}

\noindent By rearranging \cref{eq:int_qtp1}, we obtain
\begin{align}
     \qdsim{t} = \frac{\qsim{t} - \qsim{t-1}}{\dt},\label{eq:int_qtp1_rearranged}
\end{align}

\noindent which we use to compute

\begin{align}
    \begin{split}
        \tor{t} &= \frac{1}{\dt^2} \M{\qsim{t}}\mleft(\qreal{t + 1} - \qsim{t} - \dt\frac{\qsim{t} - \qsim{t-1}}{\dt}\mright) \\
        &\qquad + \cor{\qsim{t}}{\frac{\qsim{t} - \qsim{t-1}}{\dt}} + \grav{\qsim{t}}
    \end{split} \\
    \begin{split}
        &= \M{\qsim{t}}\left(\frac{\qreal{t + 1} - 2\qsim{t} + \qsim{t-1}}{\dt^2} \right) \\
        &\qquad + \cor{\qsim{t}}{\frac{\qsim{t} - \qsim{t-1}}{\dt}} + \grav{\qsim{t}}.
    \end{split}
\end{align}

\noindent With $\qsim{t} = \qreal{t}$, we obtain

\begin{align}
    \begin{split}
        \tor{t} &= \M{\qreal{t}}\mleft(\frac{\qreal{t + 1} - 2\qreal{t} + \qreal{t-1}}{\dt^2}\mright) \\
        &\qquad + \cor{\qreal{t}}{\frac{\qreal{t} - \qreal{t-1}}{\dt}} + \grav{\qreal{t}} 
    \end{split}\\
    &= \invdyn\mleft(\qreal{t}, \qdreal{t}, \qddreal{t} \mright)
\end{align}

\noindent for 

\begin{align}
    \qdreal{t} = \frac{\qreal{t} - \qreal{t-1}}{\dt} \label{eq:vel_backward}
\end{align}

\noindent and 

\begin{align}
    \qddreal{t} = \frac{\qreal{t + 1} - 2\qreal{t} + \qreal{t-1}}{\dt^2} \label{eq:acc_central}.
\end{align}

\Cref{eq:vel_backward,eq:acc_central} are the first-order backward and second-order central differences equations.

\section{Effect of the history design choices on the simulation accuracy}
\label{app:history}

The history input is an important component of the actuator modeling since actuators like PAMs are prone to hysteresis effects.
Therefore, in this section, we evaluate the impact of the history length $H$ and stride $s$ for network inputs of the form $(\qreal{t}, \qreal{t-s}, \ldots, \qreal{t-s \cdot H}, \ctrl{t}, \ctrl{t-s}, \ldots, \ctrl{t-s \cdot H})$.

\begin{figure}[ht]
        \begin{subfigure}{0.495\linewidth}
        \centering
        \mbox{
\begin{tikzpicture}

\definecolor{darkgoldenrod17815132}{RGB}{178,151,32}
\definecolor{darkgray141160203}{RGB}{141,160,203}
\definecolor{darkgray176}{RGB}{176,176,176}
\definecolor{dimgray}{RGB}{105,105,105}
\definecolor{gold25521747}{RGB}{255,217,47}
\definecolor{gray16196136}{RGB}{161,96,136}
\definecolor{mediumaquamarine102194165}{RGB}{102,194,165}
\definecolor{olivedrab11615158}{RGB}{116,151,58}
\definecolor{orchid231138195}{RGB}{231,138,195}
\definecolor{salmon25214198}{RGB}{252,141,98}
\definecolor{seagreen71135115}{RGB}{71,135,115}
\definecolor{sienna1769868}{RGB}{176,98,68}
\definecolor{slategray98111142}{RGB}{98,111,142}
\definecolor{yellowgreen16621684}{RGB}{166,216,84}

\begin{axis}[
golden width landscape=0.8\linewidth,tick align=outside,
tick pos=left,
x grid style={darkgray176},
xlabel={history length \(\displaystyle H\)},
xmin=-0.635, xmax=5.635,
xtick style={color=black,draw=none},
xtick={0,1,2,3,4,5},
xticklabels={1,2,3,5,10,20},
xticklabel style={yshift=3pt},
y grid style={darkgray176},
ylabel={position error (°)},
ymajorgrids,
ymin=0, ymax=0.00984641715048088,
ytick style={color=black}
]
\draw[draw=sienna1769868,fill=salmon25214198,very thick] (axis cs:-0.35,0) rectangle (axis cs:0.35,0.0089833195944761);
\draw[draw=slategray98111142,fill=darkgray141160203,very thick] (axis cs:0.65,0) rectangle (axis cs:1.35,0.00750341386922091);
\draw[draw=seagreen71135115,fill=mediumaquamarine102194165,very thick] (axis cs:1.65,0) rectangle (axis cs:2.35,0.00677644853792893);
\draw[draw=gray16196136,fill=orchid231138195,very thick] (axis cs:2.65,0) rectangle (axis cs:3.35,0.00639357524704128);
\draw[draw=olivedrab11615158,fill=yellowgreen16621684,very thick] (axis cs:3.65,0) rectangle (axis cs:4.35,0.00636350849084566);
\draw[draw=darkgoldenrod17815132,fill=gold25521747,very thick] (axis cs:4.65,0) rectangle (axis cs:5.35,0.00677571774437333);
\path [draw=dimgray, very thick]
(axis cs:0,0.00861028460431167)
--(axis cs:0,0.00937754014331512);

\path [draw=dimgray, very thick]
(axis cs:1,0.00716959674209612)
--(axis cs:1,0.00785020897692617);

\path [draw=dimgray, very thick]
(axis cs:2,0.00649194078708402)
--(axis cs:2,0.00708600954377315);

\path [draw=dimgray, very thick]
(axis cs:3,0.00613607249620321)
--(axis cs:3,0.00667335695240922);

\path [draw=dimgray, very thick]
(axis cs:4,0.00610053180855054)
--(axis cs:4,0.00664160875410119);

\path [draw=dimgray, very thick]
(axis cs:5,0.00651817476349199)
--(axis cs:5,0.00706302130064376);

\addplot [semithick, dimgray, mark=-, mark size=3, mark options={solid}, only marks]
table {%
0 0.00861028460431167
1 0.00716959674209612
2 0.00649194078708402
3 0.00613607249620321
4 0.00610053180855054
5 0.00651817476349199
};
\addplot [semithick, dimgray, mark=-, mark size=3, mark options={solid}, only marks]
table {%
0 0.00937754014331512
1 0.00785020897692617
2 0.00708600954377315
3 0.00667335695240922
4 0.00664160875410119
5 0.00706302130064376
};
\end{axis}

\end{tikzpicture}}
        \caption{Error after 1 step / $\SI{2}{\ms}$}
    \end{subfigure}
    \begin{subfigure}{0.495\linewidth}
        \centering
        \mbox{
\begin{tikzpicture}

\definecolor{darkgoldenrod17815132}{RGB}{178,151,32}
\definecolor{darkgray141160203}{RGB}{141,160,203}
\definecolor{darkgray176}{RGB}{176,176,176}
\definecolor{dimgray}{RGB}{105,105,105}
\definecolor{gold25521747}{RGB}{255,217,47}
\definecolor{gray16196136}{RGB}{161,96,136}
\definecolor{mediumaquamarine102194165}{RGB}{102,194,165}
\definecolor{olivedrab11615158}{RGB}{116,151,58}
\definecolor{orchid231138195}{RGB}{231,138,195}
\definecolor{salmon25214198}{RGB}{252,141,98}
\definecolor{seagreen71135115}{RGB}{71,135,115}
\definecolor{sienna1769868}{RGB}{176,98,68}
\definecolor{slategray98111142}{RGB}{98,111,142}
\definecolor{yellowgreen16621684}{RGB}{166,216,84}

\begin{axis}[
golden width landscape=0.8\linewidth,tick align=outside,
tick pos=left,
x grid style={darkgray176},
xlabel={history length \(\displaystyle H\)},
xmin=-0.635, xmax=5.635,
xtick style={color=black,draw=none},
xtick={0,1,2,3,4,5},
xticklabels={1,2,3,5,10,20},
xticklabel style={yshift=3pt},
y grid style={darkgray176},
ylabel={},
ymajorgrids,
ymin=0, ymax=5.81782323948898,
ytick style={color=black}
]
\draw[draw=sienna1769868,fill=salmon25214198,very thick] (axis cs:-0.35,0) rectangle (axis cs:0.35,5.35565804991443);
\draw[draw=slategray98111142,fill=darkgray141160203,very thick] (axis cs:0.65,0) rectangle (axis cs:1.35,4.68406347807344);
\draw[draw=seagreen71135115,fill=mediumaquamarine102194165,very thick] (axis cs:1.65,0) rectangle (axis cs:2.35,4.44238790646304);
\draw[draw=gray16196136,fill=orchid231138195,very thick] (axis cs:2.65,0) rectangle (axis cs:3.35,4.20375048837283);
\draw[draw=olivedrab11615158,fill=yellowgreen16621684,very thick] (axis cs:3.65,0) rectangle (axis cs:4.35,3.88860854827611);
\draw[draw=darkgoldenrod17815132,fill=gold25521747,very thick] (axis cs:4.65,0) rectangle (axis cs:5.35,4.10474305437206);
\path [draw=dimgray, very thick]
(axis cs:0,5.18337995200525)
--(axis cs:0,5.54078403760855);

\path [draw=dimgray, very thick]
(axis cs:1,4.53859228693015)
--(axis cs:1,4.83839407713389);

\path [draw=dimgray, very thick]
(axis cs:2,4.30957979042524)
--(axis cs:2,4.58608458679169);

\path [draw=dimgray, very thick]
(axis cs:3,4.069590721767)
--(axis cs:3,4.3423827506102);

\path [draw=dimgray, very thick]
(axis cs:4,3.76345930864467)
--(axis cs:4,4.02458051077812);

\path [draw=dimgray, very thick]
(axis cs:5,3.96771033122776)
--(axis cs:5,4.25873378274922);

\addplot [semithick, dimgray, mark=-, mark size=3, mark options={solid}, only marks]
table {%
0 5.18337995200525
1 4.53859228693015
2 4.30957979042524
3 4.069590721767
4 3.76345930864467
5 3.96771033122776
};
\addplot [semithick, dimgray, mark=-, mark size=3, mark options={solid}, only marks]
table {%
0 5.54078403760855
1 4.83839407713389
2 4.58608458679169
3 4.3423827506102
4 4.02458051077812
5 4.25873378274922
};
\end{axis}

\end{tikzpicture}}
        \caption{Error after 500 steps / $\SI{1}{\s}$}
    \end{subfigure}
    \caption{
        Position error for GenANs trained with different history lengths $H$, measured as the mean absolute error between 800 simulated trajectories and real robot trajectories (lower is better).
        Stride length is set to $s=1$ for all configurations.
        The error bars visualize the 95\% confidence intervals across samples computed via bootstrapping.
        Up to $H=10$, GenANs with longer histories tend to perform better.
        There is a large difference between the accuracy for $H=1$ and $H=2$ with progressively smaller gains beyond $H=2$.
    }
    \label{fig:accuracy_history_length}
\end{figure}

\begin{figure}[ht]
    \begin{subfigure}{0.495\linewidth}
        \centering
        \mbox{
\begin{tikzpicture}

\definecolor{darkgray141160203}{RGB}{141,160,203}
\definecolor{darkgray176}{RGB}{176,176,176}
\definecolor{dimgray}{RGB}{105,105,105}
\definecolor{gray16196136}{RGB}{161,96,136}
\definecolor{mediumaquamarine102194165}{RGB}{102,194,165}
\definecolor{olivedrab11615158}{RGB}{116,151,58}
\definecolor{orchid231138195}{RGB}{231,138,195}
\definecolor{salmon25214198}{RGB}{252,141,98}
\definecolor{seagreen71135115}{RGB}{71,135,115}
\definecolor{sienna1769868}{RGB}{176,98,68}
\definecolor{slategray98111142}{RGB}{98,111,142}
\definecolor{yellowgreen16621684}{RGB}{166,216,84}

\begin{axis}[
golden width landscape=0.8\linewidth,tick align=outside,
tick pos=left,
x grid style={darkgray176},
xlabel={history stride \(\displaystyle s\)},
xmin=-0.585, xmax=4.585,
xtick style={color=black,draw=none},
xtick={0,1,2,3,4},
xticklabels={1,2,3,4,5},
xticklabel style={yshift=3pt},
y grid style={darkgray176},
ylabel={position error (°)},
ymajorgrids,
ymin=0, ymax=0.012343341213485,
ytick style={color=black}
]
\draw[draw=seagreen71135115,fill=mediumaquamarine102194165,very thick] (axis cs:-0.35,0) rectangle (axis cs:0.35,0.00677644853792893);
\draw[draw=sienna1769868,fill=salmon25214198,very thick] (axis cs:0.65,0) rectangle (axis cs:1.35,0.00861432346039333);
\draw[draw=slategray98111142,fill=darkgray141160203,very thick] (axis cs:1.65,0) rectangle (axis cs:2.35,0.00978399002070507);
\draw[draw=gray16196136,fill=orchid231138195,very thick] (axis cs:2.65,0) rectangle (axis cs:3.35,0.0109220102601993);
\draw[draw=olivedrab11615158,fill=yellowgreen16621684,very thick] (axis cs:3.65,0) rectangle (axis cs:4.35,0.0108745366101564);
\path [draw=dimgray, very thick]
(axis cs:0,0.00649194078708402)
--(axis cs:0,0.00708600954377315);

\path [draw=dimgray, very thick]
(axis cs:1,0.00829204337428097)
--(axis cs:1,0.00895693309563783);

\path [draw=dimgray, very thick]
(axis cs:2,0.00937951748375443)
--(axis cs:2,0.0101879754164144);

\path [draw=dimgray, very thick]
(axis cs:3,0.0104376221900259)
--(axis cs:3,0.0117555630604619);

\path [draw=dimgray, very thick]
(axis cs:4,0.010433319391642)
--(axis cs:4,0.0113390862379789);

\addplot [semithick, dimgray, mark=-, mark size=3, mark options={solid}, only marks]
table {%
0 0.00649194078708402
1 0.00829204337428097
2 0.00937951748375443
3 0.0104376221900259
4 0.010433319391642
};
\addplot [semithick, dimgray, mark=-, mark size=3, mark options={solid}, only marks]
table {%
0 0.00708600954377315
1 0.00895693309563783
2 0.0101879754164144
3 0.0117555630604619
4 0.0113390862379789
};
\end{axis}

\end{tikzpicture}}
        \caption{Error after 1 step / $\SI{2}{\ms}$}
    \end{subfigure}
    \begin{subfigure}{0.495\linewidth}
        \centering
        \mbox{
\begin{tikzpicture}

\definecolor{darkgray141160203}{RGB}{141,160,203}
\definecolor{darkgray176}{RGB}{176,176,176}
\definecolor{dimgray}{RGB}{105,105,105}
\definecolor{gray16196136}{RGB}{161,96,136}
\definecolor{mediumaquamarine102194165}{RGB}{102,194,165}
\definecolor{olivedrab11615158}{RGB}{116,151,58}
\definecolor{orchid231138195}{RGB}{231,138,195}
\definecolor{salmon25214198}{RGB}{252,141,98}
\definecolor{seagreen71135115}{RGB}{71,135,115}
\definecolor{sienna1769868}{RGB}{176,98,68}
\definecolor{slategray98111142}{RGB}{98,111,142}
\definecolor{yellowgreen16621684}{RGB}{166,216,84}

\begin{axis}[
golden width landscape=0.8\linewidth,tick align=outside,
tick pos=left,
x grid style={darkgray176},
xlabel={history stride \(\displaystyle s\)},
xmin=-0.585, xmax=4.585,
xtick style={color=black,draw=none},
xtick={0,1,2,3,4},
xticklabels={1,2,3,4,5},
xticklabel style={yshift=3pt},
y grid style={darkgray176},
ylabel={},
ymajorgrids,
ymin=0, ymax=18.4562492742815,
ytick style={color=black}
]
\draw[draw=seagreen71135115,fill=mediumaquamarine102194165,very thick] (axis cs:-0.35,0) rectangle (axis cs:0.35,4.44238790646304);
\draw[draw=sienna1769868,fill=salmon25214198,very thick] (axis cs:0.65,0) rectangle (axis cs:1.35,4.00769716532355);
\draw[draw=slategray98111142,fill=darkgray141160203,very thick] (axis cs:1.65,0) rectangle (axis cs:2.35,6.32256737646524);
\draw[draw=gray16196136,fill=orchid231138195,very thick] (axis cs:2.65,0) rectangle (axis cs:3.35,15.5045362461229);
\draw[draw=olivedrab11615158,fill=yellowgreen16621684,very thick] (axis cs:3.65,0) rectangle (axis cs:4.35,17.0833754539069);
\path [draw=dimgray, very thick]
(axis cs:0,4.30957979042524)
--(axis cs:0,4.58608458679169);

\path [draw=dimgray, very thick]
(axis cs:1,3.88565551474368)
--(axis cs:1,4.13772768293022);

\path [draw=dimgray, very thick]
(axis cs:2,6.00277339598085)
--(axis cs:2,6.75563340798372);

\path [draw=dimgray, very thick]
(axis cs:3,15.0300945888226)
--(axis cs:3,16.0310723100882);

\path [draw=dimgray, very thick]
(axis cs:4,16.6485028767495)
--(axis cs:4,17.5773802612205);

\addplot [semithick, dimgray, mark=-, mark size=3, mark options={solid}, only marks]
table {%
0 4.30957979042524
1 3.88565551474368
2 6.00277339598085
3 15.0300945888226
4 16.6485028767495
};
\addplot [semithick, dimgray, mark=-, mark size=3, mark options={solid}, only marks]
table {%
0 4.58608458679169
1 4.13772768293022
2 6.75563340798372
3 16.0310723100882
4 17.5773802612205
};
\end{axis}

\end{tikzpicture}}
        \caption{Error after 500 steps / $\SI{1}{\s}$}
    \end{subfigure}
    \caption{
        Position error for GenANs trained with different stride lengths $s$, measured as the mean absolute error between 800 simulated trajectories and real robot trajectories (lower is better).
        History length is set to $H=3$ for all configurations.
        The error bars visualize the 95\% confidence intervals across samples computed via bootstrapping.
        Generally, shorter strides seem to work best.
        Only for the multi-step error does $s=2$ yield a slightly lower error than $s=1$. 
    }
    \label{fig:accuracy_stride_length}
\end{figure}

\Cref{fig:accuracy_history_length} shows the simulator position error for GenANs trained with different history lengths.
Generally, longer histories result in lower errors in both the single-step and 500-step position error metrics.
There is a strong improvement for $H=2$ over $H=1$, indicating that one-step histories are insufficient to model the complex dynamics of PAMs.
Beyond $H=2$, the gains from longer histories seem to get gradually smaller.
Histories of length $H=20$ seem to be too long, resulting in a slight performance decrease.
For all other experiments, we use $H=3$ as a trade-off between accuracy and computational efficiency.

In the original actuator networks paper~\cite{hwangbo2019learning}, the authors propose to use a sparse, i.e., strided, history, mentioning overfitting as a problem of dense histories.
\Cref{fig:accuracy_stride_length} shows the simulator accuracy for different stride lengths $s$.
Generally, shorter strides seem to perform best, with only $s=2$ leading to a slightly lower error than $s=1$ in the multi-step error metric.
In the single-step error metric, $s=1$ outperforms all longer strides.
As the experiment was conducted on an unseen test dataset, this result indicates that overfitting is not an issue for GenANs with short strides.

\section{Relation between the torque and position loss}
\label{app:relation_torque_position_loss}

Let $\qreal{t}$ for $t \in \{0, \ldots, T\}$ be a trajectory from the dataset and $\qdreal{t}$ and $\qddreal{t}$ the corresponding finite-difference velocities and accelerations, computed with \cref{eq:vel_backward,eq:acc_central} for $t \in \{1, \ldots, T-1\}$.
Furthermore, let $\boldsymbol{\hat{q}}_{t+1}$ be the next simulated position resulting from the dataset position $\qpred{t} = \qreal{t}$ and velocity $\qdpred{t} = \qdreal{t}$ after applying the torque $\boldsymbol{\hat{\tau}}_{t}$ predicted by the GenAN for $t \in \{1, \ldots, T-1\}$.
Starting from the position error $\boldsymbol{q}_{t+1} - \boldsymbol{\hat{q}}_{t+1}$, we derive \cref{eq:relation_torque_position_loss} by utilizing the integration step for the position and velocity from \cref{eq:int_dotqtp1,eq:int_qtp1}.

Analogous to \cref{eq:int_dotqtp1,eq:int_qtp1}, we obtain the integration step
\begin{align}
    \qdreal{t+1} &= \qdreal{t} + \dt\,\qddreal{t} \label{eq:int_dotqtp1_real} \\[0.5\baselineskip]
    \qreal{t+1} &= \qreal{t} + \dt\,\qdreal{t+1} \label{eq:int_qtp1_real}
\end{align}
and with $\qpred{t} = \qreal{t}$ and $\qdpred{t} = \qdreal{t}$, we get
\begin{align}
    \qdpred{t+1} &= \qdpred{t} + \dt\,\qddpred{t} \\
    &= \qdreal{t} + \dt\,\qddpred{t} \label{eq:int_dotqtp1_pred} \\[0.5\baselineskip]
    \qpred{t+1} &= \qpred{t} + \dt\,\qdpred{t+1} \\
    &= \qreal{t} + \dt\,\qdpred{t+1}\label{eq:int_qtp1_pred}
\end{align}

Analogous to \cref{eq:fwd_dyn}, the forward dynamics for these two cases are given by
\begin{align}
    \qddreal{t} &= \Minv{\qreal{t}} \left(\tor{t}- \cor{\qreal{t}}{\qdreal{t}} - \grav{\qreal{t}}\right) \label{eq:fwd_dyn_real} \\[0.5\baselineskip]
    \qddpred{t} &= \Minv{\qpred{t}} \left(\torpred{t}- \cor{\qpred{t}}{\qdpred{t}} - \grav{\qpred{t}}\right) \\
    &= \Minv{\qreal{t}} \left(\torpred{t}- \cor{\qreal{t}}{\qdreal{t}} - \grav{\qreal{t}}\right). \label{eq:fwd_dyn_pred}
\end{align}

By first inserting \cref{eq:int_qtp1_real,eq:int_qtp1_pred}, then \cref{eq:int_dotqtp1_real,eq:int_dotqtp1_pred}, and finally \cref{eq:fwd_dyn_real,eq:fwd_dyn_pred}, we obtain \cref{eq:relation_torque_position_loss}.
\begin{align*}
    &\boldsymbol{q}_{t+1} - \boldsymbol{\hat{q}}_{t+1} \\
    &~~= \boldsymbol{q}_t + \Delta t\,\boldsymbol{\dot{q}}_{t+1} - \boldsymbol{q}_t - \Delta t \boldsymbol{\dot{\hat{q}}}_{t+1} \\
    &~~= \Delta t (\boldsymbol{\dot{q}}_{t+1} - \boldsymbol{\dot{\hat{q}}}_{t+1}) \\
    &~~= \Delta t (\boldsymbol{\dot{q}}_t + \Delta t\,\boldsymbol{\ddot{q}}_t - \boldsymbol{\dot{q}}_t - \Delta t\,\boldsymbol{\ddot{\hat{q}}}_t) \\
    &~~= \Delta t^2 (\boldsymbol{\ddot{q}}_t - \boldsymbol{\ddot{\hat{q}}}_t) \\
    &~~= \Delta t^2 \big(\boldsymbol{M}(\boldsymbol{q}_t)^{-1} \left(\boldsymbol{\tau}_t - c(\boldsymbol{q}_t, \boldsymbol{\dot{q}}_t) - g(\boldsymbol{q}_t)\right) \\
    &\qquad\qquad~ - \boldsymbol{M}(\boldsymbol{q}_t)^{-1} \left(\boldsymbol{\hat{\tau}}_t- c(\boldsymbol{q}_t, \boldsymbol{\dot{q}}_t) - g(\boldsymbol{q}_t)\right)\big) \\
    &~~= \Delta t^2 \boldsymbol{M}(\boldsymbol{q}_t)^{-1} (\boldsymbol{\tau}_t - \boldsymbol{\hat{\tau}}_t)
\end{align*}

\section{Multi-step position loss}
\label{app:multi_step_loss}

In \cref{subsec:actuator_network_training}, we define a single-step position loss for training the GenAN.
A natural extension is to roll out the simulator for $R$ steps, starting from some step $t \in \{H, \ldots, T - R\}$.
We define $\qpred{t \rightarrow r}$ as notation for simulating $r$ steps from $\qreal{t}$ into the future.
For ease of notation, we define $\qpred{t \rightarrow r} = \qreal{t + r}$ and $\qdpred{t \rightarrow r} = \qdreal{t + r}$ for all $r \in \{-H, \ldots, 0\}$.
We then simulate
\begin{align}
    \qpred{t \rightarrow r + 1} &= \step(\qpred{t \rightarrow r}, \qdpred{t \rightarrow r}, \torpred{t \rightarrow r}) \\[0.3\baselineskip]
    \qdpred{t \rightarrow r +1} &= \frac{\qpred{t \rightarrow r + 1} - \qpred{t \rightarrow r}}{\Delta t} \\[0.3\baselineskip]
    \torpred{t \rightarrow r} &= f_{\boldsymbol{\theta}}\mleft(\qpred{t \rightarrow r-H:r}, \ctrl{t+r-H:t+r}\mright) 
\end{align} 
for all $r \in \{0, \ldots, R - 1\}$, where we define $\qpred{t \rightarrow r-H:r} = (\qpred{t \rightarrow r-H}, \qpred{t \rightarrow r-H+1}, \ldots,\qpred{t \rightarrow r})$.
By adding the losses for the individual simulation steps, we obtain the following \emph{multi-step position loss}
\begin{equation}
    \mathcal{L}_\mathrm{mul}(\boldsymbol{\theta}) = \frac{1}{R}\sum_{r=1}^{R} \left\lVert\frac{\qpred{t \rightarrow r} - \qreal{t + r}}{\mathbf{c}_{r}}\right\rVert^2, \label{eq:multi_step_loss}
\end{equation}
where the division is elementwise and $\mathbf{c}_{r}$ is a normalization constant, described below.

The normalization is required because position errors at different rollout steps can differ substantially in magnitude, which results in the network ignoring errors early in the rollouts in favor of reducing the later errors.
For the normalization, we first compute positions $\qsimzero{t \rightarrow r}$ resulting from applying constant zero torques for each time step of the rollout by setting $\qsimzero{t \rightarrow 0} = \qreal{t}$ and $\qdsimzero{t \rightarrow 0} = \qdreal{t}$ and simulating 
\begin{align}
    \qsimzero{t \rightarrow r + 1} &= \step(\qsimzero{t \rightarrow r}, \qdsimzero{t \rightarrow r}, \boldsymbol{0}) \\[0.2\baselineskip]
    \qdsimzero{t \rightarrow r + 1} &= \frac{\qsimzero{t \rightarrow r + 1} - \qsimzero{t \rightarrow r}}{\Delta t}
\end{align} for all $r \in \{0, \ldots, R - 1\}$.
The normalization constant is then computed by calculating the absolute position error per joint, averaged across all possible $r$-step rollouts in the training dataset.
\begin{equation}
    \boldsymbol{c}^{(j)}_{r} = \mathbb{E}\left[\left\lvert\qsimzero{t \rightarrow r}^{(j)} - \qreal{t + r}^{(j)}\right\rvert\right]
\end{equation}
The notation $\boldsymbol{x}^{(j)}$ denotes selecting the value for the $j$th joint from the vector.

This normalization essentially compares the error of the network with that of the simplest possible, i.e., zero-torque, model.
Naturally, the errors of this baseline tend to increase over the rollout horizon.
Dividing by these errors, therefore, puts more weight on predictions early in the rollouts and mitigates the effects of the error magnitude imbalance described above.

\begin{figure}[ht]
        \begin{subfigure}{0.495\linewidth}
        \centering
        \mbox{
\begin{tikzpicture}

\definecolor{darkgoldenrod17815132}{RGB}{178,151,32}
\definecolor{darkgray141160203}{RGB}{141,160,203}
\definecolor{darkgray176}{RGB}{176,176,176}
\definecolor{dimgray}{RGB}{105,105,105}
\definecolor{gold25521747}{RGB}{255,217,47}
\definecolor{gray16196136}{RGB}{161,96,136}
\definecolor{mediumaquamarine102194165}{RGB}{102,194,165}
\definecolor{olivedrab11615158}{RGB}{116,151,58}
\definecolor{orchid231138195}{RGB}{231,138,195}
\definecolor{salmon25214198}{RGB}{252,141,98}
\definecolor{seagreen71135115}{RGB}{71,135,115}
\definecolor{sienna1769868}{RGB}{176,98,68}
\definecolor{slategray98111142}{RGB}{98,111,142}
\definecolor{yellowgreen16621684}{RGB}{166,216,84}

\begin{axis}[
golden width landscape=0.8\linewidth,tick align=outside,
tick pos=left,
x grid style={darkgray176},
xmin=-0.635, xmax=5.635,
xtick style={color=black,draw=none},
xtick={0,1,2,3,4,5},
xticklabels={1,3,5,10,20,30},
xticklabel style={yshift=3pt},
y grid style={darkgray176},
xlabel={rollout length $R$},
ylabel={position error (°)},
ymajorgrids,
ymin=0, ymax=0.00889401954885113,
ytick style={color=black}
]
\draw[draw=seagreen71135115,fill=mediumaquamarine102194165,very thick] (axis cs:-0.35,0) rectangle (axis cs:0.35,0.00677644853792893);
\draw[draw=sienna1769868,fill=salmon25214198,very thick] (axis cs:0.65,0) rectangle (axis cs:1.35,0.0067174526955966);
\draw[draw=slategray98111142,fill=darkgray141160203,very thick] (axis cs:1.65,0) rectangle (axis cs:2.35,0.0069152463695227);
\draw[draw=gray16196136,fill=orchid231138195,very thick] (axis cs:2.65,0) rectangle (axis cs:3.35,0.00731241369887846);
\draw[draw=olivedrab11615158,fill=yellowgreen16621684,very thick] (axis cs:3.65,0) rectangle (axis cs:4.35,0.00764521783448808);
\draw[draw=darkgoldenrod17815132,fill=gold25521747,very thick] (axis cs:4.65,0) rectangle (axis cs:5.35,0.0081498539215658);
\path [draw=dimgray, very thick]
(axis cs:0,0.00649194078708402)
--(axis cs:0,0.00708600954377315);

\path [draw=dimgray, very thick]
(axis cs:1,0.006442750871027)
--(axis cs:1,0.00700240909918281);

\path [draw=dimgray, very thick]
(axis cs:2,0.00664188927372681)
--(axis cs:2,0.00719685433479995);

\path [draw=dimgray, very thick]
(axis cs:3,0.00703422303364536)
--(axis cs:3,0.00762279428062108);

\path [draw=dimgray, very thick]
(axis cs:4,0.00736699716589334)
--(axis cs:4,0.00794761419594215);

\path [draw=dimgray, very thick]
(axis cs:5,0.00785275915623239)
--(axis cs:5,0.00847049480842964);

\addplot [semithick, dimgray, mark=-, mark size=3, mark options={solid}, only marks]
table {%
0 0.00649194078708402
1 0.006442750871027
2 0.00664188927372681
3 0.00703422303364536
4 0.00736699716589334
5 0.00785275915623239
};
\addplot [semithick, dimgray, mark=-, mark size=3, mark options={solid}, only marks]
table {%
0 0.00708600954377315
1 0.00700240909918281
2 0.00719685433479995
3 0.00762279428062108
4 0.00794761419594215
5 0.00847049480842964
};
\end{axis}

\end{tikzpicture}}
        \caption{Error after 1 step / $\SI{2}{\ms}$}
    \end{subfigure}
    \begin{subfigure}{0.495\linewidth}
        \centering
        \mbox{
\begin{tikzpicture}

\definecolor{darkgoldenrod17815132}{RGB}{178,151,32}
\definecolor{darkgray141160203}{RGB}{141,160,203}
\definecolor{darkgray176}{RGB}{176,176,176}
\definecolor{dimgray}{RGB}{105,105,105}
\definecolor{gold25521747}{RGB}{255,217,47}
\definecolor{gray16196136}{RGB}{161,96,136}
\definecolor{mediumaquamarine102194165}{RGB}{102,194,165}
\definecolor{olivedrab11615158}{RGB}{116,151,58}
\definecolor{orchid231138195}{RGB}{231,138,195}
\definecolor{salmon25214198}{RGB}{252,141,98}
\definecolor{seagreen71135115}{RGB}{71,135,115}
\definecolor{sienna1769868}{RGB}{176,98,68}
\definecolor{slategray98111142}{RGB}{98,111,142}
\definecolor{yellowgreen16621684}{RGB}{166,216,84}

\begin{axis}[
golden width landscape=0.8\linewidth,tick align=outside,
tick pos=left,
x grid style={darkgray176},
xmin=-0.635, xmax=5.635,
xtick style={color=black,draw=none},
xtick={0,1,2,3,4,5},
xticklabels={1,3,5,10,20,30},
xticklabel style={yshift=3pt},
y grid style={darkgray176},
xlabel={rollout length $R$},
ymajorgrids,
ymin=0, ymax=4.84073706257572,
ytick style={color=black}
]
\draw[draw=seagreen71135115,fill=mediumaquamarine102194165,very thick] (axis cs:-0.35,0) rectangle (axis cs:0.35,4.44238790646304);
\draw[draw=sienna1769868,fill=salmon25214198,very thick] (axis cs:0.65,0) rectangle (axis cs:1.35,4.06942030419251);
\draw[draw=slategray98111142,fill=darkgray141160203,very thick] (axis cs:1.65,0) rectangle (axis cs:2.35,4.05044731933867);
\draw[draw=gray16196136,fill=orchid231138195,very thick] (axis cs:2.65,0) rectangle (axis cs:3.35,4.39673065318243);
\draw[draw=olivedrab11615158,fill=yellowgreen16621684,very thick] (axis cs:3.65,0) rectangle (axis cs:4.35,4.2246401950625);
\draw[draw=darkgoldenrod17815132,fill=gold25521747,very thick] (axis cs:4.65,0) rectangle (axis cs:5.35,3.87830520791314);
\path [draw=dimgray, very thick]
(axis cs:0,4.30957979042524)
--(axis cs:0,4.58608458679169);

\path [draw=dimgray, very thick]
(axis cs:1,3.93830922296619)
--(axis cs:1,4.21886242817578);

\path [draw=dimgray, very thick]
(axis cs:2,3.92389616483405)
--(axis cs:2,4.18844212876546);

\path [draw=dimgray, very thick]
(axis cs:3,4.23793429725355)
--(axis cs:3,4.61022577388164);

\path [draw=dimgray, very thick]
(axis cs:4,4.07814312148243)
--(axis cs:4,4.38089577524741);

\path [draw=dimgray, very thick]
(axis cs:5,3.75502250220657)
--(axis cs:5,4.00344033226554);

\addplot [semithick, dimgray, mark=-, mark size=3, mark options={solid}, only marks]
table {%
0 4.30957979042524
1 3.93830922296619
2 3.92389616483405
3 4.23793429725355
4 4.07814312148243
5 3.75502250220657
};
\addplot [semithick, dimgray, mark=-, mark size=3, mark options={solid}, only marks]
table {%
0 4.58608458679169
1 4.21886242817578
2 4.18844212876546
3 4.61022577388164
4 4.38089577524741
5 4.00344033226554
};
\end{axis}

\end{tikzpicture}}
        \caption{Error after 500 steps / $\SI{1}{\s}$}
    \end{subfigure}
    \caption{
        Position error for GenANs trained with the position loss rolled out for different rollout lengths $R$, measured as the mean absolute error between 800 simulated trajectories and real robot trajectories (lower is better).
        In the single-step error, shorter rollouts perform best, while in the multi-step error, longer rollout lengths tend to perform best. 
    }
    \label{fig:accuracy_rollout_length}
\end{figure}

\Cref{fig:accuracy_rollout_length} compares the position accuracy of GenANs trained with the multi-step loss of \cref{eq:multi_step_loss} for different rollout lengths $R$.
For the multi-step error, the plots show a slight downward trend for models trained with longer rollouts.
For the single-step error, however, this trend reverses, and shorter rollout lengths result in higher accuracy.
Furthermore, the multi-step training is computationally significantly more demanding as it requires differentiating through the simulator for multiple steps, resulting in vastly longer training times.
While the single-step training completes in about 25 minutes, the multi-step training with $R=30$ takes about 12 hours to converge on an Nvidia A100 GPU.
Overall, the multi-step loss does not yield clear improvements, and we deem it not worth the additional computational cost.
Therefore, we use the single-step loss throughout the main text.

\section{Details on the system identification baseline}
\label{app:sysid}

As a physically interpretable baseline for the learned actuator network, we identify an analytical model of the pneumatic actuation.
Like GenAN, the analytical model maps the antagonistic control signals $\ctrl{}$ to joint torques $\tor{}$.

\subsection{Antagonistic actuation}
As mentioned in \cref{sec:experiments}, we actuate the two muscles of each joint in an antagonistic fashion with a one-dimensional control signal.
For each joint $j \in \{1, 2, 3, 4\}$, the cocontraction parameter $\nu^{(j)} \in [0,1]$ specifies the amount of cocontraction in the muscle pair, resulting in the following mapping from the control signal $u^{(j)}$ to the commanded pressure $\pdes{j,x}$ for muscle $x \in \{a, b\}$.
\begin{align}
  \pdes{j,\mathrm{a}} &= \mathrm{clip}(\bigl(\pmax{j,\mathrm{a}} - \pmin{j,\mathrm{a}}\bigr)\bigl(\nu^{(j)} - c^{(j)} u^{(j)}\bigr) + \pmin{j,\mathrm{a}}, \pmin{j,\mathrm{a}}, \pmax{j, \mathrm{a}}) \\
  \pdes{j,\mathrm{b}} &= \mathrm{clip}(\bigl(\pmax{j,\mathrm{b}} - \pmin{j,\mathrm{b}}\bigr)\bigl(\nu^{(j)} + c^{(j)} u^{(j)}\bigr) + \pmin{j,\mathrm{b}}, \pmin{j,\mathrm{b}}, \pmax{j, \mathrm{b}})
\end{align}
The commanded pressures are clipped to the pressure limits $\pmin{j,x}$ and $\pmax{j,x}$, and the correction $c^{(j)} = 0.5 + \operatorname{clip}\bigl(|\nu^{(j)} - 0.5|,\, 0,\, 0.5\bigr)$ ensures that at least one of the commanded pressures changes across the full command range.

\subsection{Pressure dynamics}
The commanded muscle pressures are realized by a low-level pressure controller.
As it takes time for the controller to achieve a desired pressure, there is a delay between a command and its realization.
We model this as a first-order lag with the time constant $T_p$.
\begin{align}
  \pobs{j,x}_t &= \sum_{i=0}^{N} w_i \, \pdes{j,x}(t - i\Delta t) \label{eq:pressure_lag} \\
  w_i &= \frac{e^{-i\Delta t/T_p}}{\sum_{k=0}^{N} e^{-k\Delta t /T_p}} \label{eq:pressure_lag_weight}
\end{align}
$\Delta t = \SI{2}{\milli\second}$ is the spacing of the command history of the last $N=20$ steps.

\subsection{Muscle-tendon kinematics}

Each muscle acts on its joint via a tendon routed over a pulley of radius $r^{(j,x)}$.
Furthermore, the tendon routing includes an initial slack offset $\delta^{(j,x)} > 0$. 
A muscle does not transmit force to the joint until the tendon becomes taut.
The relation between the muscle contraction ratio and the corresponding joint position $q^{(j)}$ is therefore
\begin{equation}
  h^{(j,x)} = \frac{1}{L_0}\left( \delta^{(j,x)} + \epsilon^{(x)} \, r^{(j,x)}\, q^{(j)} \right),
  \label{eq:contraction}
\end{equation}
where $L_0$ is the muscle length at rest and $\epsilon^{(\mathrm{a})} = -1$ and $\epsilon^{(\mathrm{b})} = 1$ encodes the muscle's pulling direction.
Note that $h^{(j,x)}$ specifies the ratio of how far the muscle has contracted, i.e., $h^{(j,x)} = 0$ at rest and $h^{(j,x)} = h_{\max} \in (0, 1)$ at maximum contraction.

\subsection{Muscle force}

The static characteristic of a PAM is described by the Chou-Hannaford model~\cite{chou1996measurement} with an additional parameter $k_{\mathrm{el}}$ for the elasticity of the membrane
\begin{align}
  \mathcal{F}(\pobs{}, h) &= \max\Bigl(A\,\pobs{} \left( a\,\phi(h)^2 - b \right) - k_{\mathrm{el}}\, h,\; 0 \Bigr) \label{eq:muscle_force} \\
  \phi(h) &= \operatorname{clip}\bigl(1 - c\,h,\, 0,\, 1\bigr),
\end{align}
where $A = \pi D_0^2/4$ is the nominal cross-section, $a = 3\cot^2\alpha_0$ and $b = \csc^2\alpha_0$ follow from the braid angle $\alpha_0$ at rest, and $c$ corrects the effective diameter change.

The truncation at zero in \cref{eq:muscle_force} models the fact that tendons can only pull on the joint and not push.
Since the untruncated force expression decreases monotonically with $h$, it reaches zero at the pressure-dependent free-contraction ratio $h_0(\pobs{})$, i.e., the contraction of an unloaded muscle at pressure $\pobs{}$.
Thus, if $h \geq h_0(\pobs{})$, the joint position permits the muscle to reach its free-contraction length without tensioning the tendon, which means that the tendon is slack and transmits no force.

The manufacturer specifies a theoretical force $F_{\mathrm{nom}}$ at rest length and the maximum contraction $h_{\max}$, both at the maximum operating pressure $p_{\max,\mathrm{ds}}$.
By setting $c = 1$, i.e., no correction of the effective diameter change, we compute the braid angle and membrane stiffness from \cref{eq:muscle_force} with $\mathcal{F}(p_{\max,\mathrm{ds}}, 0) = F_{\mathrm{nom}}$ and $\mathcal{F}(p_{\max,\mathrm{ds}}, h_{\max}) = 0$.
\begin{align}
  \sin^2\alpha_0 &= \frac{2}{\dfrac{F_{\mathrm{nom}}}{A\,p_{\max,\mathrm{ds}}} + 3} \\
  k_{\mathrm{el}} &= \frac{A\,p_{\max,\mathrm{ds}}\left(a (1 - h_{\max})^2 - b\right)}{h_{\max}}
  \label{eq:datasheet_init}
\end{align}
For the Festo DMSP-20-1200N-AM-CM PAMs ($D_0 = \SI{20}{\milli\metre}$, $F_{\mathrm{nom}} = \SI{1500}{\newton}$, $h_{\max} = 0.25$, $p_{\max,\mathrm{ds}} = \SI{6}{\bar}$) this gives $\alpha_0 = \SI{25.3}{\degree}$, $a = 13.44$, $b = 5.48$ and $k_{\mathrm{el}} = \SI{1568}{\newton}$.
We use these parameter values as initialization for the system identification in \cref{subapp:sysid}.

\subsection{Joint torque}

The tendon tension is computed as $F^{(j,x)} = g^{(j,x)} \mathcal{F}\mleft(\pobs{j,x}, h^{(j,x)}\mleft(q^{(j)}\mright)\mright)$, where $g^{(j,x)}$ is a dimensionless per-muscle gain to correct the force scale of each muscle.
The resulting torque at joint $j$ is modeled as
\begin{equation}
  \tau^{(j)} = r^{(j,\mathrm{b})} F^{(j,\mathrm{b})} - r^{(j,\mathrm{a})} F^{(j,\mathrm{a})} + \tau_{\mathrm{f}}^{(j)} + \tau_0^{(j)},
\end{equation}
which consists of the contributions of the two antagonistic muscles, a friction torque $\tau_{\mathrm{f}}^{(j)}$ and a constant torque offset $\tau_0^{(j)}$ that models residual static bias.
\begin{equation}
  \tau_{\mathrm{f}}^{(j)} = -\tanh\!\left(\frac{\dot{q}^{(j)}}{\omega_s}\right) \Bigl( \tau_{\mathrm{c}}^{(j)} + \mu^{(j)} \sum_{x \in \{\mathrm{a}, \mathrm{b}\}} r^{(j,x)} F^{(j,x)} \Bigr) - d^{(j)}\, \dot{q}^{(j)}
  \label{eq:friction}
\end{equation}
The Coulomb term has a constant part $\tau_{\mathrm{c}}^{(j)}$ and a term proportional to the total tendon tension, scaled by the load-dependent friction coefficient $\mu^{(j)}$.
The factor $\tanh(\dot{q}^{(j)}/\omega_s)$ smooths the sign of the Coulomb term over a small velocity scale $\omega_s$, keeping~\eqref{eq:friction} differentiable.
$d^{(j)}$ is a viscous coefficient.

\subsection{System Identification}
\label{subapp:sysid}

\begin{table}[t]
  \centering
  \caption{System identification hyperparameters.}
  \label{tab:hyperparameters_sys_id}
  \begin{tabular}{llr}
    \toprule
    Stage & Setting & Value \\
    \midrule
    Stage 1                     & time constant grid                 & \SIrange{2}{50}{\milli\second} \\
    \midrule
    \multirow{8}{*}{Stage 2}    & rollout horizon $R$                & 300 steps (\SI{0.6}{\second}) \\
                                & optimizer                          & Adam \\
                                & optimizer steps                    & \num{3000} \\
                                & batch size                         & 128 \\
                                & initial learning rate              & \num{0.015} \\
                                & learning rate schedule             & cosine decay \\
                                & gradient clipping (global norm)    & \num{1.0} \\
                                & random restarts                    & 8 \\
    \bottomrule
  \end{tabular}
\end{table}

\begin{table}[t]
\centering
\caption{Parameters of the analytical muscle model.}
\label{tab:parameters_sys_id}
\begin{tabular}{clcl}
    \toprule
    Symbol                              & Description                               & Identification            & Value \\
    \midrule
    \multirow{2}{*}{$r^{(j,x)}$}        & \multirow{2}{*}{pulley radius}            & \multirow{2}{*}{--}       & $(38.5, 25.1, 23.5, 17.0)$ \si{\milli\metre} \\
                                        &                                           &                           & $(38.5, 25.1, 23.5, 17.0)$ \si{\milli\metre} \\  
    $L_0$                               & muscle length at rest                     & --                        & \SI{1.2}{\metre} \\
    $D_0$                               & muscle diameter at rest                   & --                        & \SI{20}{\milli\metre} \\
    \midrule
    $T_p$                              & pressure lag time                         & stage 1                   & \SI{50}{\milli\second} \\
    \midrule             
    $a$, $b$, $c$                       & braid geometry                            & stage 2                   & $15.6$, $6.39$, $2.61$ \\
    $k_{\mathrm{el}}$                   & membrane stiffness                        & stage 2                   & \SI{1475}{\newton} \\
    \multirow{2}{*}{$\delta^{(j,x)}$}   & \multirow{2}{*}{tendon slack}             & \multirow{2}{*}{stage 2}  & $(112, 80.2, 71.6, 88.5)$ \si{\milli\metre} \\
                                        &                                           &                           & $(127, 60.7, 74.1, 93.6)$ \si{\milli\metre} \\  
    \multirow{2}{*}{$g^{(j,x)}$}        & \multirow{2}{*}{per-muscle force gain}    & \multirow{2}{*}{stage 2}  & $(1.03, 1.97, 1.46, 0.904)$ \\
                                        &                                           &                           & $(1.95, 1.82, 1.47, 1.10)$ \\
    $\tau_{\mathrm{c}}^{(j)}$           & const.\ Coulomb friction                  & stage 2                   & $(0.0191, 0.361, 2.41, 0.118)$ \si{\newton\metre} \\    
    $\mu^{(j)}$                         & load-dependent friction                   & stage 2                   & $(0.380, 0.0173, 0.0293, 0.399)$ \\
    $d^{(j)}$                           & viscous friction                          & stage 2                   & $(0.198, 1.77, 0.448, 0.0539)$ \si{\newton\metre\second\per\radian} \\
    $\tau_{0}^{(j)}$                    & static torque offset                      & stage 2                   & $(0.126, -0.101, 0.159, 0.275)$ \si{\newton\metre} \\    
    \bottomrule
\end{tabular}
\end{table}

We identify the parameters in two stages.
In stage 1, we identify the lag in the commanded pressures from the measured pressure responses.
The pressure dynamics parameters are then fixed for stage 2, where we identify the remaining parameters. 

\paragraph{Stage 1: pressure dynamics.}
We identify the time constant $T_p$ from recorded pairs of commanded and measured muscle pressures by maximizing the explained variance of~\cref{eq:pressure_lag} over a grid, pooled across muscles.

\paragraph{Stage 2: torque response.}
All remaining parameters are identified from joint trajectories by minimizing a multi-step simulation error over a horizon $R$, akin to \cref{eq:multi_step_loss}.
\begin{equation}
    \mathcal{L}_\mathrm{sysid}(\boldsymbol{\theta}) = \frac{1}{R}\sum_{r=1}^{R} \left\lVert \qpred{t \rightarrow r} - \qreal{t + r}\right\rVert^2, \label{eq:sys_id_multi_step_loss}
\end{equation}
We refer to \cref{app:multi_step_loss} for details on this loss formulation.
Gradients are propagated through the differentiable simulator, and the model is optimized with Adam.
Positive parameters are optimized in logarithmic coordinates so that gradients are of comparable magnitude despite scales spanning several orders of magnitude.

\Cref{tab:hyperparameters_sys_id} lists all the hyperparameters of the system identification procedure, and \cref{tab:parameters_sys_id} displays the identified parameters.

\section{Additional transfer evaluation}
\label{app:transfer_results}

The following section extends the transfer evaluation from \cref{subsec:transfer} with task-specific metrics, a comparison with the results of~\citet{guist2024safe}, and a discussion of common failure modes.

\subsection{Reacher}
\label{subapp:accuracy_reacher}

\begin{figure}[ht]
    \centering
\begin{tikzpicture}

\definecolor{darkgray141160203}{RGB}{141,160,203}
\definecolor{darkgray176}{RGB}{176,176,176}
\definecolor{dimgray}{RGB}{105,105,105}
\definecolor{gray16196136}{RGB}{161,96,136}
\definecolor{mediumaquamarine102194165}{RGB}{102,194,165}
\definecolor{olivedrab11615158}{RGB}{116,151,58}
\definecolor{orchid231138195}{RGB}{231,138,195}
\definecolor{salmon25214198}{RGB}{252,141,98}
\definecolor{seagreen71135115}{RGB}{71,135,115}
\definecolor{sienna1769868}{RGB}{176,98,68}
\definecolor{slategray98111142}{RGB}{98,111,142}
\definecolor{yellowgreen16621684}{RGB}{166,216,84}

\begin{axis}[
golden width landscape=0.7\linewidth,scale only axis,tick align=outside,
tick pos=left,
x grid style={darkgray176},
xmin=-0.585, xmax=4.585,
xtick style={color=black,draw=none},
xtick={0,1,2,3,4},
xticklabel style={align=center,yshift=3pt},
xticklabels={GenAN,Torque\\GenAN,No disag.\\penalty,No\\ensemble,Low action\\penalty},
y grid style={darkgray176},
ylabel={distance to goal (°)},
ymajorgrids,
ymin=0, ymax=2.48352632507056,
ytick={0,0.5,1,1.5,2},
ytick style={color=black},
legend columns=3,
legend cell align=left,
legend style={
  at={(0.5,1.01)},
  anchor=south,
  font=\footnotesize,
  draw=none,
},
/tikz/every odd column/.append style={column sep=2pt},
/tikz/every even column/.append style={column sep=6pt},
]
\draw[draw=seagreen71135115,fill=mediumaquamarine102194165,very thick] (axis cs:-0.35,0) rectangle (axis cs:0.35,1.32422583440086);
\draw[draw=sienna1769868,fill=salmon25214198,very thick] (axis cs:0.65,0) rectangle (axis cs:1.35,1.01383980958286);
\draw[draw=slategray98111142,fill=darkgray141160203,very thick] (axis cs:1.65,0) rectangle (axis cs:2.35,1.30304723875571);
\draw[draw=gray16196136,fill=orchid231138195,very thick] (axis cs:2.65,0) rectangle (axis cs:3.35,1.21690395219681);
\draw[draw=olivedrab11615158,fill=yellowgreen16621684,very thick] (axis cs:3.65,0) rectangle (axis cs:4.35,1.94000981693621);
\path [draw=dimgray, very thick]
(axis cs:0,1.22108663717604)
--(axis cs:0,1.44246761086637);

\path [draw=dimgray, very thick]
(axis cs:1,0.928227313446513)
--(axis cs:1,1.11772021558841);

\path [draw=dimgray, very thick]
(axis cs:2,1.18480269107107)
--(axis cs:2,1.43214911389579);

\path [draw=dimgray, very thick]
(axis cs:3,1.11588089577752)
--(axis cs:3,1.33367495244384);

\path [draw=dimgray, very thick]
(axis cs:4,1.64135134974179)
--(axis cs:4,2.37545618331761);

\addplot [semithick, dimgray, mark=-, mark size=5, mark options={solid}, only marks]
table {%
0 1.22108663717604
1 0.928227313446513
2 1.18480269107107
3 1.11588089577752
4 1.64135134974179
};
\addplot [semithick, dimgray, mark=-, mark size=5, mark options={solid}, only marks]
table {%
0 1.44246761086637
1 1.11772021558841
2 1.43214911389579
3 1.33367495244384
4 2.37545618331761
};
\draw (axis cs:0,0.09) node[
  scale=0.9,
  text=white,
  rotate=0.0
]{\bfseries 1.32};
\draw (axis cs:1,0.09) node[
  scale=0.9,
  text=white,
  rotate=0.0
]{\bfseries 1.01};
\draw (axis cs:2,0.09) node[
  scale=0.9,
  text=white,
  rotate=0.0
]{\bfseries 1.30};
\draw (axis cs:3,0.09) node[
  scale=0.9,
  text=white,
  rotate=0.0
]{\bfseries 1.22};
\draw (axis cs:4,0.09) node[
  scale=0.9,
  text=white,
  rotate=0.0
]{\bfseries 1.94};
\end{axis}

\end{tikzpicture}
    \caption{
        Mean absolute distance between the final joint positions and the goal for the reacher task (lower is better).
        The error bars visualize the 95\% confidence intervals across 100 trials computed via bootstrapping.
    }
    \label{fig:avg_pos_errs_reacher}
\end{figure}

\Cref{fig:avg_pos_errs_reacher} displays the average joint position error for the reacher task, measured as the joint-space distance to the goal at the final step of the episode.
All configurations yield precise reacher policies on the real robot.
The policy trained with the baseline Torque GenAN achieves the highest accuracy with an average error of $\SI{1.01}{\degree}$.
The GenAN configurations with and without the ensemble achieve similar accuracies, while reducing the action penalty results in increased position errors of $\SI{1.94}{\degree}$.

\subsection{Ball-in-a-cup}
\label{subapp:ball_in_cup}

\begin{figure}[ht]
    \begin{subfigure}[t]{\linewidth}
        \centering
        \includeMoreTrimmedShiftedLeftBallInCupImg{0.305\linewidth}{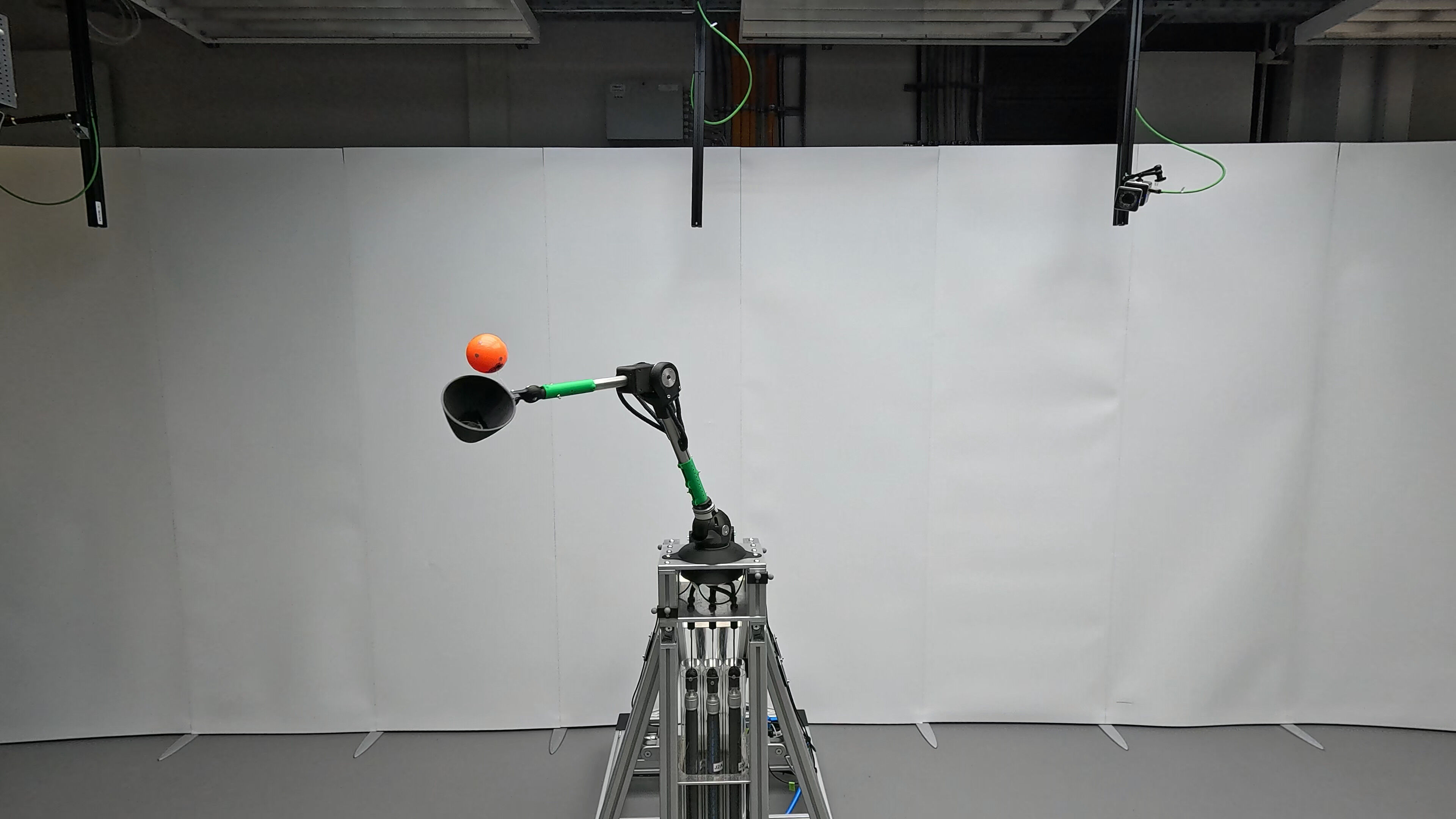}
        \includeMoreTrimmedShiftedLeftBallInCupImg{0.305\linewidth}{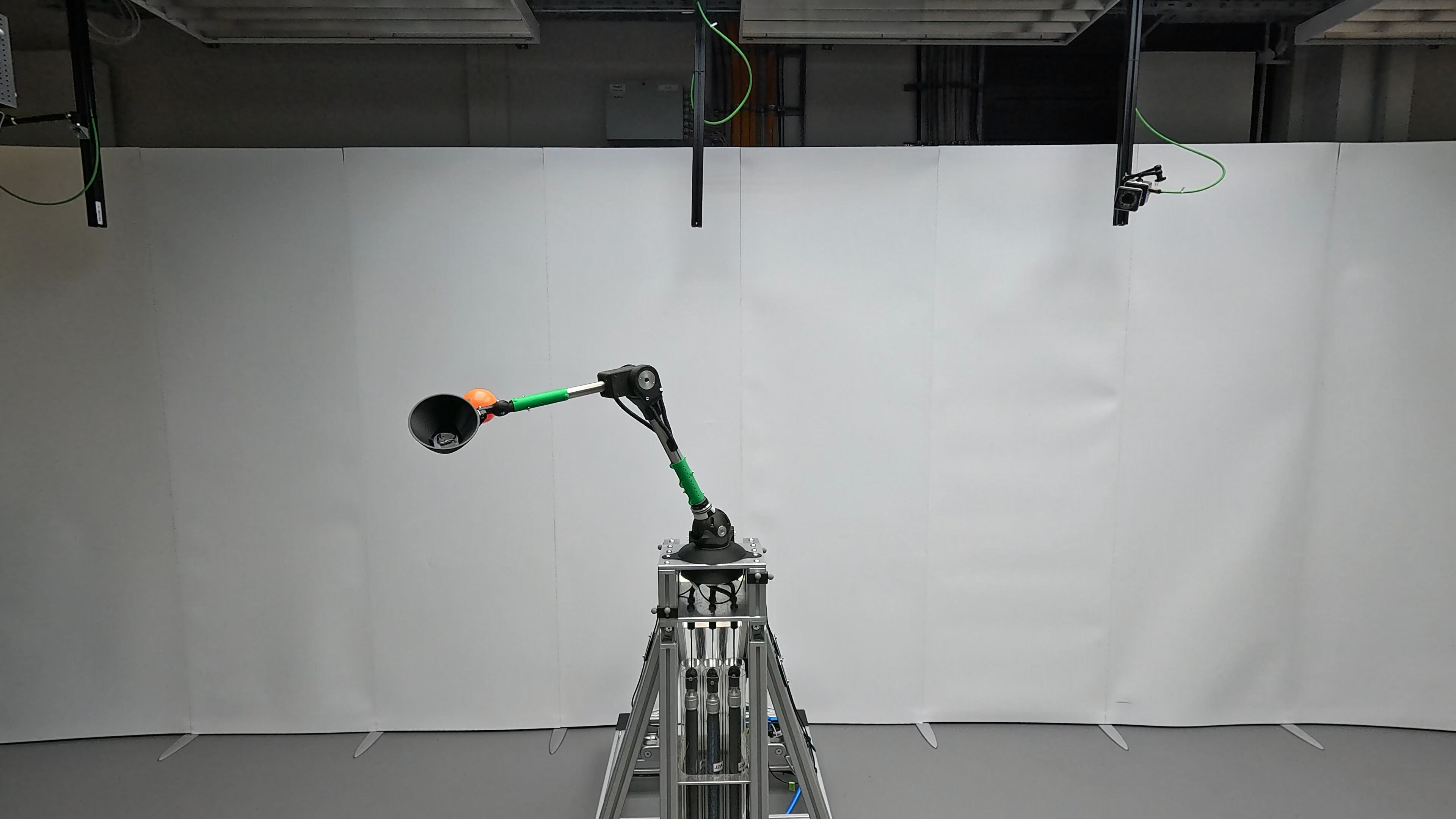}
        \includeMoreTrimmedBallInCupImg{0.305\linewidth}{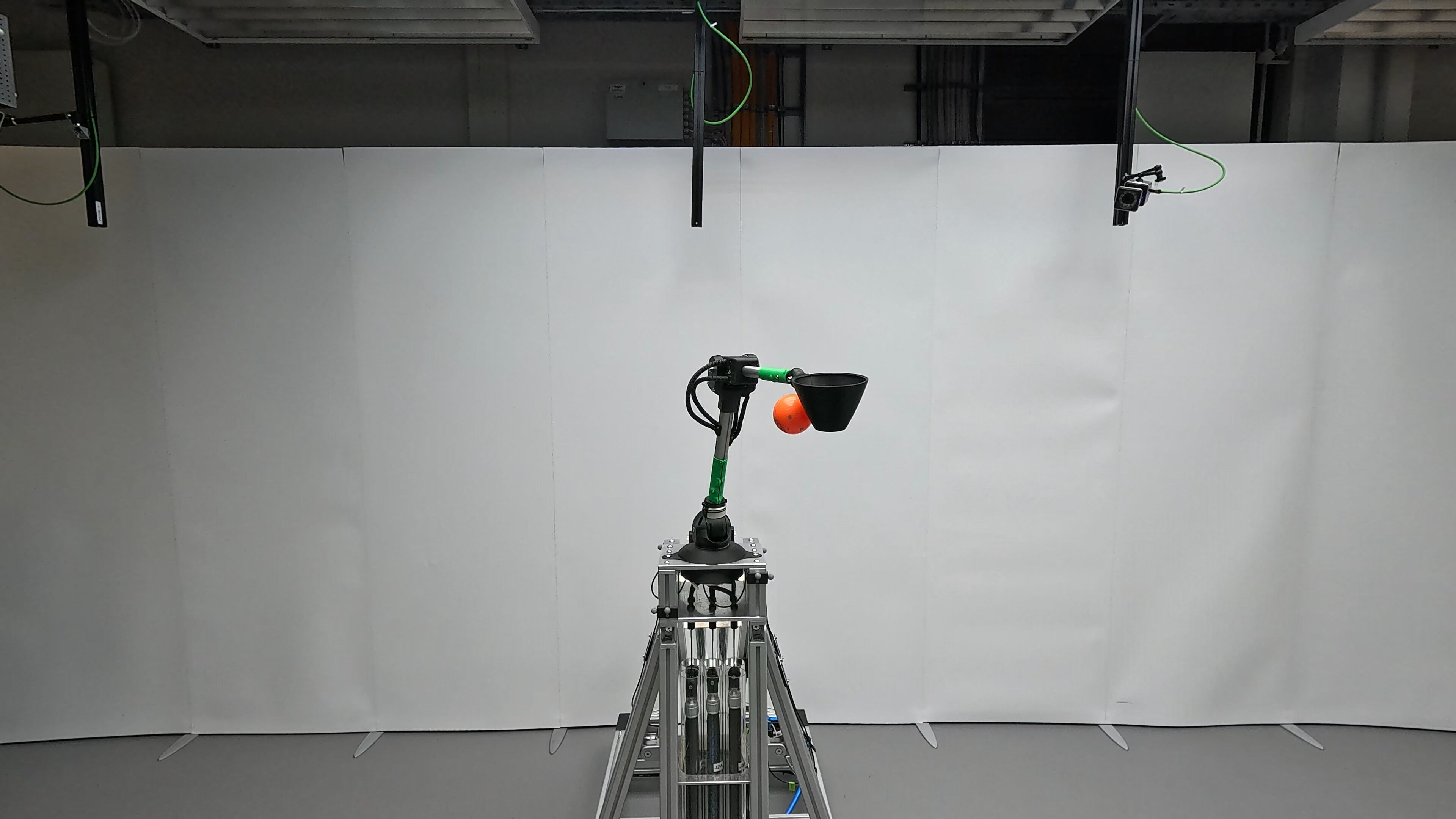}
        \caption{The string wraps around the link.}
        \label{subfig:ball_in_cup_failure_modes_string_wrapping}
    \end{subfigure}\\[0.1cm]
    \begin{subfigure}[t]{\linewidth}
        \centering
        \includeMoreTrimmedShiftedLeftBallInCupImg{0.305\linewidth}{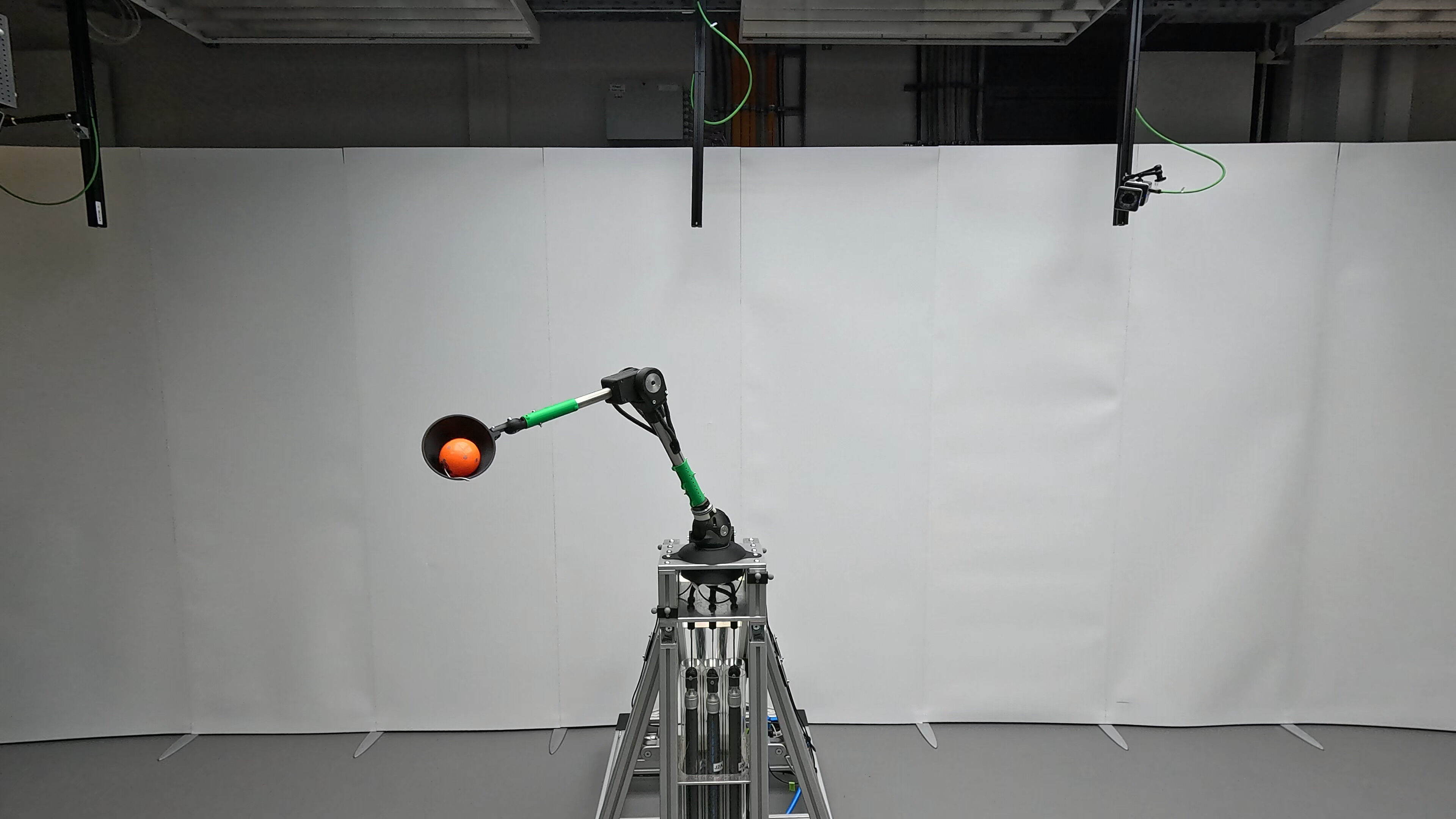}
        \includeMoreTrimmedShiftedLeftBallInCupImg{0.305\linewidth}{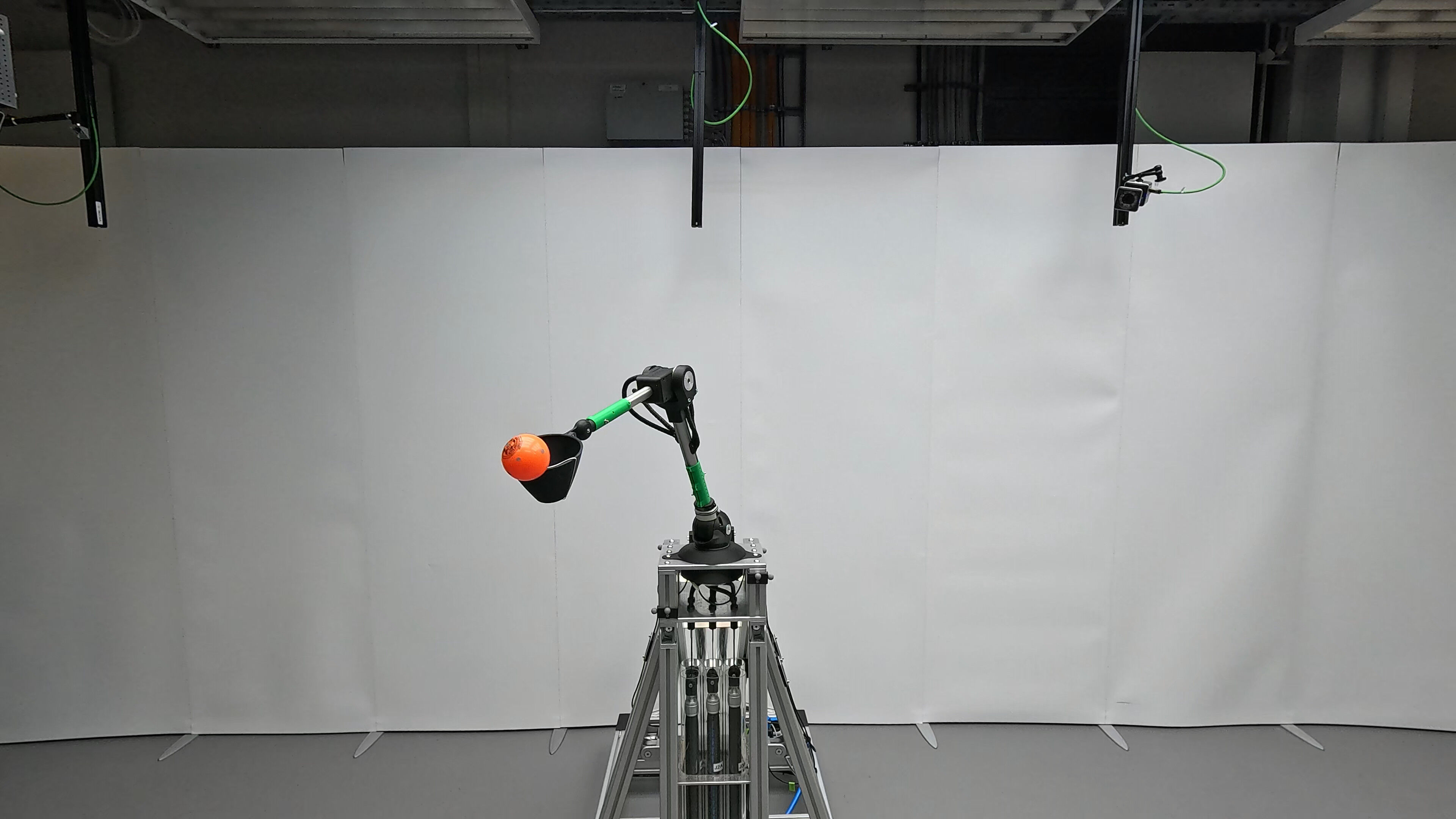}
        \includeMoreTrimmedShiftedLeftBallInCupImg{0.305\linewidth}{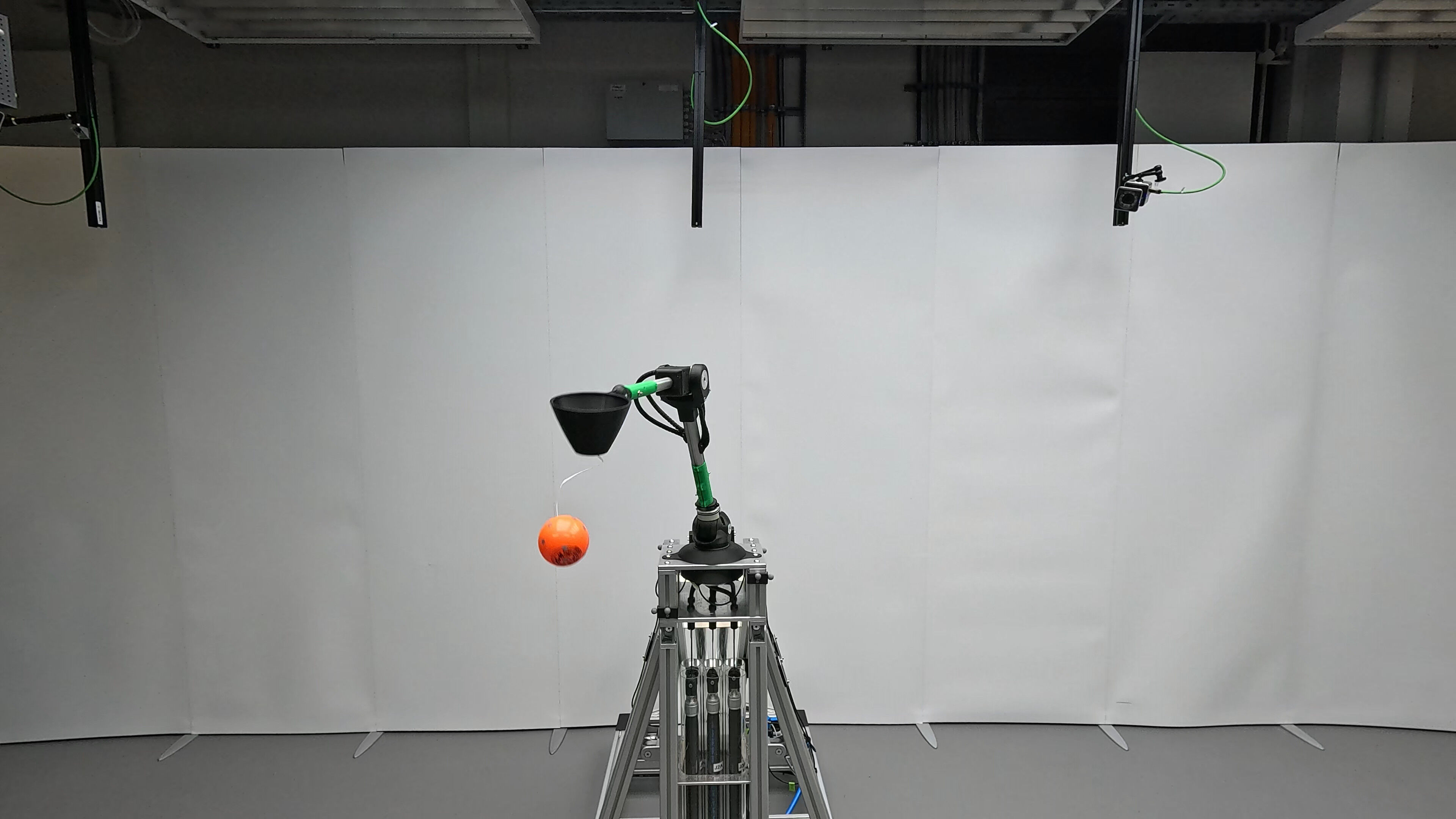}
        \caption{The ball bounces out of the cup.}
        \label{subfig:ball_in_cup_failure_modes_bounce}
    \end{subfigure}\\[0.1cm]
    \caption{
        Two common failure modes of the ball-in-a-cup policy caused by differences in the ball and string dynamics.
    }
    \label{fig:ball_in_cup_failure_modes}
\end{figure}

Common failure modes of the ball-in-a-cup policy include the string wrapping around the link (see \cref{subfig:ball_in_cup_failure_modes_string_wrapping}) and the ball bouncing out of the cup after an attempted catch (see \cref{subfig:ball_in_cup_failure_modes_bounce}).
These failure modes originate from differences in the ball and string dynamics between the simulated and real environments and are therefore independent of the actuator modeling.
For simplicity, we use a MuJoCo tendon to simulate the string, which does not model collisions with the robot geometry.
Therefore, the agent does not learn how to recover in this scenario by unwinding the string.
Using a more sophisticated string model could enable the agent to learn effective recovery maneuvers for these situations.
Preventing the ball from bouncing out of the cup is challenging, as the ball is partly occluded while in the cup, which makes tracking less reliable.
Nevertheless, carefully tuning the contact properties of the ball and cup in simulation to match the real dynamics could reduce the risk of the ball bouncing out of the cup.

\subsection{Table tennis}
\label{subapp:table_tennis}

\begin{table}[ht]
    \caption{Detailed outcomes of the table tennis episodes (out of 100 episodes).}
    \label{tab:table_tennis_outcomes}
    \centering
    \begin{tabular}{l c c c c c}
        \toprule
        Outcome & \makecell[c]{GenAN} & \makecell[c]{Torque\\GenAN} & \makecell[c]{No disag.\\penalty} & \makecell[c]{No\\ensemble} & \makecell[c]{Low action\\penalty} \\
        \midrule
        Success & 96 & 75 & 82 & 83 & 54 \\
        Return too long & 3 & 0 & 1 & 0 & 16 \\
        Return too short & 0 & 13 & 12 & 17 & 8 \\
        Missed ball & 1 & 12 & 5 & 0 & 22 \\
        \bottomrule
    \end{tabular}
\end{table}

For the table tennis task, we observed three distinct failure modes:
\begin{enumerate}
    \item Returning the ball too far, such that it misses the opponent's side of the table.
    \item Returning the ball too short, such that it lands on the robot's side of the table.
    \item Missing the ball entirely.
\end{enumerate}
Notably, we never observed the ball missing the table laterally.
\Cref{tab:table_tennis_outcomes} shows the distribution of these outcomes for the policies from \cref{subsec:transfer}.

\begin{figure}[ht]
    \centering
\begin{tikzpicture}

\definecolor{darkgray141160203}{RGB}{141,160,203}
\definecolor{darkgray176}{RGB}{176,176,176}
\definecolor{dimgray}{RGB}{105,105,105}
\definecolor{gray16196136}{RGB}{161,96,136}
\definecolor{mediumaquamarine102194165}{RGB}{102,194,165}
\definecolor{olivedrab11615158}{RGB}{116,151,58}
\definecolor{orchid231138195}{RGB}{231,138,195}
\definecolor{salmon25214198}{RGB}{252,141,98}
\definecolor{seagreen71135115}{RGB}{71,135,115}
\definecolor{sienna1769868}{RGB}{176,98,68}
\definecolor{slategray98111142}{RGB}{98,111,142}
\definecolor{yellowgreen16621684}{RGB}{166,216,84}
\definecolor{darkgoldenrod17815132}{RGB}{178,151,32}
\definecolor{gold25521747}{RGB}{255,217,47}

\begin{axis}[
golden width landscape=0.65\linewidth,tick align=outside,
tick pos=left,
x grid style={darkgray176},
xmin=-0.585, xmax=5.585,
xtick style={color=black},
xtick={0,1,2,3,4,5},
xticklabel style={align=center,yshift=3pt},
xticklabels={HySR\\(virtual balls),GenAN,Torque\\GenAN,No disag.\\penalty,No\\ensemble,Low action\\penalty},
y grid style={darkgray176},
ylabel={success rate (\%)},
ymajorgrids,
ymin=0, ymax=100,
ytick style={color=black}
]
\draw[draw=darkgoldenrod17815132,fill=gold25521747,very thick] (axis cs:-0.35,0) rectangle (axis cs:0.35,88);
\draw[draw=seagreen71135115,fill=mediumaquamarine102194165,very thick] (axis cs:0.65,0) rectangle (axis cs:1.35,89);
\draw[draw=sienna1769868,fill=salmon25214198,very thick] (axis cs:1.65,0) rectangle (axis cs:2.35,74);
\draw[draw=slategray98111142,fill=darkgray141160203,very thick] (axis cs:2.65,0) rectangle (axis cs:3.35,66);
\draw[draw=gray16196136,fill=orchid231138195,very thick] (axis cs:3.65,0) rectangle (axis cs:4.35,74);
\draw[draw=olivedrab11615158,fill=yellowgreen16621684,very thick] (axis cs:4.65,0) rectangle (axis cs:5.35,44);
\path [draw=dimgray, very thick]
(axis cs:1,81.3687034969197)
--(axis cs:1,93.7458036429354);

\path [draw=dimgray, very thick]
(axis cs:2,64.6290105508128)
--(axis cs:2,81.5953015352519);

\path [draw=dimgray, very thick]
(axis cs:3,56.2777288547246)
--(axis cs:3,74.5384792026518);

\path [draw=dimgray, very thick]
(axis cs:4,64.6290105508128)
--(axis cs:4,81.5953015352519);

\path [draw=dimgray, very thick]
(axis cs:5,34.6720266301608)
--(axis cs:5,53.771895348323);

\addplot [semithick, dimgray, mark=-, mark size=5, mark options={solid}, only marks]
table {%
1 81.3687034969197
2 64.6290105508128
3 56.2777288547246
4 64.6290105508128
5 34.6720266301608
};
\addplot [semithick, dimgray, mark=-, mark size=5, mark options={solid}, only marks]
table {%
1 93.7458036429354
2 81.5953015352519
3 74.5384792026518
4 81.5953015352519
5 53.771895348323
};
\draw (axis cs:0,4) node[
  scale=0.9,
  text=white,
  rotate=0.0
]{\bfseries 88};
\draw (axis cs:1,4) node[
  scale=0.9,
  text=white,
  rotate=0.0
]{\bfseries 89};
\draw (axis cs:2,4) node[
  scale=0.9,
  text=white,
  rotate=0.0
]{\bfseries 74};
\draw (axis cs:3,4) node[
  scale=0.9,
  text=white,
  rotate=0.0
]{\bfseries 66};
\draw (axis cs:4,4) node[
  scale=0.9,
  text=white,
  rotate=0.0
]{\bfseries 74};
\draw (axis cs:5,4) node[
  scale=0.9,
  text=white,
  rotate=0.0
]{\bfseries 44};
\end{axis}

\end{tikzpicture}
    \caption{
        Table tennis transfer success rate in terms of the success criterion of~\citet{guist2024safe}.
        Success is defined as returning the table tennis ball to a circle of radius $\SI{68.5}{\cm}$ around the target position in the middle of the opponent's side of the table.
        Even though Guist et al.\ train with HySR~\cite{buchler2022learning} directly on the real robot and evaluate only with virtual balls, our policy achieves the same performance with real table tennis balls, despite training entirely in simulation.
    }
    \label{fig:table_tennis_successes_0685}
\end{figure}

\begin{figure}[ht]
    \centering
    \includegraphics[width=0.6\linewidth]{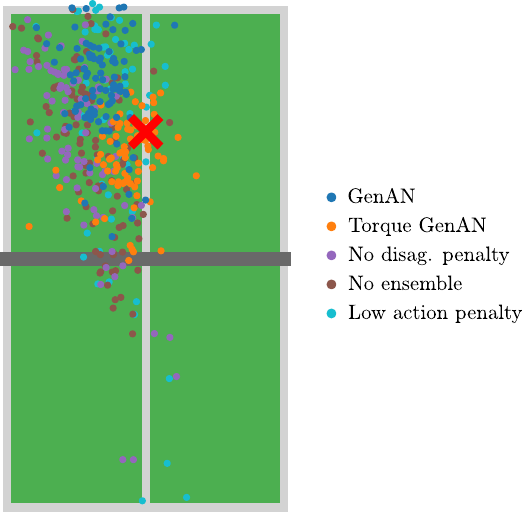}
    \caption{
        Distribution of the table tennis ball landing points.
        The red cross marks the target landing point on the opponent's side of the table.
        For each configuration, this plot shows only balls that landed on the table.
    }
    \label{fig:table_tennis_landing_points}
\end{figure}

To assess the impact of the remaining sim-to-real gap on the table tennis transfer performance, we compare to the results of \citet{guist2024safe}.
They use hybrid-sim-and-real (HySR) training to train directly on the real robot with virtual balls.
Since they only evaluate with virtual balls, this setting is not affected by the sim-to-real gap.
Furthermore, the virtual ball trajectories used for training and evaluation are the same, reducing the difficulty of the evaluation setting.
While we define table tennis success as returning the ball to the opponent's side of the table, Guist et al.\ use a different success criterion, where the robot has to return the ball into a circle of radius $\SI{68.5}{\cm}$ on the opponent's side of the table.
Note that this circle covers most of the $\SI{137}{\cm} \times \SI{152.5}{\cm}$ table half.

\Cref{fig:table_tennis_successes_0685} shows the performance of our policies with this success criterion.
We compare the performance to the HySR results from~\cite{guist2024safe}.
Although Guist et al.\ train directly on the real robot and evaluate with seen virtual ball trajectories, our method achieves the same performance on the real robot.
This result highlights that the GenAN is a faithful model of PAMY2's actuator dynamics, which results in a small sim-to-real gap and enables a robust sim-to-real transfer.

\Cref{fig:table_tennis_landing_points} shows the landing points of the real table tennis balls on the table.
Note that the figure visualizes only balls that landed on the table.
The visualization shows a clear bias toward the left side of the table.
This bias is likely caused by the kinematics of the robot.
The robot's 4 DoFs constrain the possible end-effector orientations for a given position.
In particular, the robot lacks a wrist joint, which is essential for precise lateral ball placement.

\end{document}